%% file: 0.0-main.tex
\documentclass[10pt,twocolumn,letterpaper]{article}
\usepackage{changepage}
\usepackage{algorithm2e,setspace}
\usepackage[document]{ragged2e}

\usepackage{cvpr}              % To produce the CAMERA-READY version
\usepackage[export]{adjustbox}
\usepackage{booktabs}
\usepackage{times}
\usepackage{epsfig}
\usepackage{ifthen}
\usepackage{graphicx}
\usepackage{amsmath}
\usepackage{algorithm2e,setspace}
\usepackage{mathtools}
\usepackage{amssymb}
\usepackage[justification=justified]{subcaption}
\usepackage{wrapfig}
\usepackage{multirow}
\usepackage{parskip}
\usepackage{array}
\usepackage{xr-hyper}
\usepackage{enumitem}       % enumerations
\usepackage[font=small,labelfont=bf,
   justification=justified,
   format=plain]{caption}
\DeclareCaptionJustification{justified}{\justifying}
\usepackage{xcolor}
\makeatletter
    \renewcommand\@makefntext[1]{\raggedright\leftskip=-.1em\hskip0em\@makefnmark#1}
\makeatother

\usepackage[pagebackref,breaklinks,colorlinks]{hyperref}

\usepackage{amsthm}

\usepackage[capitalize]{cleveref}
\crefname{section}{Sec.}{Secs.}
\crefname{algorithm}{Alg.}{Algs.}
\Crefname{section}{Section}{Sections}
\Crefname{table}{Table}{Tables}
\crefname{table}{Tab.}{Tabs.}

\newcommand\myeq{\mkern1.25mu{=}\mkern1.25mu}
\newcommand\myminus{\mkern1.25mu{-}\mkern1.25mu}
\newcommand\myplus{\mkern1.25mu{+}\mkern1.25mu}
\newcommand\mygreater{\mkern1.25mu{>}\mkern1.25mu}

\newcommand\mysubsub[1]{\vspace{-.8em}%
\subsubsection{#1}%
\vspace{-.6em}
}

\newcommand\mysub[1]{\subsection{#1}}

\newcolumntype{R}[2]{%
    >{\adjustbox{angle=#1,lap=\width-(#2)}\bgroup}%
    l%
    <{\egroup}%
}

\makeatletter
\def\thickhline{%
  \noalign{\ifnum0=`}\fi\hrule \@height \thickarrayrulewidth \futurelet
   \reserved@a\@xthickhline}
\def\@xthickhline{\ifx\reserved@a\thickhline
               \vskip\doublerulesep
               \vskip-\thickarrayrulewidth
             \fi
      \ifnum0=`{\fi}}
\makeatother

\newlength{\thickarrayrulewidth}
\def\cvprPaperID{6365} % *** Enter the CVPR Paper ID here
\def\confName{CVPR}
\def\confYear{2022}

\newcommand{\mat}[1]{\MakeUppercase{\mathbf{#1}}}
\renewcommand{\vec}[1]{\MakeLowercase{\mathbf{#1}}}

\newcommand{\bcos}{B-cos}
\newcommand{\mbcos}{\text{B-cos}}

\newboolean{showcomments}
\setboolean{showcomments}{false}
\newcommand{\myparagraph}[2][-.1]{\vspace{#1em}\noindent{\bf #2}}
\begin{document}

\title{\bcos{} Networks: Alignment is All We Need for Interpretability}

\author{Moritz Böhle\\
\normalsize{MPI for Informatics}\\ 
\normalsize{Saarland Informatics Campus}
\and
Mario Fritz\\
\normalsize{CISPA Helmholtz Center}\\
\normalsize{for Information Security}
\and
Bernt Schiele\\
\normalsize{MPI for Informatics}\\ 
\normalsize{Saarland Informatics Campus}
}
\twocolumn[{%
\renewcommand\twocolumn[1][]{#1}%
\maketitle
\begin{center}
\begin{adjustbox}{minipage=\linewidth, scale=.95}
    \centering
    \vspace{-.5em}
    \captionsetup{type=figure}
    \input{resources/figures/global_protos_v4}
\end{adjustbox}
\end{center}%
\smallskip
}]
\justify
%%%%%%%%% ABSTRACT
\input{0.v1-abstract}

\section{Introduction}
\label{sec:intro}
\input{1.v4-introduction}
\section{Related work}
\label{sec:related}
\input{2.v2-related_work}
\vspace{-.5em}
\section{\bcos{} neural networks}
\label{sec:method}
\vspace{-.25em}
\input{3.v4-method}
\section{Experimental setting}
\label{sec:experiments}
\input{4.0-experiments}
\vspace{-1em}
\section{Results}
\label{sec:results}
\input{5.0-results}
\section{Conclusion}
\label{sec:discussion}
\input{6.0-discussion}

\clearpage
%%%%%%%%% REFERENCES
{\small
\bibliographystyle{ieee_fullname}
\bibliography{egbib}
}

\input{supplement/0.1-main}

\end{document}

%% file: resources/figures/global_protos_v4.tex
\newcommand\mysize{0.076}
\newcommand\secondsize{0.95}
    \centering
    \begin{subfigure}[b]{\mysize\textwidth}\begin{subfigure}[b]{\secondsize\textwidth}
    \includegraphics[width=\textwidth, trim=1em 1em 1em 1em]{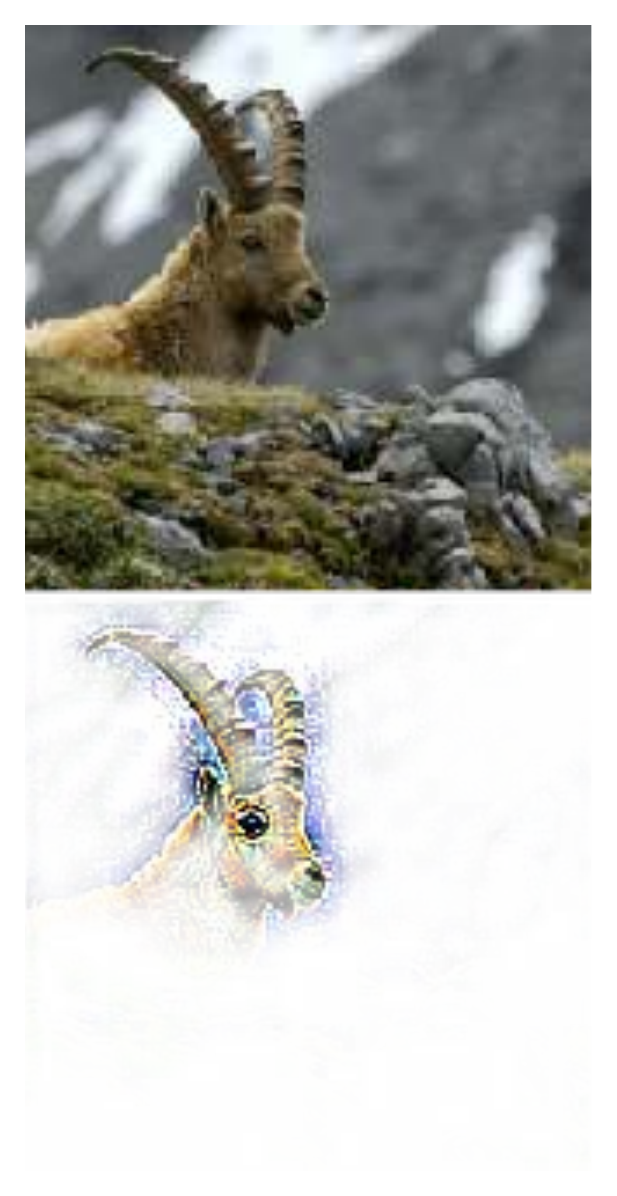}
    \end{subfigure}\end{subfigure}
    \begin{subfigure}[b]{\mysize\textwidth}\begin{subfigure}[b]{\secondsize\textwidth}
    \includegraphics[width=\textwidth, trim=1em 1em 1em 1em]{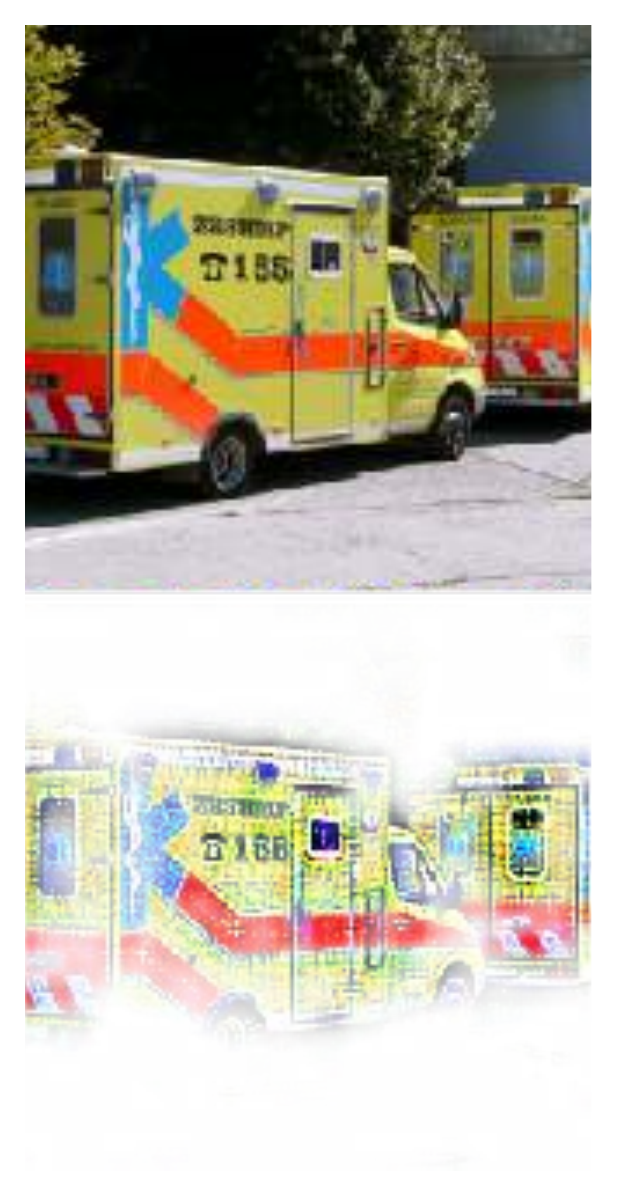}
    \end{subfigure}\end{subfigure}
    \begin{subfigure}[b]{\mysize\textwidth}\begin{subfigure}[b]{\secondsize\textwidth}
    \includegraphics[width=\textwidth, trim=1em 1em 1em 1em]{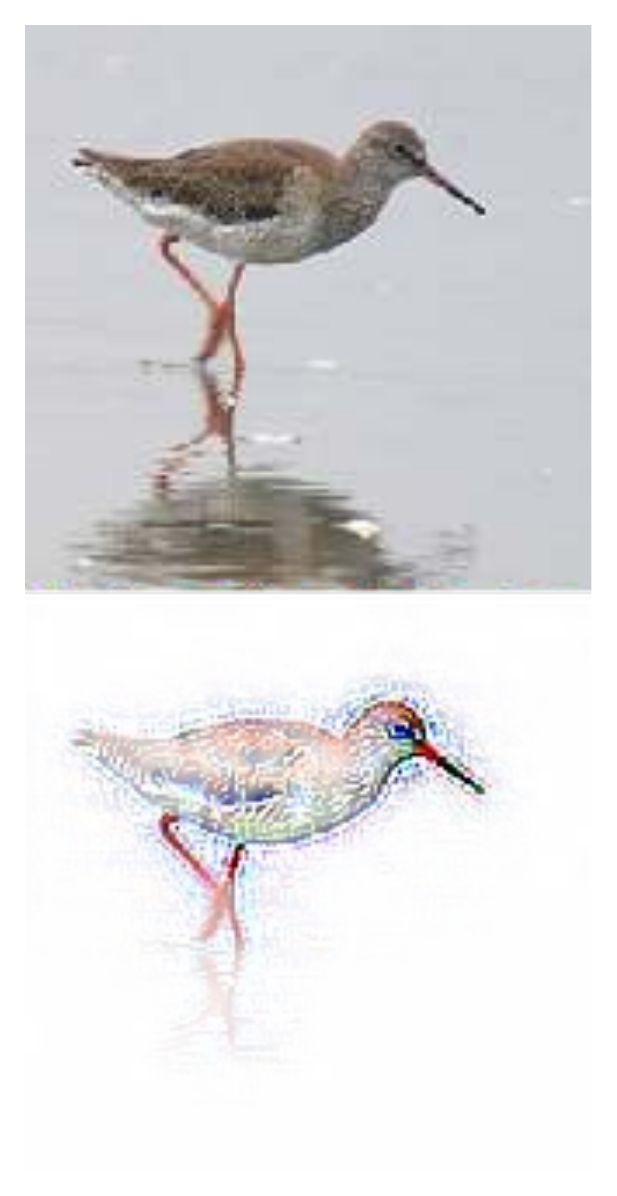}
    \end{subfigure}\end{subfigure}
    \begin{subfigure}[b]{\mysize\textwidth}\begin{subfigure}[b]{\secondsize\textwidth}
    \includegraphics[width=\textwidth, trim=1em 1em 1em 1em]{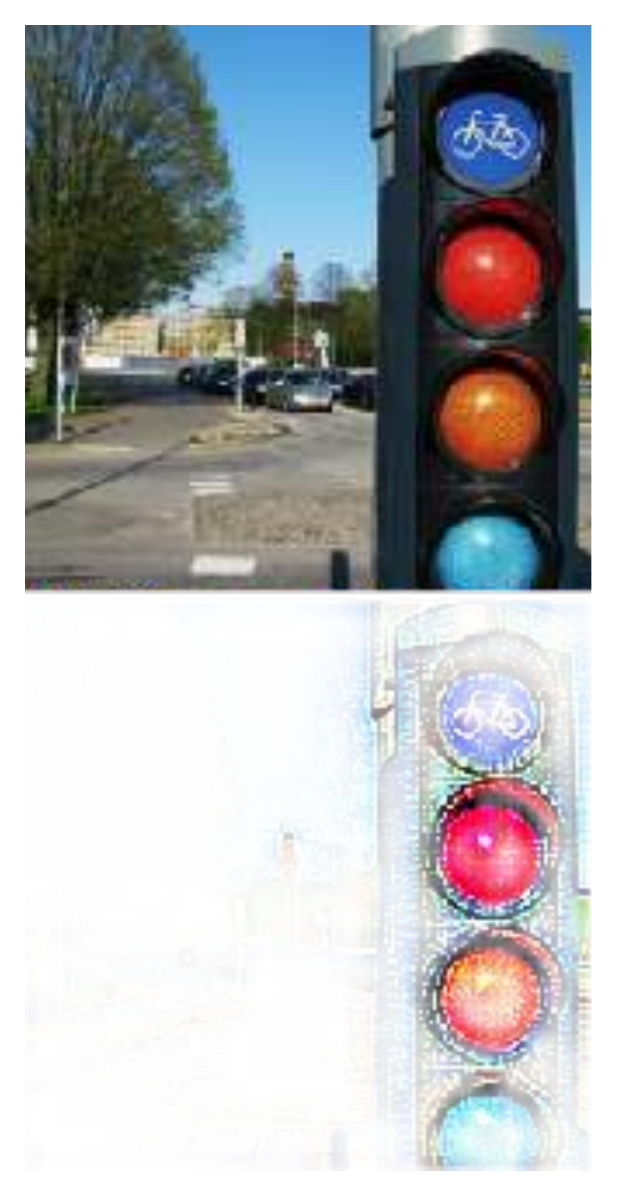}
    \end{subfigure}\end{subfigure}
    \begin{subfigure}[b]{\mysize\textwidth}\begin{subfigure}[b]{\secondsize\textwidth}
    \includegraphics[width=\textwidth, trim=1em 1em 1em 1em]{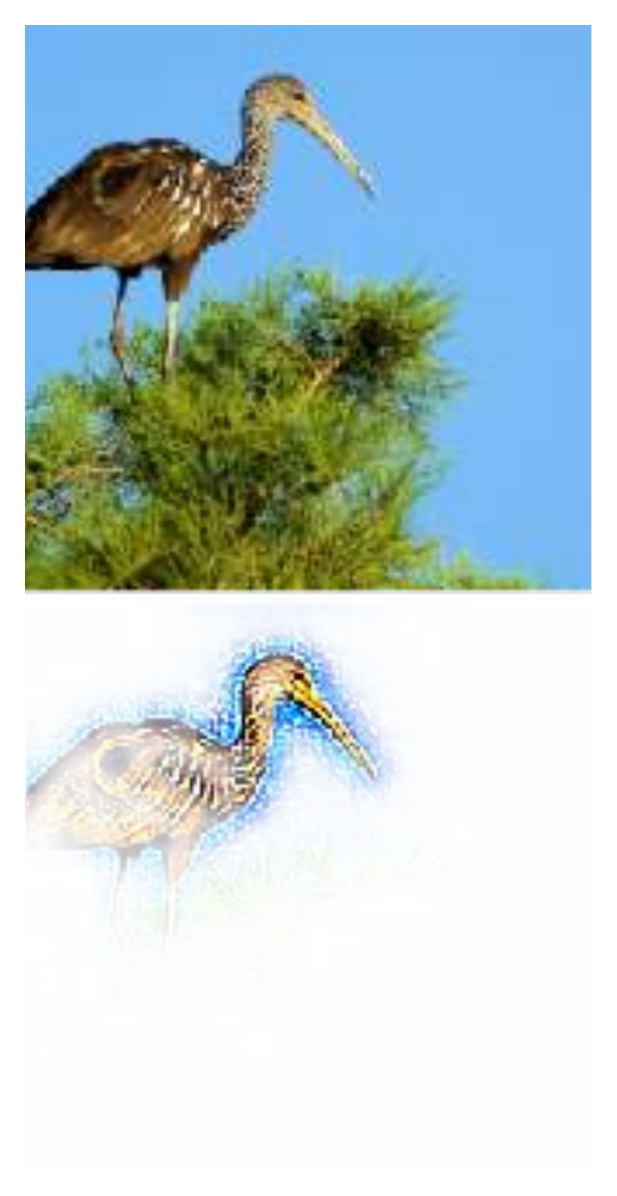}
    \end{subfigure}\end{subfigure}
    \begin{subfigure}[b]{\mysize\textwidth}\begin{subfigure}[b]{\secondsize\textwidth}
    \includegraphics[width=\textwidth, trim=1em 1em 1em 1em]{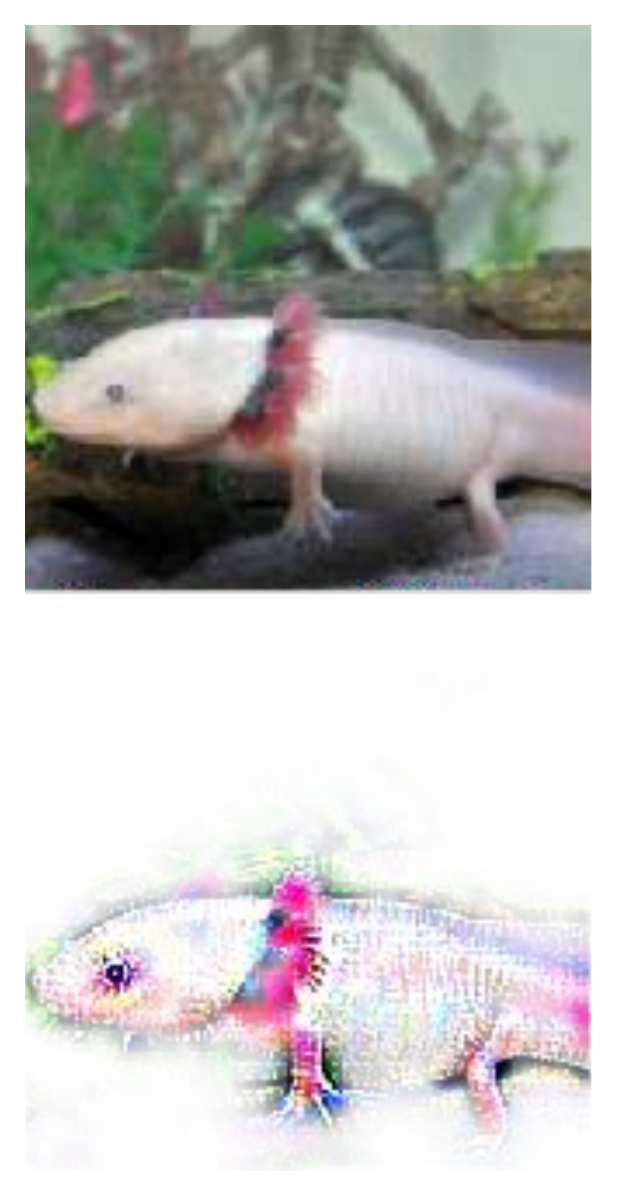}
    \end{subfigure}\end{subfigure}
    \begin{subfigure}[b]{\mysize\textwidth}\begin{subfigure}[b]{\secondsize\textwidth}
    \includegraphics[width=\textwidth, trim=1em 1em 1em 1em]{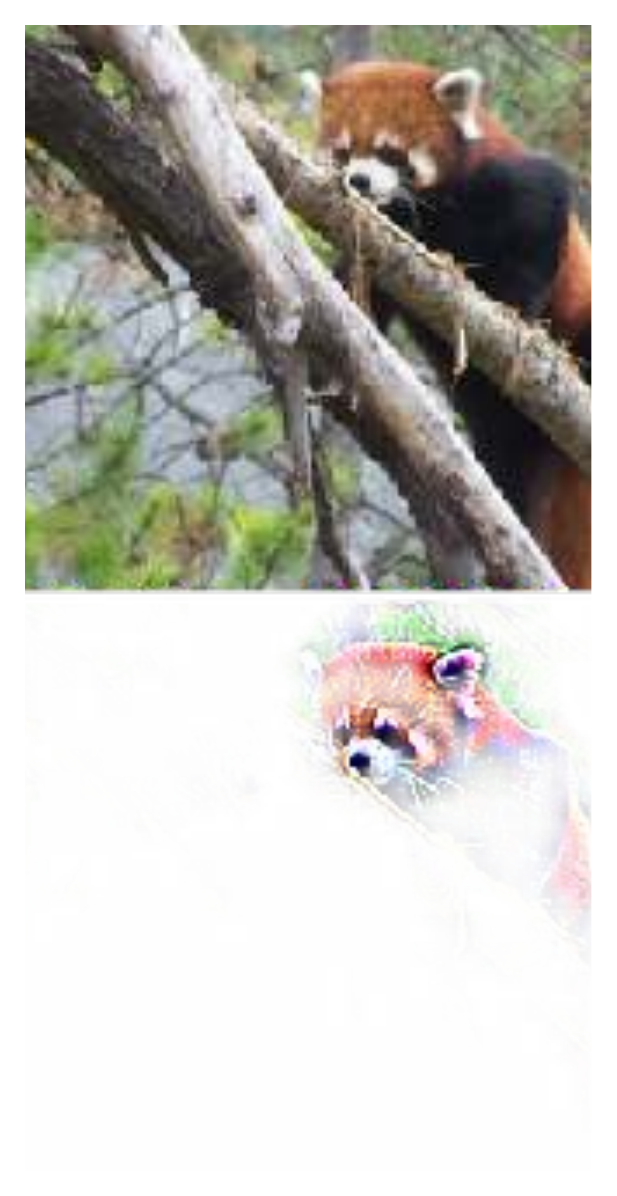}
    \end{subfigure}\end{subfigure}
    \begin{subfigure}[b]{\mysize\textwidth}\begin{subfigure}[b]{\secondsize\textwidth}
    \includegraphics[width=\textwidth, trim=1em 1em 1em 1em]{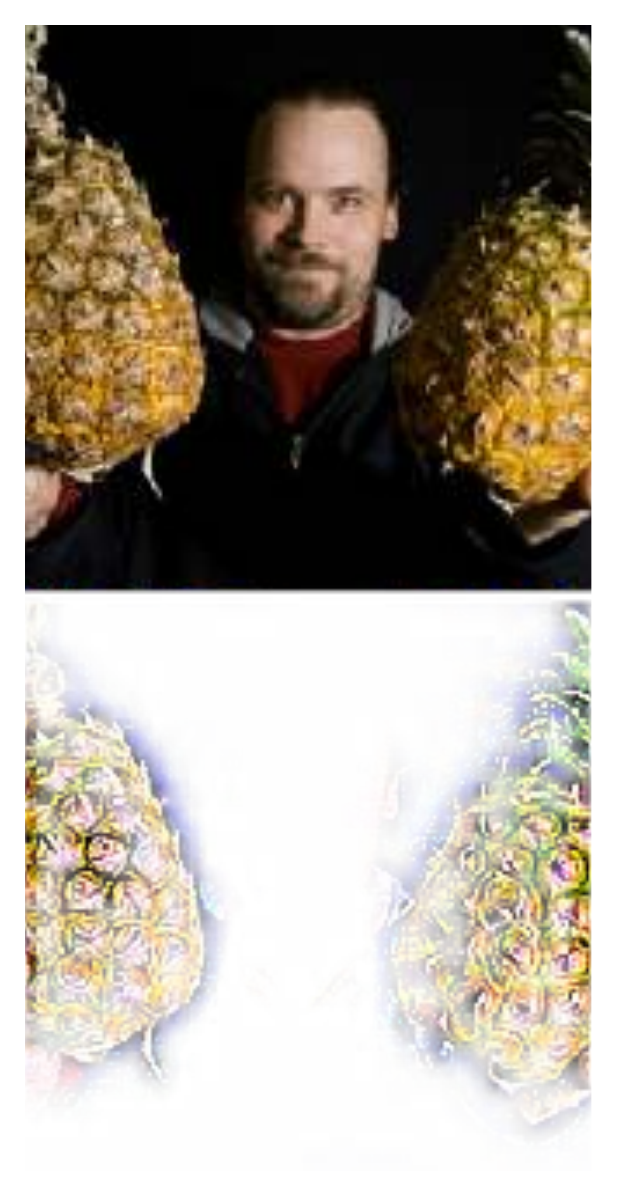}
    \end{subfigure}\end{subfigure}
    \begin{subfigure}[b]{\mysize\textwidth}\begin{subfigure}[b]{\secondsize\textwidth}
    \includegraphics[width=\textwidth, trim=1em 1em 1em 1em]{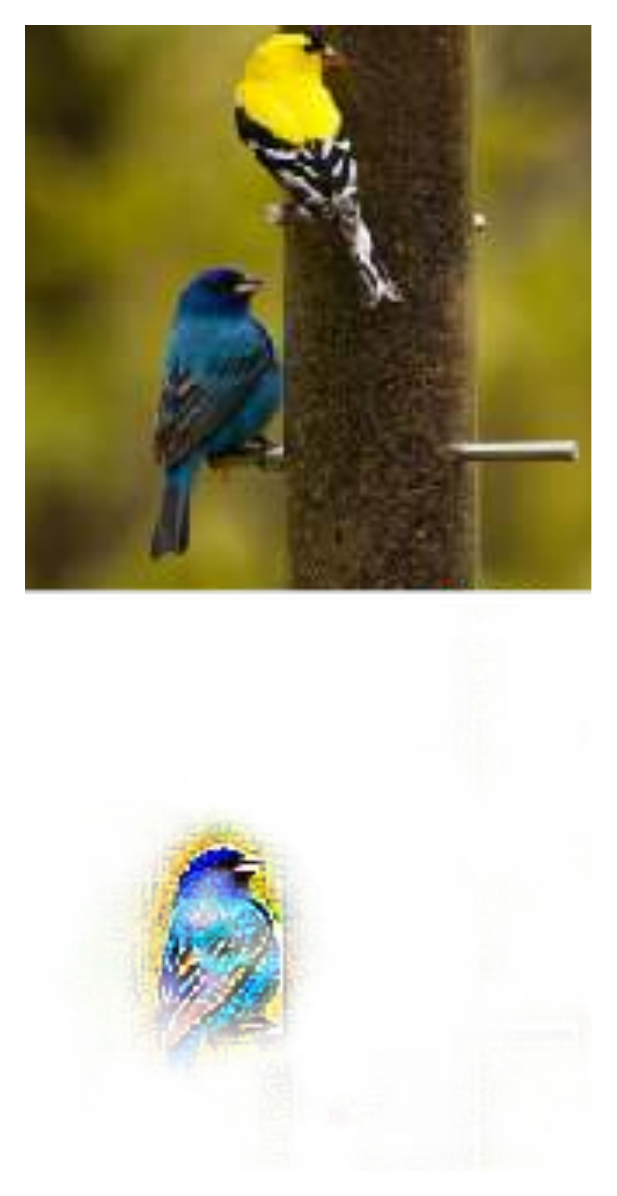}
    \end{subfigure}\end{subfigure}
    \begin{subfigure}[b]{\mysize\textwidth}\begin{subfigure}[b]{\secondsize\textwidth}
    \includegraphics[width=\textwidth, trim=1em 1em 1em 1em]{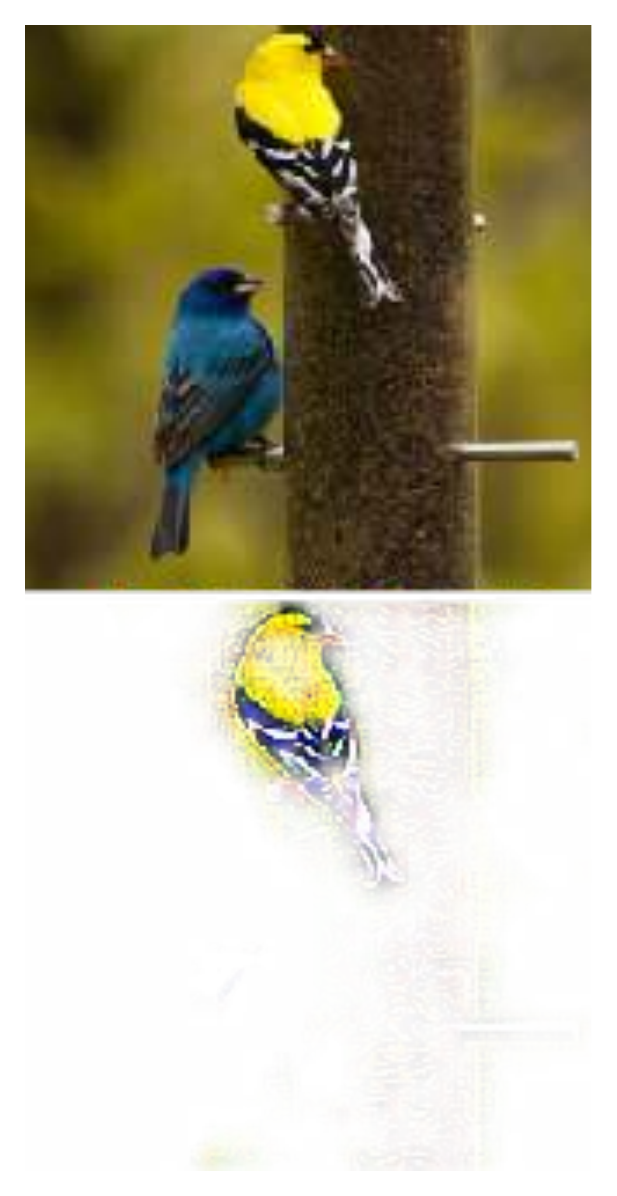}
    \end{subfigure}\end{subfigure}
    \begin{subfigure}[b]{\mysize\textwidth}\begin{subfigure}[b]{\secondsize\textwidth}
    \includegraphics[width=\textwidth, trim=1em 1em 1em 1em]{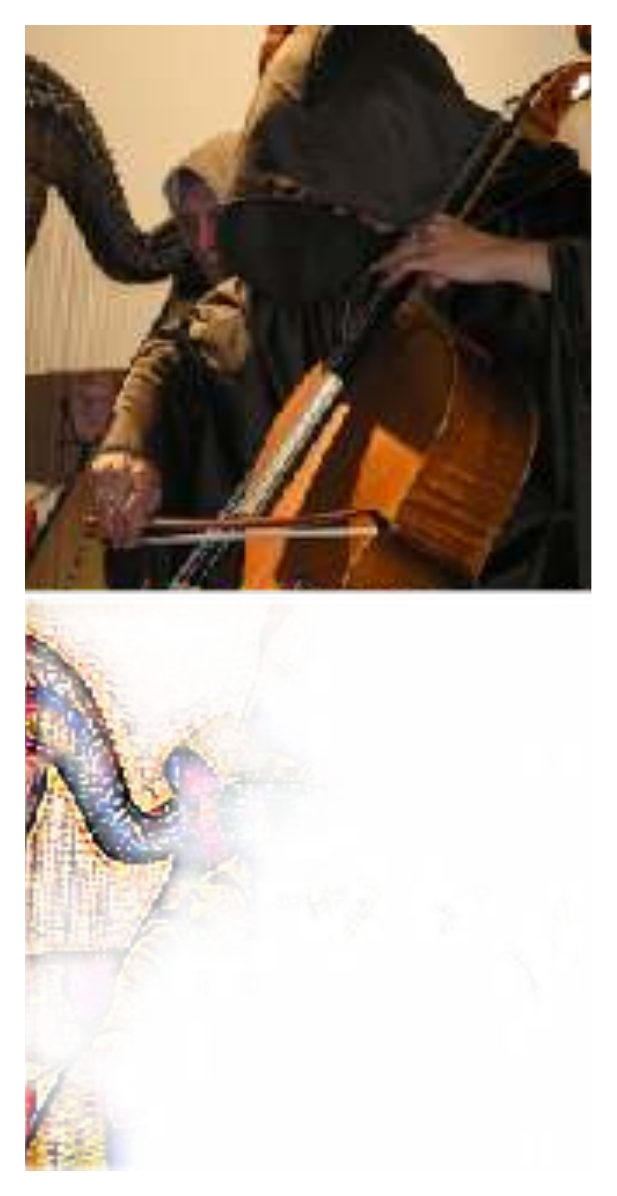}
    \end{subfigure}\end{subfigure}
    \begin{subfigure}[b]{\mysize\textwidth}\begin{subfigure}[b]{\secondsize\textwidth}
    \includegraphics[width=\textwidth, trim=1em 1em 1em 1em]{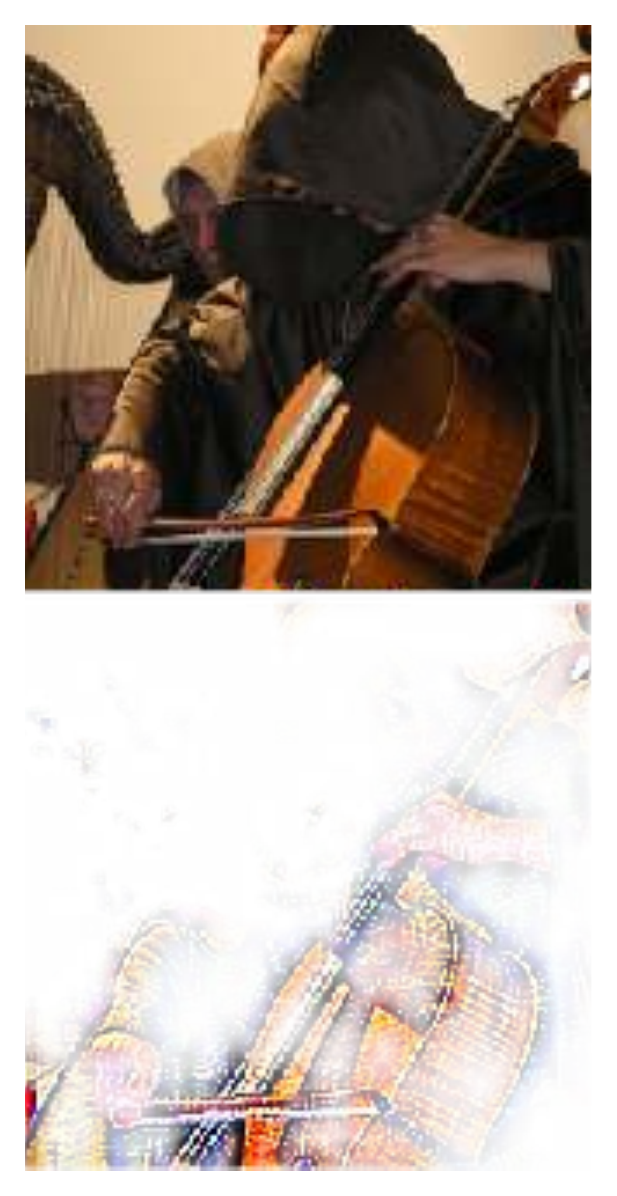}
    \end{subfigure}\end{subfigure}
    \vspace{.25em}
    \caption{\textbf{Top:} Inputs $\vec x_i$ to a \bcos{} DenseNet-121. \textbf{Bottom:} \bcos{} network explanation for class $c$ ($c$: image label). Specifically, we visualise the $c$-th row of $\mat w_{1\rightarrow L}(\vec x_i)$ as applied by the model, see \cref{eq:collapse}; \textbf{no masking of the original image is used for these visualisations}. For the last 2 images, we also show the explanation for the 2nd most likely class. For details on the visualisation of $\mat w_{1\rightarrow L}(\vec x_i)$, see \cref{sec:experiments}. \newline
    }
    
    \label{fig:global_protos}
    \vspace{-1em}

%% file: 0.v1-abstract.tex
\begin{abstract}
\vspace{-.45em}\hspace{.75em}
   We present a new direction for increasing the interpretability of deep neural networks (DNNs) by promoting weight-input alignment during training. 
   % Quick description of method
   For this, % to achieve this
   we propose to replace the linear transforms in DNNs by our \bcos{} transform. {As we show, a sequence (network) of such transforms induces a single linear transform that faithfully summarises the full model computations. Moreover, the \bcos{} transform 
   }
   introduces 
   alignment pressure on the weights during optimisation. 
   As a result, those induced linear transforms become highly interpretable and align with 
   task-relevant features. 
%
   % Results
   {Importantly, the \bcos{} transform is designed to be compatible with existing architectures and we show that it can easily be integrated into common
   models such as VGGs, ResNets, InceptionNets, and DenseNets,  whilst maintaining similar performance on ImageNet.} 
   The resulting explanations are of high visual quality and perform well under quantitative metrics for interpretability.\\
   {Code available at \href{https://www.github.com/moboehle/B-cos}{github.com/moboehle/B-cos}.}
\end{abstract}

%% file: 1.v4-introduction.tex
While deep neural networks (DNNs) are highly successful in a wide range of tasks, explaining their decisions remains an open research problem~\cite{xai_overview}.
The difficulty here lies in the fact that such explanations need to faithfully summarise the internal model computations \emph{and} present them in a {human-interpretable} manner. 
E.g., it is well known that 
piece-wise linear models (e.g., ReLU-based \cite{relu})
are accurately summarised by a linear transform 
for every input~\cite{montufar2014number}.
However, despite providing an {accurate summary}, these 
piece-wise linear transforms are {generally 
not intuitively interpretable} for humans and typically perform poorly under quantitative interpretability metrics, cf.~\cite{BAM2019, shrikumar2017deeplift}. 
{Recent work thus aimed to improve the explanations' interpretability, often focusing on their visual quality~\cite{adebayo2018sanity}. However, gains in the visual quality of the explanations often came at the cost of their model-faithfulness~\cite{adebayo2018sanity}.}
{Instead of optimising the explanation method, in this work we aim to optimise the DNNs to inherently provide an explanation that fulfills the aforementioned requirements---the resulting explanations constitute both a faithful summary and have a clear interpretation for humans.
For this, we propose the \textbf{\bcos{} transform} as a drop-in replacement for linear transforms. 
As such, the \bcos{} transform can easily be integrated into a wide range of existing DNN architectures and we show that the resulting B-cos DNNs provide high-quality explanations for their decisions, see \cref{fig:global_protos}.}

To ensure that these explanations constitute a {faithful summary} of the models, we {design} 
the \bcos{} transform as an input-dependent linear transform. 
Importantly, any sequence of such transforms therefore induces a 
single linear transform that faithfully summarises the entire sequence. 
 In order to make the induced linear transforms interpretable,
the \bcos{} transform is designed to induce 
{alignment pressure} on the weights during optimisation,
which optimises the model weights to align with task-relevant input patterns. 
The linear transform induced by the model thus has a clear interpretation: it is a direct reflection of the weights the model has learnt during training and specifically reflects those weights that best align with a given input.

In summary, we make the following contributions:
\begin{enumerate}[wide, label={\textbf{(\arabic*)}}, itemsep=0em, topsep=-1.25em, labelwidth=0em, labelindent=0pt]
    \item We introduce the \bcos{} transform to improve neural network interpretability. By promoting weight-input alignment, these transforms are explicitly designed to yield explanations that highlight task-relevant patterns in the input.
    \item Specifically, the \bcos{} transform is designed such that any sequence of \bcos{} transforms can be {faithfully summarised} by a single linear transform. 
    We show that this allows to explain not only the models' output neurons, but also neurons from arbitrary intermediate network layers. 
    \item 
    {
    We demonstrate that a plain \bcos{} convolutional neural network without any additional non-linearities, batch-norm layers~\cite{ioffe2015batchnorm}, or regularisation schemes can achieve competitive performance 
    on  CIFAR10~\cite{krizhevsky2009cifar10}. 
    In an ablation study we also show that the
    parameter B allows for fine-grained control over the increase in weight alignment and thus the interpretability of the \bcos{} networks.
    }
    \item 
    {
    To highlight the generality of our approach, we show that the \bcos{} transform can easily be integrated into
    various commonly used DNNs such as
    InceptionNet~\cite{szegedy2015inception},
    ResNet~\cite{he2016deep},
    VGG~\cite{simonyan2015vgg},
    and DenseNet~\cite{huang2017densely} models, whilst maintaining similar performance.
    More importantly, the resulting architectures are {highly interpretable} under the \bcos{} explanations and outperform other explanation methods across all tested architectures, both under quantitative metrics as well as under qualitative inspection.}
\end{enumerate}

%% file: 2.v2-related_work.tex
\myparagraph[-.25]{Approaches for understanding DNNs}
 typically focus on explaining individual model decisions \emph{post-hoc}, i.e., they are designed to work on any pre-trained DNN. Examples of this include perturbation-based, \cite{lundberg2017unified,petsiuk2018rise,ribeiro2016lime}, activation-based, 
 \cite{kim2018tcav,das2020prototypes}, or backpropagation-based explanations, \cite{simonyan2013deep,springenberg2014striving,zhou2016CAM,bach2015pixel,selvaraju2017grad,shrikumar2017deeplift,srinivas2019full,sundararajan2017axiomatic}. 
 In order to obtain explanations for the \bcos{} networks, we also rely on a backpropagation-based approach. In contrast to post-hoc explanation methods, however, {we optimise the \bcos{} networks to be explainable under this particular form of backpropagation and the resulting explanations are thus {model-inherent.}}
  
 {The design of such \emph{inherently interpretable} models has gained increased attention recently. Examples include prototype-based networks~\cite{chen2019looks}, BagNets~\cite{brendel2018approximating}, and CoDA Nets~\cite{Boehle2021CVPR}. 
 Similar to the BagNets and the CoDA Nets, our \bcos{} networks can be faithfully summarised by a single linear transform. Moreover, similar to~\cite{Boehle2021CVPR}, we rely on a structurally induced alignment pressure to make those transforms interpretable. 
 In contrast to those works, however, our method is specifically designed to be compatible with existing neural network architectures, which allows us to improve the interpretability of a wide range of DNNs.}

\myparagraph{Weight-input alignment} in DNNs has recently received increased attention. 
E.g., it has been observed that adversarial training promotes alignment~\cite{tsipras2019atodds} and recent studies suggest that this could increase interpretability via gradient-based explanations~\cite{Shah2021grads,kim2019safeml}. Further, \cite{srinivas2021rethink} %
introduce a loss to increase alignment. Instead of relying on loss-based model regularisation, the increase in alignment in \bcos{} networks is based on \emph{architectural constraints} that require weight-input alignment for solving the optimisation task. 

\myparagraph{Non-linear transforms.}
While the linear transform is the default operation for most neural network architectures, 
many non-linear transforms have been investigated, 
\cite{zoumpourlis2017non,liu2017hyperspherical,Liu2018CVPR,luo2018cosine,ghiasi2019generalizing,wang2019kervolutional}. Most similar to our work are~\cite{liu2017hyperspherical,Liu2018CVPR,luo2018cosine}, which assess transforms that emphasise the cosine similarity (i.e., `alignment') between weights and inputs to improve model performance. 
In fact, we found that amongst other transforms, \cite{Liu2018CVPR} evaluates a non-linear transform that is equivalent to our \bcos{} operator with B=2. 
{In contrast to \cite{Liu2018CVPR}, we explicitly introduce this non-linear transform to increase interpretability and show that such models can be scaled to large-scale classification problems.}

%% file: 3.v4-method.tex
In this section, we introduce the \bcos{} transform as a replacement for the linear units in DNNs, which are (almost) ``at the heart of every deep network''~\cite{swish}, and discuss how this can increase the interpretability of DNNs. 

For this, we first introduce the \bcos{} transform as a variation of the linear transform in \cref{subsec:bcos} and highlight its most important properties.
In \cref{subsec:bcos_net}, we show how to construct {\bcos{} networks} and how to {faithfully summarise} the network computations to obtain explanations for their outputs (\ref{subsubsec:explain}).
Then, we discuss how the \bcos{} transform---combined with the binary cross entropy (BCE) loss---affects the parameter optima of the models (\ref{subsubsec:optim}). Specifically, by inducing 
alignment pressure, the \bcos{} transform aligns the model weights with task-relevant patterns in the input.
{Finally, in \cref{subsec:deep_bcos} we integrate the \bcos{} transform into conventional DNNs by using it as a drop-in replacement for the ubiquitously used linear units.}

\mysub{The \bcos{} transform}
\label{subsec:bcos}
Typically, the individual `neurons' in a DNN compute the dot product between their weights $\vec w$ and an input $\vec x$:\vspace{-.8em}
\begin{align}
\label{eq:lc}
    f(\vec x; \vec w) &= \vec w^T\,\vec x = ||\vec w||\, ||\vec x|| \, c(\vec x, \vec w)\;\text{,}\\
    \text{with}\quad c(\vec x, {\vec w})&= \cos\left(\angle(\vec x, {\vec w})\right) \; {.}
\end{align}
Here, $\angle(\vec x, {\vec w})$ returns the angle between the vectors $\vec x$ and ${\vec w}$.
In this work, we seek to improve the interpretability of DNNs by promoting weight-input alignment during optimisation. To achieve this, we propose the 
\textbf{\bcos{} transform}:
\vspace{-.9em}
\begin{align} 
    \label{eq:bcos}
    \hspace{-.07em}\mbcos (\vec x; \vec w) 
    &\hspace{.075em}\myeq\underbrace{||\widehat{\vec w} ||}_{\color[RGB]{10, 132, 180}=1}\hspace{.05em}||\vec x ||\hspace{.05em} {\color[RGB]{10, 132, 180}|}
    c(\vec x, \widehat{\vec w}){\color[RGB]{10, 132, 180}|^{\text{B}} \times \,\text{sgn}\left(c(\vec x, \widehat{\vec w})\right)}\,.
 \vspace{-.45em}\end{align}
 Here, B is a hyperparameter, the hat-operator scales $\widehat{\vec w}$ to unit norm, and sgn denotes the sign function. Note that this only introduces {\emph{minor changes}} %\footnote{
(highlighted in blue) with respect to~\cref{eq:lc}; e.g., note that for B$\,=\,$%
$1$, the \bcos{} transform is equivalent to a linear transform with $\widehat{\vec w}$.
 However, albeit small, these changes are important for {three reasons}. 
 
 \myparagraph{{First}}, they bound the output of \bcos{} neurons, i.e., 
 \begin{align} 
\label{eq:bound}
    ||\widehat{\vec w}||=1\; \Rightarrow \;\mbcos{}(\vec x; \vec w) \leq ||\vec x|| \;.
\end{align}
As becomes clear from \cref{eq:bcos}, equality in \cref{eq:bound} is only achieved if $\vec x$ and $\vec w$ are collinear, i.e., \emph{aligned}. 

\myparagraph{{Secondly}}, by increasing the exponent B, the output for badly aligned weights can be further suppressed, 
{\begin{align}
\label{eq:supp}
    \text{B}\gg1 \land
    |c(\vec x, \widehat{\vec w})|<1 \,\Rightarrow\, \mbcos(\vec x; {\vec w}) \ll ||\vec x||\;,
\end{align}}
and the respective \bcos{} unit can only produce outputs close to its maximum (i.e., $||\vec x||$) 
for a small range of angular deviations from $\vec x$.
In combination, these two properties can significantly alter the optima of the weight vectors $\vec w$.
To illustrate this, we show in \cref{fig:toy_example} how increasing B affects a simple linear classification problem. In particular, \cref{eq:bound,,eq:supp} shift the optimum of the optimisation problem such that for large B the optimal weights align with the red data cluster, independent of the other class. In contrast to the \emph{discriminative explanation} of a linear classifier, which is highly task-dependent (see, e.g., first row in \cref{fig:toy_example})
the \bcos{} transform allows for a \emph{similarity-based explanation}: a sample is confidently classified as the red class if it is aligned well with the corresponding weight vector. 

{
\myparagraph{{Lastly}}, these changes maintain an important property of the linear transform: similar to sequences of linear transforms, sequences of \bcos{} transforms can still be faithfully summarised by a single linear transform (\cref{eq:collapse}). 
Given the bound (\cref{eq:bound}) and the suppression of outputs for badly aligned weights (\cref{eq:supp}) these linear transforms will align with discriminative patterns when optimising a \bcos{} network for classification, see \cref{subsubsec:optim}. As a result, these transforms are well suited to explain the model outputs.}

\mysub{Simple (convolutional) \bcos{} networks}
\label{subsec:bcos_net}
\input{resources/figures/toy_example_v3}

In this section, we first discuss how to construct simple (convolutional) DNNs based on the \bcos{} transform. Then, we show how to summarise the network outputs by a single linear transform and, finally, why this transform aligns with discriminative input patterns in classification tasks.

\myparagraph{\bcos{} networks.} The \bcos{} transform is designed as a drop-in replacement of the linear transform, i.e., it can be used in exactly the same way. For example, first consider a \emph{conventional} fully connected multi-layer neural network
$\vec f(\vec x; \theta)$ of $L$ layers, represented by
\begin{align} 
\label{eq:relu_net}
    \vec f(\vec x; \theta) = \vec l_L   \circ\vec l_{L-1}   \circ...   \circ\vec l_{2}   \circ\vec l_1 (\vec x)\; \text{,}
\end{align}
with $\vec l_j$ denoting layer $j$ with parameters $\vec w^k_j$ for neuron $k$ in layer $j$, and $\theta$ the collection of all model parameters. In such a model, each layer $\vec l_j$ typically computes 
\begin{align} 
    \vec l_j(\vec a_j; \mat w_j) = \phi\left(\mat w_j \,\vec a_j\right)\;,
\end{align}
with $\vec a_j$ the input to layer $j$, $\phi$ a non-linear activation function (e.g., ReLU), and the row $k$ of $\mat w_j$ given by the weight vector $\vec w_j^k$ of the $k$-th neuron in that layer.
Note that the non-linear activation function $\phi$ is \emph{required} to be able to model non-linear relationships with multiple layers in sequence. \\
A corresponding \bcos{} network $\vec f^*$ with layers $\vec l^*_j$ can be formulated in exactly the same way as
\vspace{-.225em}\begin{align} 
\label{eq:bcos_net}
    \vec f^*(\vec x; \theta) =\vec l^*_L   \circ\vec l^*_{L-1}   \circ...   \circ\vec l^*_{2}   \circ\vec l^*_1 (\vec x)\,,
 \vspace{-2em}\end{align}
with the only difference being that every dot product (here between rows of $\mat w_j$ and the input $\vec a_j$) is replaced by the \bcos{} transform in \cref{eq:bcos}.
In matrix form, this equates 
to
\vspace{-.75em}
\begin{align} 
    \vec l^*_j(\vec a_j; \mat w_j) = 
         |c(\vec a_j; \widehat{\mat w}_j)|^{\text{B}-1}\times\left(\widehat{\mat w}_j \,\vec a_j \right)\label{eq:bcos_suppress}
        \; .
 \vspace{-3em}\end{align}
Here, the power, absolute value, and $\times$ operators are applied element-wise, $c(\vec a_j; \widehat{\mat w}_j)$ computes the cosine similarity between input $\vec a_j$ and the rows of $\widehat{\mat w}_j$, and the hat operator scales the rows of $\widehat{\mat w}_j$ to unit norm.
{To see the equivalence of \cref{eq:bcos_suppress,,eq:bcos}, note that $\widehat{\mat w}_j \,\vec a_j$ computes the scalar product between each row of $\widehat{\mat w}_j$ and $\vec a_j$, which includes a cosine factor. We account for this by reducing the exponent to B--$1$ in \cref{eq:bcos_suppress}; for a derivation, see supplement (Sec.~D).}
Finally, note that for B$>$1 the layer transform $\vec l_j^*$ is \emph{non-linear}. As a result, {a non-linearity $\phi$ is not required for a \bcos{} network} to model non-linear relationships.

The above discussion readily generalises to 
convolutional neural networks (CNNs): in CNNs, we replace the linear transforms computed by the convolutional kernels by \bcos{}, {see Alg.~1 in supplement}. Further, although we assumed a plain multi-layer network without add-ons such as skip connections, we show in \cref{sec:results} that the benefits of \bcos{} also transfer to more advanced architectures (\cref{subsec:deep_bcos}).

\mysubsub{Computing explanations for \bcos{} networks}
\label{subsubsec:explain}
\input{3.v3.3-explanations}

\mysubsub{Optimising \bcos{} networks for classification}
\label{subsubsec:optim}
\input{3.v3.3-optimising}

\mysubsub{MaxOut to increase modelling capacity}
\label{subsubsec:maxout}
As discussed in \cref{subsec:bcos_net}, a deep \bcos{} network  
with B$>$1 does not \emph{require} a non-linearity between subsequent layers to model non-linear relationships. This, of course, does not mean that it could not \emph{benefit} from it. While there are many potential non-linearities to choose from, in this work, we specifically explore the option of combining the \bcos{} transform with the MaxOut~\cite{goodfellow2013maxout} operation. In particular, we model every neuron in a \bcos{} network by 2 \bcos{} transforms\footnote{\mbox{Initial experiments showed no added benefit when using more than 2 units.}} of which the maximal activation is forwarded:
\begin{align} 
    \text{MaxOut} (\vec x) = \textstyle \max_{i\in\{1, 2\}} \left\{\mbcos{}(\vec x; \vec w_i)\right\} \;.
\end{align}
We do so for two reasons. First, in order to forward a large signal, one such MaxOut unit still needs to have at least one weight vector that highly aligns with a given input and the alignment pressure is thus maintained during optimisation. Secondly, while the latter is also true for the ReLU~\cite{relu} operation, we noticed that networks with the MaxOut operation were much easier to optimise. This could be due to the `dying neuron' problem, cf.~\cite{goodfellow2013maxout}, and could potentially be remedied by better initialisation schemes.

\mysub{Advanced \bcos{} networks}
\label{subsec:deep_bcos}
To test the generality of our approach, we evaluate how integrating the \bcos{} transform into commonly used DNN architectures affects their classification performance and interpretability.
In order to `convert' such models to \bcos{} networks we proceed as follows. First, every convolutional kernel / fully connected layer is replaced by the corresponding \bcos{} version with two MaxOut units (see \cref{subsubsec:maxout}). Secondly, any other non-linearities (e.g., ReLU, MaxPool, etc.), as well as any batch norm layers are removed to maintain the alignment pressure and to ensure that the model can be summarised via a single linear transform.

%% file: resources/figures/toy_example_v3.tex
\begin{figure}
    \centering
    \begin{subfigure}[b]{\linewidth}
    \includegraphics[width=\linewidth, trim=1em 0 0 0, clip]{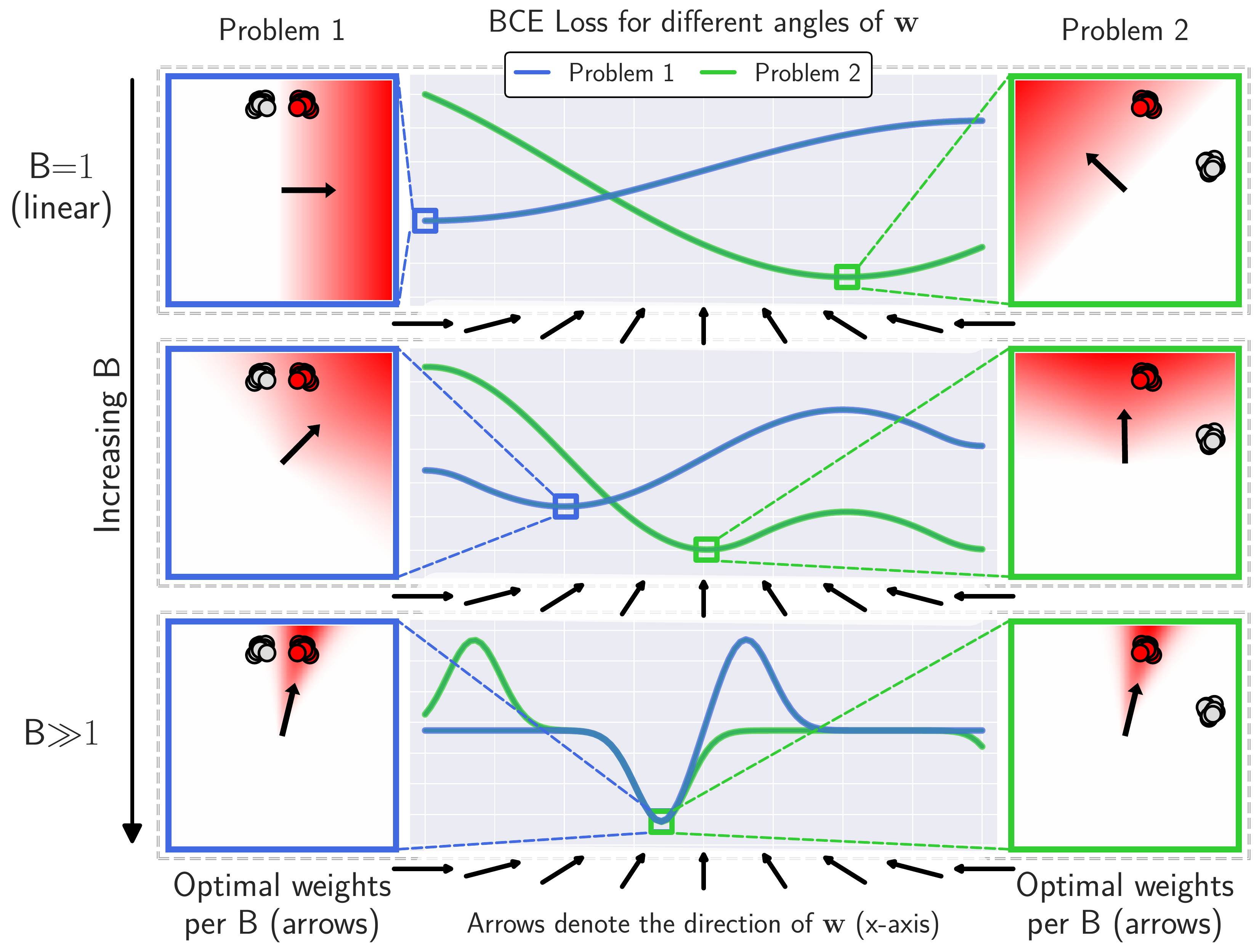}
    \end{subfigure}
    % \vspace{-1em}
    \caption{
    {
    \textbf{Col.~2:} BCE loss for different angles of $\vec w$ for \bcos{} classifiers (\cref{eq:bcos}) with different values of B (rows) for two classification problems. \textbf{Cols.~1+3:} Visualisation of the classification problems and the corresponding optimal weights (arrows) per B.     For B=1 (first row) the weights $\vec w$ represent the decision boundary of a linear classifier. 
    Although the red cluster is the same in both cases, the optimal weight vectors differ significantly (compare within row). In contrast, for higher values of B the weights converge to the same optimum in both tasks (see last row).} 
    }
    \label{fig:toy_example}
    % \vspace{-.5em}
\end{figure}

%% file: 3.v3.3-explanations.tex
{As can be seen by rewriting \cref{eq:bcos_suppress}, a \bcos{} layer effectively computes an input-dependent linear transform:
\begin{align} 
\label{eq:dynlin}
    \vec l^*_j(\vec a_j; \mat w_j) &\;=\; 
        \widetilde{\mat w}_j(\vec a_j) \,\vec a_j \, ,\\[.25em]
        \text{with} \quad {\widetilde{\mat w}_j(\vec a_j)}&\;=\;{|c(\vec a_j; \widehat{\mat w}_j)|^{\text{B}-1}\odot \widehat{\mat w}_j} \; \text{.}
        \label{eq:dynlin2}
 \end{align}
Here, $\odot$ scales the rows of the matrix to its right by the scalar entries of the vector to its left.
Hence, the output of a \bcos{} network, see \cref{eq:bcos_net}, is effectively calculated as
\begin{align}
\label{eq:linear_sequence}
    \vec f^*(\vec x; \vec \theta) = \widetilde{\mat w}_L(\vec a_L)\widetilde{\mat w}_{L-1}(\vec a_{L-1})...\widetilde{\mat w}_1(\vec a_1\myeq\vec x)\vec x\;.
\end{align}
As multiple linear transforms in sequence can be collapsed to a single one,
the output $\vec f^*(\vec x; \vec\theta)$ can be written as
\begin{align} 
\label{eq:collapse}
    \vec f^*(\vec x; \vec \theta) &= \mat w_{1\rightarrow L}(\vec x)\,\vec x\; \text{,} \\
    \text{with} \quad
        \mat{W}_{{1}\rightarrow {L}}\left(\vec{x}\right) &= \textstyle\prod_{j={1}}^{L} \widetilde{\mat{W}}_j \left(\vec{a}_{j}\right)\; \text{.}
 \end{align}
Thus, $\mat w_{1\rightarrow L}(\vec x)$ {faithfully summarises} the network computations (\cref{eq:bcos_net}) by a single linear transform (\cref{eq:collapse}).

To explain an activation (e.g., the class logit), we can now either {directly visualise} the corresponding row in $\mat w_{1\rightarrow L}$, see \cref{fig:global_protos,,fig:intermediate_protos}, or the \emph{contributions} according to $\mat w_{1\rightarrow L}$ coming from individual input dimensions. 
We use the resulting spatial \textbf{contributions maps} to quantitatively evaluate the explanations. In detail, the input contributions  $\vec s_j^l(\vec x)$ to neuron $j$ in layer $l$ for an input $\vec x$ are given by 
\begin{align}
    \label{eq:contrib}
        \vec{s}_{j}^l(\vec x) = \left[\mat W_{1\rightarrow l} (\vec{x})\right]_j^T \odot \vec x\; \text{,}
    \end{align}
with $[\mat w_{1\rightarrow l}]_j$ denoting the $j$th row in matrix $\mat w_{1\rightarrow l}$; as such, the contribution from a single pixel location ${(x, y)}$ is given by $\sum_c[\vec s_j^l(\vec x)]_{(x, y, c)}$ with $c$ the color channels.
}

%% file: 3.v3.3-optimising.tex
In the following, we discuss why the linear transforms $\mat W_{1\rightarrow L}$ (see \cref{eq:collapse}) can be expected to be interpretable, i.e., to align with relevant input patterns.

{For this, first note that the output of each neuron---and thus of each layer---is bounded, 
cf.~\cref{eq:bound,,eq:bcos_suppress}. Since the output of a \bcos{} network is computed as a sequence of such bounded transforms, see \cref{eq:linear_sequence}, the output of the network as a whole is also bounded. 
Secondly, note that a \bcos{} network as a whole can only achieve its upper bound for a given input if the units in each layer achieve their upper bound. 
Importantly, as discussed in \cref{subsec:bcos} (\cref{eq:bound}), the individual units, in turn, can only achieve their maxima by aligning with their inputs. Hence, optimising a \bcos{} network to maximise its output over a set of inputs will optimise the model weights to align with those inputs.}

In order to take advantage of this when optimising for classification, we train the \bcos{} networks with the binary cross entropy (BCE) loss
\begin{align} 
    \mathcal{L}(\vec x_i, \vec y_i) = \text{BCE}\left(\sigma(\vec f^*(\vec x_i; \vec \theta) + \vec b), \vec y_i\right)\; {,}
\end{align}
for input $\vec x_i$ and its corresponding one-hot encoded class label $\vec y_i$. Here, $\sigma$ denotes the sigmoid function, $\vec b$ a bias, and $\vec \theta$ the model parameters. 
In particular, we choose the BCE loss because it directly entails output maximisation. Specifically, in order to reduce the BCE loss, the network is optimised to maximise the (negative) class logit for the correct (incorrect) classes. As discussed in the previous paragraph, this will optimise the weights in each layer of the network to align with their inputs. In particular, they will need to align with class-specific input patterns such that these result in large outputs for the respective class logits.
 
{Finally, note that increasing B allows to specifically reduce the output of badly aligned weights in each layer (cf.~\cref{eq:bound}). This will decrease the layer's output strength and thus the output of the network as a whole for badly aligned weights, which increases the alignment pressure during optimisation (thus, higher B$\rightarrow$higher alignment).}

%% file: 4.0-experiments.tex
\myparagraph[0]{Datasets.}
We evaluate the accuracies of several \bcos{} networks on the CIFAR-10~\cite{krizhevsky2009cifar10} and the ImageNet~\cite{deng2009imagenet} datasets. We use the same datasets for the qualitative and quantitative evaluations of the model-inherent explanations.

\myparagraph{Models.}
For the CIFAR10 experiments, we develop a simple fully-convolutional \bcos{} DNN, consisting of 9 convolutional layers, each with a kernel size of 3, followed by a global pooling operation. We evaluate a network without additional non-linearities as well as with MaxOut units, see \cref{subsubsec:maxout}. For the ImageNet experiments, we rely on the publicly available~\cite{pytorch} implementations of the VGG-11~\cite{simonyan2015vgg}, ResNet-34~\cite{he2016deep}, InceptionNet (v3)~\cite{szegedy2015inception}, and DenseNet-121~\cite{huang2017densely} model architectures. We adapt those architectures to \bcos{} networks as described in \cref{subsec:deep_bcos}. For details on the training procedure, see supplement (Sec.~C).

\myparagraph{Image encoding.} We add three additional channels and encode images
as \mbox{[$r$, $g$, $b$, $1$%
$-$%
$r$, $1$%
$-$%
$g$, $1$%
$-$%
$b$]}, with $r, g, b\,$%
$\in\,$%
$[0, 1]$ the red, green, and blue color channels. On the one hand, this reduces a bias towards bright regions in the image\footnote{The network is trained to maximise its output, which is bounded by the input norm. In the conventional encoding, however, black pixels, e.g., have a norm of zero and thus cannot contribute to the class logits.} \cite{Boehle2021CVPR}. On the other hand, colors with the \emph{same angle in the original encoding}---i.e., $[r_1, g_1, b_1]\propto[r_2, g_2, b_2]$---\emph{are unambiguously encoded by their angles under the new encoding}. Therefore, the linear transformation $\mat w_{1\rightarrow l}$ can be decoded into colors just based on the angles of each pixel, see \cref{fig:global_protos}.
For a detailed discussion, see supplement (Sec.~D).

\myparagraph{Evaluating explanations.} To compare explanations for the model decisions and {evaluate their faithfulness}, we employ the \emph{grid pointing game} \cite{Boehle2021CVPR}. That means we evaluate the trained models on a synthetic 3x3 grid of images of different classes and for each of the corresponding class logits measure how much positive attribution an explanation method assigns to the correct location in the grid; for a visualisation of a 2x2 grid, see \cref{fig:local_quali}.
Following \cite{Boehle2021CVPR}, we construct 500 image grids from the most confidently and correctly classified images. 
We compare the {model-inherent} contribution maps, see \cref{eq:contrib}, against several commonly employed {post-hoc} explanation methods under two settings. First, we evaluate all methods on the \bcos{} networks to investigate which method provides the best explanation \emph{for the same model}. Secondly, we further evaluate the post-hoc methods on pre-trained versions of the original models (VGG, ResNet, DenseNet, InceptionNet). 
This allows to compare explanations \emph{between different models} and  to assess the \emph{explainability gain} obtained by converting conventional models to \bcos{} networks. Lastly, all non-perturbation-based attribution maps are smoothed by a 15$\times$15 (3$\times$3) kernel to better account for negative attributions in the localisation metric for ImageNet (CIFAR-10) images, which is negligible with respect to the overall image size.
\input{resources/figures/localisation_qualitative}

\myparagraph{Visualisations details.} 
{For generating the visualisations of the linear transforms for individual neurons $n$ in layer $l$ (cf.~\cref{fig:global_protos,,fig:intermediate_protos}), we proceed as follows. First, we select all pixel locations  $(x, y)$ that positively contribute to the respective activation (e.g., class logit) as computed by \cref{eq:contrib};
i.e., $\{(x, y)\; \text{s.t.}\, \sum_c\, [\vec s_n^l(\vec x)]_{(x, y, c)}\mygreater0\}$ with $c$ the 6 color channels (see image encoding). Then, we normalise the weights of each color channel such that the corresponding weights (e.g., for $r$ and $1\myminus r$) sum to 1.
Note that this normalisation maintains the angle for each color channel pair (i.e., $r$ and $1\myminus r$), but produces values in the allowed range $r,g,b\,$%
$\in\,$%
$[0,1]$. These normalised weights can then directly be visualised as color images. The opacity of a pixel is set to $\min(||\vec w_{(x, y)}||_2/p_{99.5}, 1)$, with $p_{99.5}$ the $99.5$th percentile over the weight norms $||\vec w_{(x, y)}||_2$ across all $(x, y)$.}

%% file: resources/figures/localisation_qualitative.tex
\begin{figure}
    \centering
    \begin{subfigure}[b]{\linewidth}
    \includegraphics[width=\linewidth, trim=1em 1em .5em 0, clip]{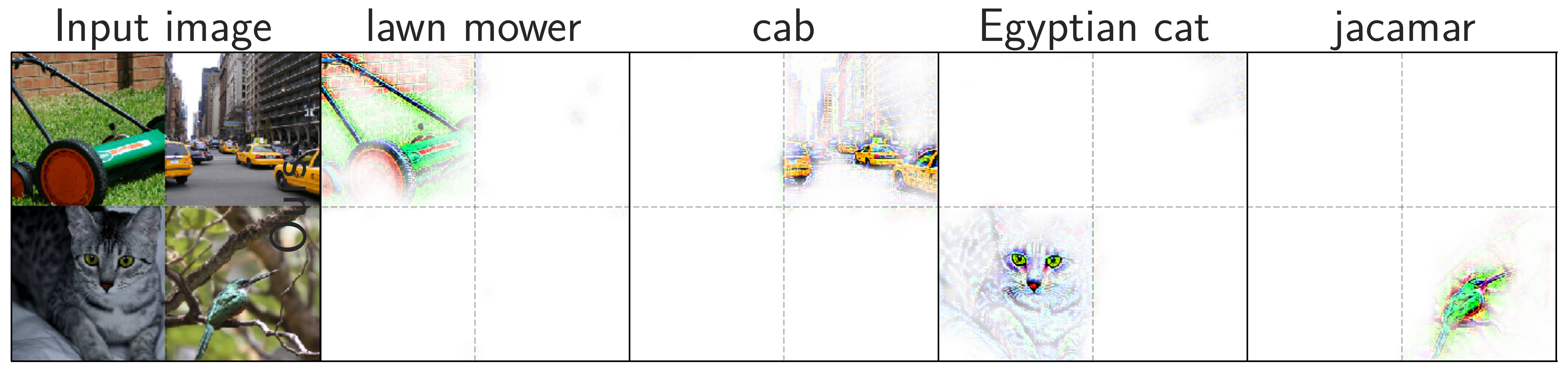}
    \end{subfigure}
    \caption{2$\times$2 example for the pointing game. \textbf{Column 1}: input image. \textbf{Columns 2 -- 5}: explanations for individual class logits.}
    \label{fig:local_quali}
\end{figure}

%% file: 5.0-results.tex
In this section, we analyse the performance and interpretability of \bcos{} networks. For this, in \cref{subsec:plain_results} we show results of  `simple' \bcos{} networks without advanced architectural elements such as skip connections or inception modules. In this context, we investigate how the B parameter influences \bcos{} networks in terms of performance and interpretability.
Thereafter, in \cref{subsec:advanced_results}, we present quantitative results of the \emph{advanced} 
\bcos{} networks, i.e., \bcos{} networks based on common DNN architectures (\cref{subsec:deep_bcos}). Finally, in \cref{subsubsec:quali_results}, we present and qualitatively discuss explanations for outputs of individual neurons.

\subsection{Simple \bcos{} networks}
\label{subsec:plain_results}
In the following, we discuss the experimental results of simple \bcos{} networks evaluated on the CIFAR-10 dataset. 
\input{resources/figures/b-ablation}
\input{resources/figures/local_c10}
\input{resources/tables/c10}

\myparagraph{Accuracy.}
In \cref{tbl:c10_table}, 
we present the test accuracies of various \bcos{} networks trained on CIFAR-10. We show that a {plain \bcos{} network} (B$\myeq2$) 
{without any add-ons} (ReLU, batch norm, etc.) can achieve competitive\footnote{A ResNet-20 achieves 91.2\%~\cite{he2016deep} under the same data augmentation.} performance. 
By modelling each neuron via 2 MaxOut units (\cref{subsubsec:maxout}), the performance can be 
increased and the resulting model (B$\myeq2$) performs on par with a ResNet-56 (achieving 93.0\%, see \cite{he2016deep}). Further, we see that an increase in the parameter B leads to a decline in performance from $93.8\%$ for 
B$\myeq1.25$ to $92.4\%$  for 
B$\myeq2.5$. Notably, despite its simple design, our strongest model with B$\myeq1.25$ performs similarly to the strongest ResNet model ($93.6\%$) reported in \cite{he2016deep}.

\myparagraph{Model interpretability.}
As discussed in \cref{subsubsec:optim}, we expect an increase in B to increase the alignment pressure on the weights during optimisation and thus influence the models' optima, similar to the single unit case in \cref{fig:toy_example}. 
This is indeed what we observe. For example, in \cref{fig:b-abl-quali}, we visualise $[\mat W_{1\rightarrow l}(\vec x_i)]_{y_i}$ (see \cref{eq:collapse}) for different samples $i$ from the CIFAR-10 test set. For higher values of B, the weight alignment increases notably from piece-wise linear models (B=1) to \bcos{} networks with higher B (B=2.5). Importantly,  this does not only lead to an increase in the {visual quality} of the explanations, but also to quantifiable gains in model interpretability. In particular, as we show in \cref{fig:local_c10}, the spatial contribution maps defined by $\mat W_{1\rightarrow l}(\vec x_i)$ (see \cref{eq:contrib}) of models with larger B values score significantly higher in the localisation metric (see \cref{sec:experiments}).
\subsection{Advanced \bcos{} networks}
\label{subsec:advanced_results}
In this section, we first quantitatively evaluate the performance and interpretability of the advanced \bcos{} networks, see \cref{subsec:deep_bcos,,sec:experiments}. Then, we qualitatively investigate the interpretability of the \bcos{} networks 
in more detail.
\input{resources/figures/localisation_results}
\input{resources/tables/imnet}

\myparagraph{Classification accuracies}
 of the pretrained~\cite{torchvision} %
models and their corresponding \bcos{} counterparts (trained from scratch) are presented in \cref{tbl:imnet_table}. 
The \bcos{} networks (except bottom row in \cref{tbl:imnet_table}) were trained for 100 epochs with the Adam optimiser, a learning rate of $2.5e^{-4}$, a batch size of 256 and no weight decay. The learning rate was decreased by a factor of 10 after 60 epochs and we used RandAugment for data augmentation; for further details on training and evaluation, see supplement (Sec.~C).
We would like to highlight that these results  are thus obtained `out of the box', i.e., with a simple and commonly used optimisation scheme for all models. Thus, in spite of the drastic changes to the model architectures (no batch norm, no ReLU, no MaxPool), we are able to achieve competitive results: {our B-cos VGG-11 outperforms its conventional counterpart and we only observe minor drops in accuracy w.r.t.~the baseline models for the other networks, e.g., $1.1$ p.p.~for the DenseNet-121 (74.4\% vs. 73.3\%). By training a DenseNet-121 model for 200 epochs, a batch size of 128, learning rate warm-up and a cosine learning rate schedule, we are able to close the gap between the pretrained DenseNet-121 model and its \bcos{} counterpart (see last row in \cref{tbl:imnet_table}, `training$^+$').} 

\myparagraph{Model interpretability.} In \cref{fig:localisation_results}, we present the explanation quality results as assessed by the localisation metric for various post-hoc attribution methods as well as the model-inherent contribution maps (\cref{eq:contrib}). We evaluated the post-hoc methods both on the conventional pretrained models (cf.~\cref{tbl:imnet_table}) as well as on their corresponding \bcos{} counterparts; in \cref{fig:localisation_results}, we show results for two \bcos{} networks, for the remaining results we kindly refer the reader to the supplement (Sec.~B). In particular, we evaluated various gradient-based methods (the `vanilla gradient' (Grad)~\cite{baehrens2010explain}; Input$\times$Gradient (IxG), cf.~\cite{adebayo2018sanity}; Integrated Gradients (IntGrad)~\cite{sundararajan2017axiomatic}; DeepLIFT~\cite{shrikumar2017deeplift}; GradCam (GCam)~\cite{selvaraju2017grad}) and two perturbation-based methods (LIME~\cite{ribeiro2016lime}, RISE~\cite{petsiuk2018rise}) for comparison. We would like to highlight the following two results. {First, for all converted \bcos{} architectures, the model-inherent explanations not only outperform any post-hoc explanation for the models' decisions, but achieve close to optimal scores on the localisation metric.} Secondly, as we show in the supplement (Sec.~B), none of the post-hoc methods that we evaluated for the conventional models provides a better explanation for those models than the linear transform $\mat W_{1\rightarrow l}(\vec x)$ provides for the \bcos{} networks. Hence, by using \bcos{} networks instead of conventional models, it is possible to drastically improve the models' interpretability. For a qualitative comparison between the model-inherent explanations and post-hoc methods, see \cref{fig:quali_comp}.

\input{resources/figures/quali_comp}
\mysubsub{Qualitative evaluation of explanations}
\label{subsubsec:quali_results}
{The following results are based on the DenseNet-121 training$^+$ model, cf.~\cref{tbl:imnet_table}; other advanced \bcos{} networks yield similar results, see supplement (Sec.~A).}

Every activation in a \bcos{} network is the result of a sequence of \bcos{} transforms. Hence, every neuron $n$ in any layer $l$ can be explained via the corresponding linear transform $[\mat W_{1\rightarrow l}(\vec x)]_n$, see \cref{eq:collapse}. 

For example, in \cref{fig:global_protos}, we visualise the linear transforms of the respective \emph{class logits} for various input images.
Given the alignment pressure during optimisation, these linear transforms align with class-discriminant patterns in the input and thus actually resemble the class objects. 

Similarly, in \cref{fig:intermediate_protos,,fig:other_layers}, we visualise explanations for \emph{intermediate neurons}. 
Specifically, in \cref{fig:other_layers}, we show explanations for some of the most highly activating neurons over the validation set. 
We find that neurons in early layers seem to represent low-level concepts (e.g., curves), and become more complex in later layers ($l\myeq87$: hands, $l\myeq120$: streetcars); for additional results, see supplement Sec.~A.
\input{resources/figures/other_layers}
\input{resources/figures/ambiguous}
\input{resources/figures/neurons}

\cref{fig:intermediate_protos} shows additional results for neurons in layer 87. {We observed that some neurons become highly specific to certain concepts, such as wheels (neuron 739), faces (neuron 797), or eyes (neuron 938). These neurons do not just learn to align with simple, fixed patterns---instead, they represent semantic concepts and are robust to changes in colour, size, and pose.} 
{Further, we found that several neurons respond preferentially to watermarks, emphasising the importance of explainability for debugging DNNs: while watermarks do not seem \emph{semantically} meaningful, they can represent an informative feature for classification if they are only present in a subset of classes, see supplement (Sec.~B).}

Lastly, 
in \cref{fig:ambiguous_classifications}, we show explanations of the two most likely classes for images for which the model produces predictions with high uncertainty; additionally, we show the $\Delta$-Explanation, i.e., the difference in contribution maps for the two classes, see \cref{eq:contrib}. By means of the model-inherent linear mappings $\mat w_{1\rightarrow L}$, the model can provide a human-interpretable explanation for its uncertainty: there are indeed features in each of those images that provide evidence for both of the predicted classes.
\footnotetext{We manually evaluated the first 100 images for each neuron and found the respective neurons to reliably highlight the assigned concept, i.e., their linear explanations $[\mat w_{1\rightarrow 87}(\vec x)]_n$ are similar to those shown in \cref{fig:intermediate_protos}.}

\mysubsub{Limitations}
\label{subsubsec:limitations}
{By normalising the weights and computing the additional down-scaling factor (see \cref{eq:bcos}), the \bcos{} transform adds computational overhead, which we observed to increase training and inference time by up to 60\% in comparison to baseline models of the same size. However, we expect this cost to decrease significantly in the future with an optimised implementation of the \bcos{} transform. 

Moreover, in this work we specifically investigated how to integrate the B-cos transform into CNNs for image classification. How to integrate the B-cos transform into other types of architectures, such as (vision) transformers \cite{vaswani2017attention,dosovitskiy2021an}, and how it affects model interpretability on other tasks and domains, remains an open question. Given the increasing dominance of transformers, we believe extending our method to such models to be an important next step.
}

%% file: resources/figures/b-ablation.tex
\begin{figure}
    \centering
    \begin{subfigure}[b]{\linewidth}\centering
    \includegraphics[width=\linewidth]{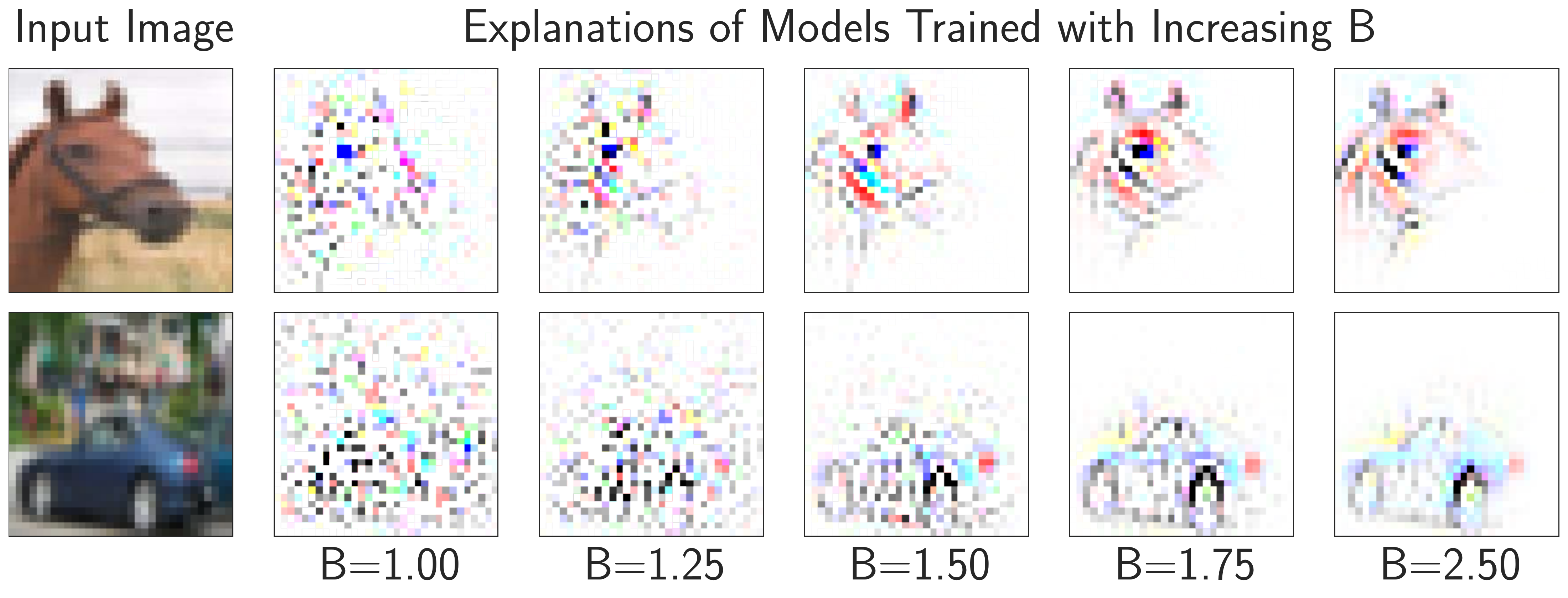}
    \end{subfigure}
    % \vspace{-1.75em}
    \caption{%\hspace{-.22em}
    \textbf{Col. 1}: Input images.
    \textbf{Cols. 2-6}: Explanations for classes `horse' and `car' of models trained with increasing values of B. 
    With higher B, the linear transforms $\mat w_{1\rightarrow l}$ increasingly align with discriminative patterns and thus become more interpretable.
    }
    \label{fig:b-abl-quali}
    % \vspace{-1em}
\end{figure}

%% file: resources/figures/local_c10.tex
\begin{figure}[t]
    \centering
    % \vspace{.5em}
    \begin{subfigure}[b]{1.0\linewidth}
    \includegraphics[width=\textwidth, trim=.5em 1em .5em 2em, clip]{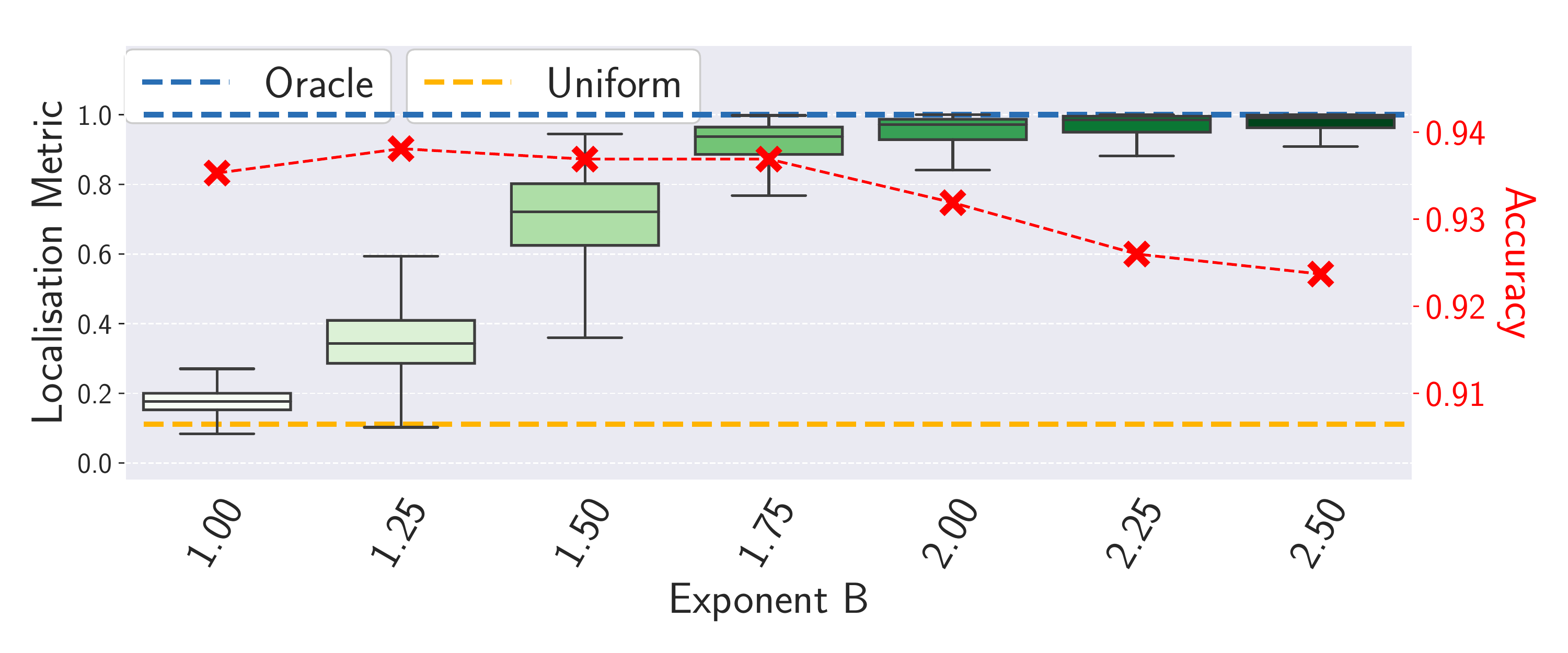}
    \end{subfigure}
    % \vspace{-2em}
    \caption{Accuracy (crosses) and localisation (box plots) results for a \bcos{} network trained with different B values. While decreasing accuracy, a higher B yields significant gains in localisation.     }
    \label{fig:local_c10}
\end{figure}

%% file: resources/tables/c10.tex
\begin{table}[h!]
    \vspace{.25em}
    \centering
    {\setlength{\tabcolsep}{0.25em}\setlength\extrarowheight{-1pt}
    \begin{tabular}{|>{\centering\arraybackslash}  >{\centering\arraybackslash}p{1.7cm} |
    >{\centering\arraybackslash}p{.6cm} |
    >{\centering\arraybackslash}p{.6cm} |
    >{\centering\arraybackslash}p{.6cm} |
    >{\centering\arraybackslash}p{.6cm} |
    >{\centering\arraybackslash}p{.6cm} |
    >{\centering\arraybackslash}p{.6cm} |
    >{\centering\arraybackslash}p{.6cm} |
    >{\centering\arraybackslash}p{.6cm} |
    }
    \cline{2-9}
  \multicolumn{1}{c|}{}&\footnotesize \bf plain&\multicolumn{7}{c|}{\footnotesize \bf MaxOut \bcos{} networks}
    \\\hline
    \footnotesize \textbf{B}
    & {\footnotesize 2.00}
    & {\footnotesize 1.00}
    & {\footnotesize 1.25}
    & {\footnotesize 1.50}
    & {\footnotesize 1.75}
    & {\footnotesize 2.00}
    & {\footnotesize 2.25}
    & {\footnotesize 2.50}
    \\\hline
    \footnotesize \textbf{Accuracy (\%)}
    &\footnotesize 91.5 % Vanilla
    &\footnotesize 93.5 % B=1
    &\footnotesize 93.8 % B=1.25
    &\footnotesize 93.7 % B=1.5
    &\footnotesize 93.7 % B=1.75
    &\footnotesize 93.2 % B=2
    &\footnotesize 92.6 % B=2.25
    &\footnotesize 92.4 % B=2.5
    \\\hline
    \end{tabular}}
    % \vspace{-.5em}
    \caption{\textbf{CIFAR-10.} Model accuracy of a \bcos{} network
    without any additional non-linearity (\textbf{plain}) and for \bcos{} networks with MaxOut (\cref{subsubsec:maxout}) and increasing values for B (left to right).
    }
    \vspace{-.25em}
    \label{tbl:c10_table}
\end{table}

%% file: resources/figures/localisation_results.tex
\begin{figure}
    \centering
    \hspace{-.65em}
    \begin{subfigure}[b]{1.015\linewidth}
    \includegraphics[width=\textwidth, trim=0 1.425em 5.1em 5em, clip]{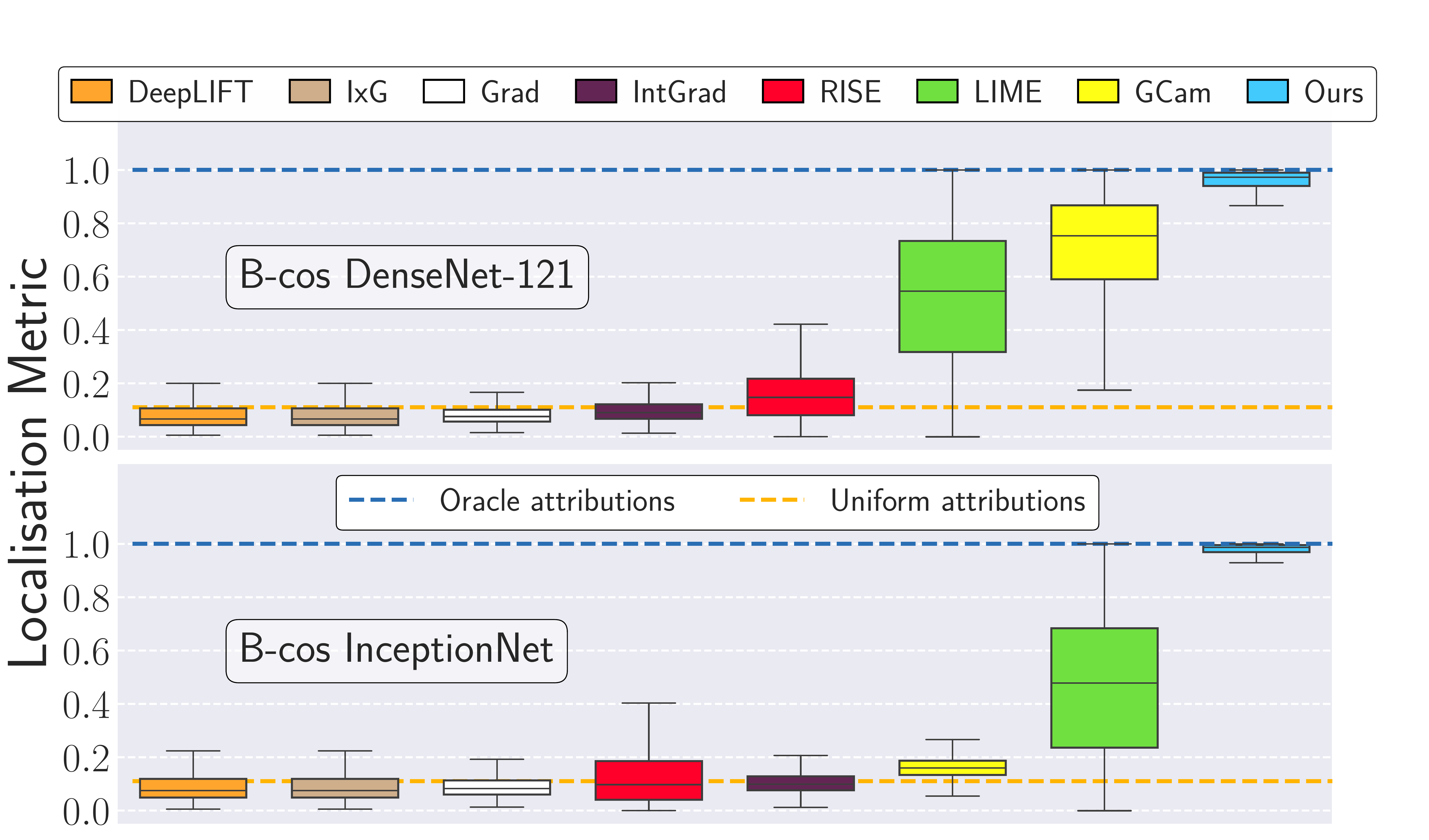}
    \end{subfigure}
    \caption{Localisation results of model-inherent contribution maps (`Ours'), \cref{eq:contrib}, and post-hoc methods.     For more results (VGG-11, ResNet-34, pretrained baselines), see supplement (Sec.~B).
    }
    \label{fig:localisation_results}
\end{figure}

%% file: resources/tables/imnet.tex
\bgroup
\newcolumntype{?}{!{\vrule width 1pt}}
% \vspace{-.5em}
\def\arraystretch{1.1}
\begin{table}[ht]
    \centering
    {\setlength{\tabcolsep}{0.15em}\setlength\extrarowheight{-1pt}
    \begin{tabular}{
    ?>{\centering\arraybackslash}  
    >{\centering\arraybackslash}m{.11\linewidth}| 
    >{\centering\arraybackslash}m{.11\linewidth}| 
    >{\centering\arraybackslash}m{.11\linewidth}| 
    >{\centering\arraybackslash}m{.11\linewidth}| 
    >{\centering\arraybackslash}m{.11\linewidth}| 
    >{\centering\arraybackslash}m{.11\linewidth}| 
    >{\centering\arraybackslash}m{.11\linewidth}| 
    >{\centering\arraybackslash}m{.11\linewidth}?
    }
    % \cline{2-3}
    % \multicolumn{1}{c|}{}&\multicolumn{2}{c|}{\footnotesize \textbf{Top-1 Accuracy (\%)} }
    % \\
    \thickhline
    \multicolumn{2}{?c|}{\footnotesize \textbf{VGG-11}}&
    \multicolumn{2}{c|}{\footnotesize \textbf{ResNet-34}}&
    \multicolumn{2}{c|}{\footnotesize \textbf{DenseNet-121}}&
    \multicolumn{2}{c?}{\footnotesize \textbf{InceptionNet}}\\\hline
    {\footnotesize pre}&
    {\footnotesize \bcos{}}&
    {\footnotesize pre}&
    {\footnotesize \bcos{}}&
    {\footnotesize pre}&
    {\footnotesize \bcos{}}&
    {\footnotesize pre}&
    {\footnotesize \bcos{}}
    \\\hline
    {\footnotesize $69.0$}&
    {\footnotesize $69.6$}&
    {\footnotesize $73.3$}&
    {\footnotesize $71.7$}&
    {\footnotesize $74.4$}&
    {\footnotesize $73.3$}&
    {\footnotesize $77.3$}&
    {\footnotesize $75.4$}\\\hline
    \multicolumn{2}{?c|}{\footnotesize $\Delta\myeq\myplus0.6$}&
    \multicolumn{2}{c|}{\footnotesize $\Delta\myeq\myminus1.6$}&
    \multicolumn{2}{c|}{\footnotesize $\Delta\myeq\myminus1.1$}&
    \multicolumn{2}{c?}{\footnotesize $\Delta\myeq\myminus1.9$}\\\thickhline
    % \noalign{\vrule width .22\linewidth}
    \multicolumn{4}{?c|}{\footnotesize \bf \bcos{} DenseNet-121 training$^{+}$}& \multicolumn{2}{c|}{\footnotesize $74.4$ ($\Delta\myeq0.0$)}
    &\multicolumn{2}{c?}{}\\\thickhline
    \end{tabular}}
    % \vspace{-.5em}
    \caption{\textbf{ImageNet.} Top-1 accuracy (\%)   for various conventional pre-trained models (pre) and their respective \bcos{} version. 
    \newline
    \textbf{Bottom row}: Top-1 accuracy of a \bcos{} DenseNet-121 trained for more epochs and a cosine learning rate schedule (training$^+$).     }
    \label{tbl:imnet_table}
    \vspace{-.75em}
% \end{table}
\end{table}
\egroup

%% file: resources/figures/quali_comp.tex
\begin{figure}[t!]
    \centering
    \begin{subfigure}[b]{\linewidth}
    \includegraphics[width=\linewidth, trim=.25em 1em .5em 0, clip]{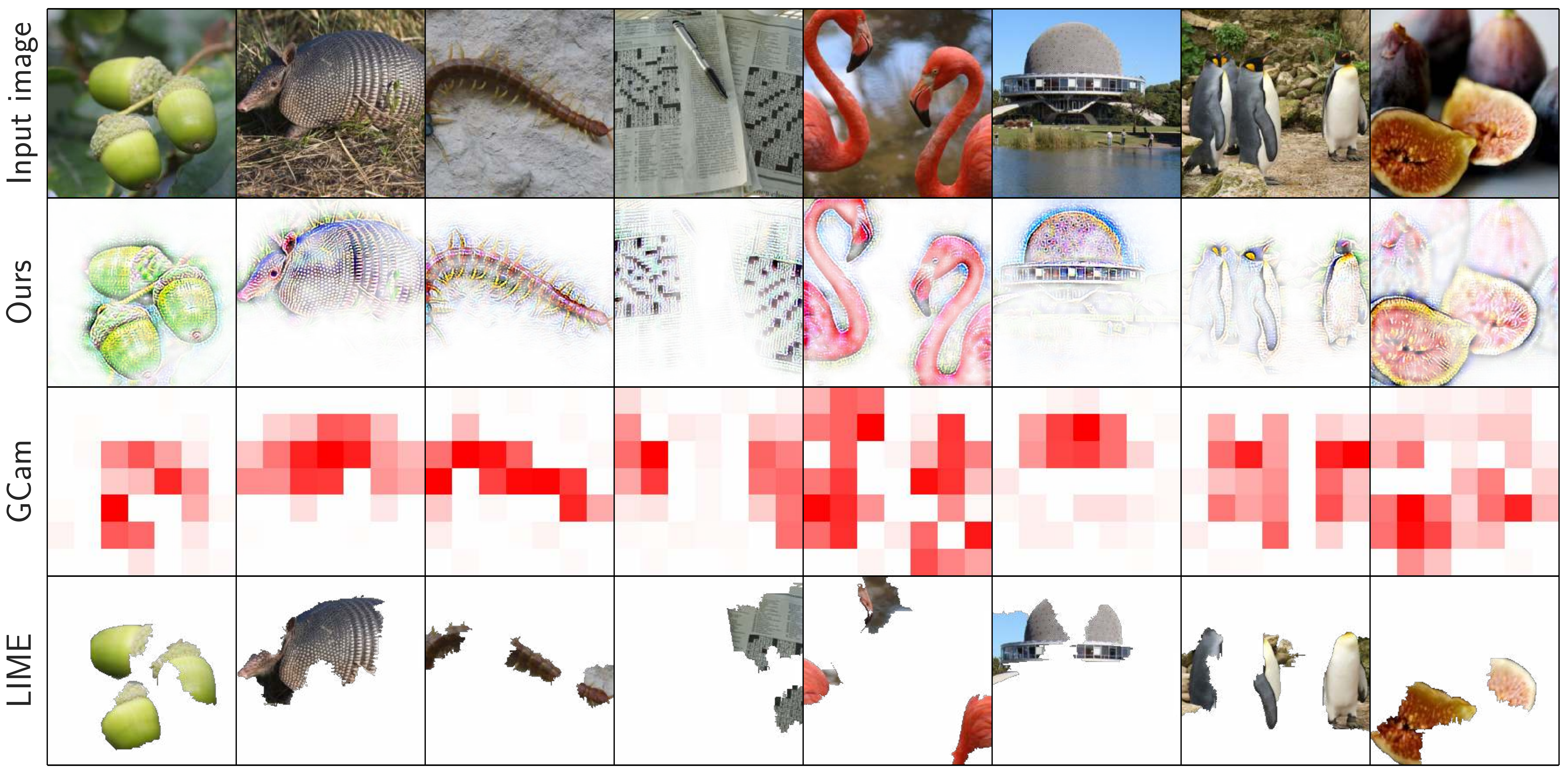}
    \end{subfigure}
    \caption{Comparison between model-inherent explanations and the strongest post-hoc methods. More results in supplement (Sec.~A.)}
    \label{fig:quali_comp}
\end{figure}

%% file: resources/figures/other_layers.tex
\begin{figure}[t!]
    \centering
    \vspace{-.2em}
    \hspace{-.6em}
    \begin{subfigure}[b]{.975\linewidth}
    \begin{subfigure}[b]{\linewidth}
    \includegraphics[width=\textwidth, ]{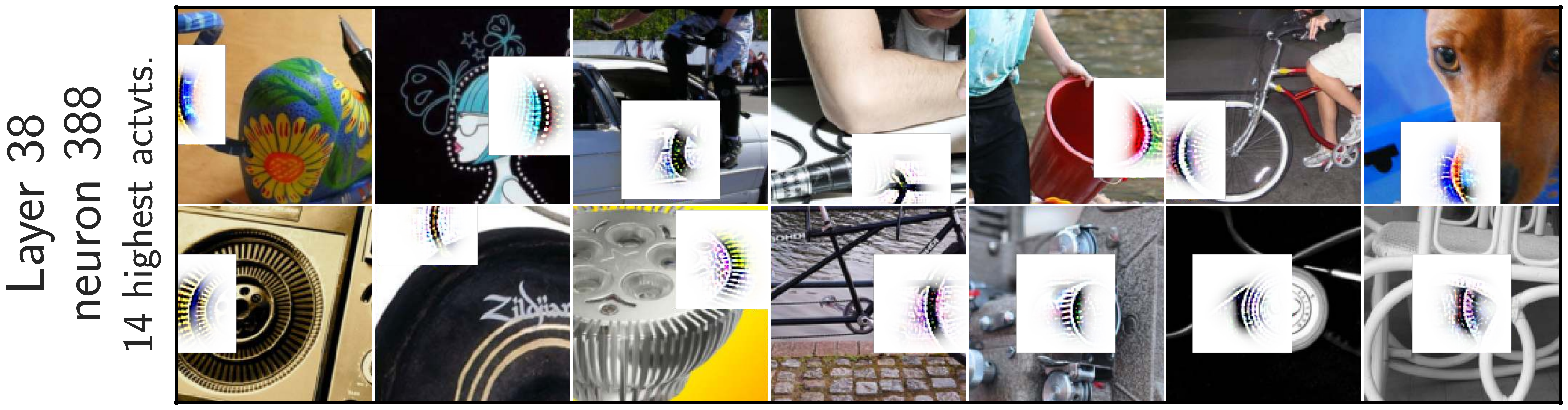}
    \end{subfigure}
    \begin{subfigure}[b]{\linewidth}
    \includegraphics[width=\textwidth, ]{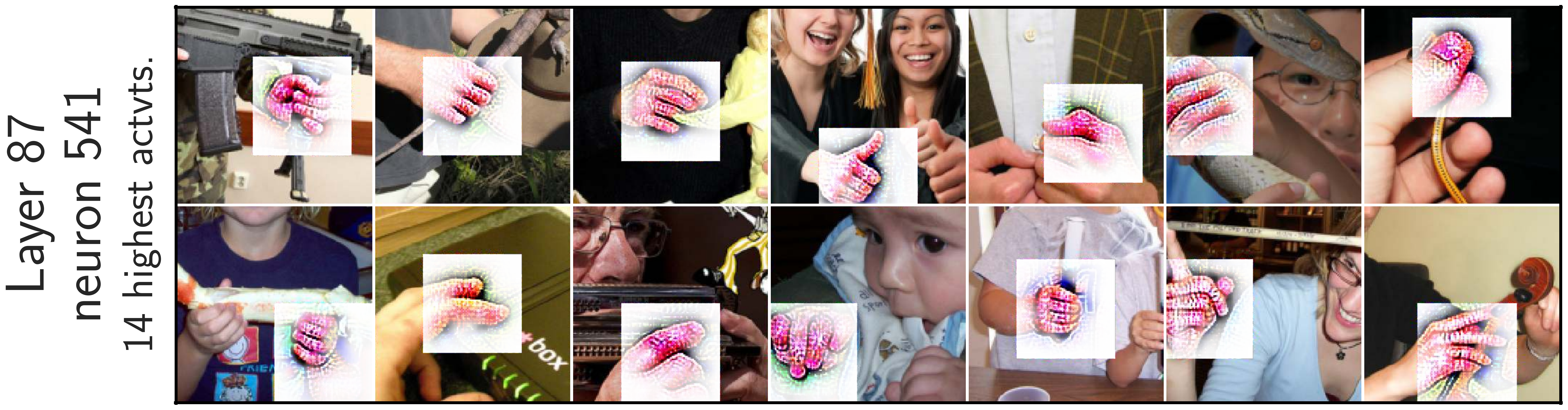}
    \end{subfigure}
    \begin{subfigure}[b]{\linewidth}
    \includegraphics[width=\textwidth, trim=0 .5em 0 0, clip]{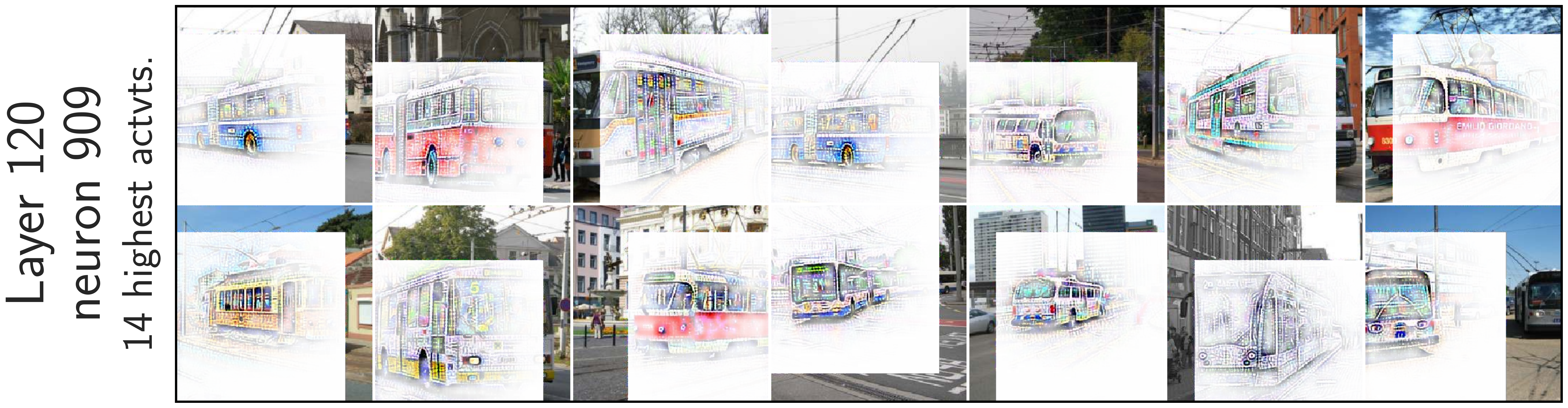}
    \end{subfigure}
    \end{subfigure}
    \caption{Explanations for high activations of neurons from various layers. In early layers the neurons seem to encode low-level concepts (e.g., curves, see layer 38) and represent more high-level concepts in later layers (e.g., layers 87 and 120), see also \cref{fig:intermediate_protos}.}
    \label{fig:other_layers}
    \vspace{-.2em}
\end{figure}

%% file: resources/figures/ambiguous.tex
\begin{figure}[b!]
    \centering
    \begin{subfigure}[b]{.9\linewidth}
    \begin{subfigure}[b]{\linewidth}
    \includegraphics[width=\linewidth, ]{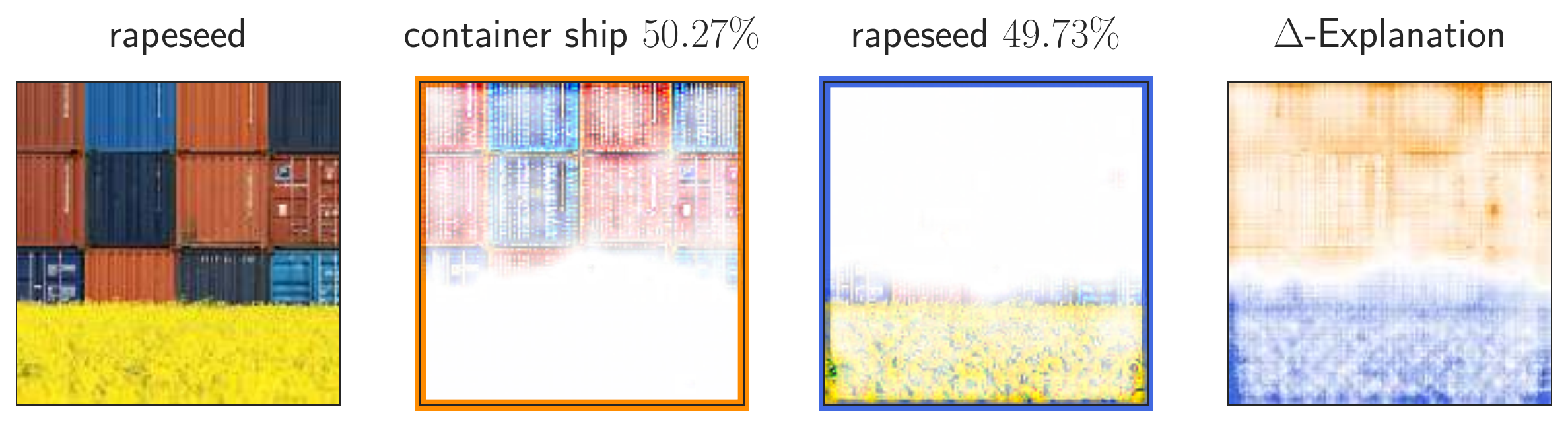}
    \end{subfigure}
    \begin{subfigure}[b]{\linewidth}
    \includegraphics[width=\linewidth, ]{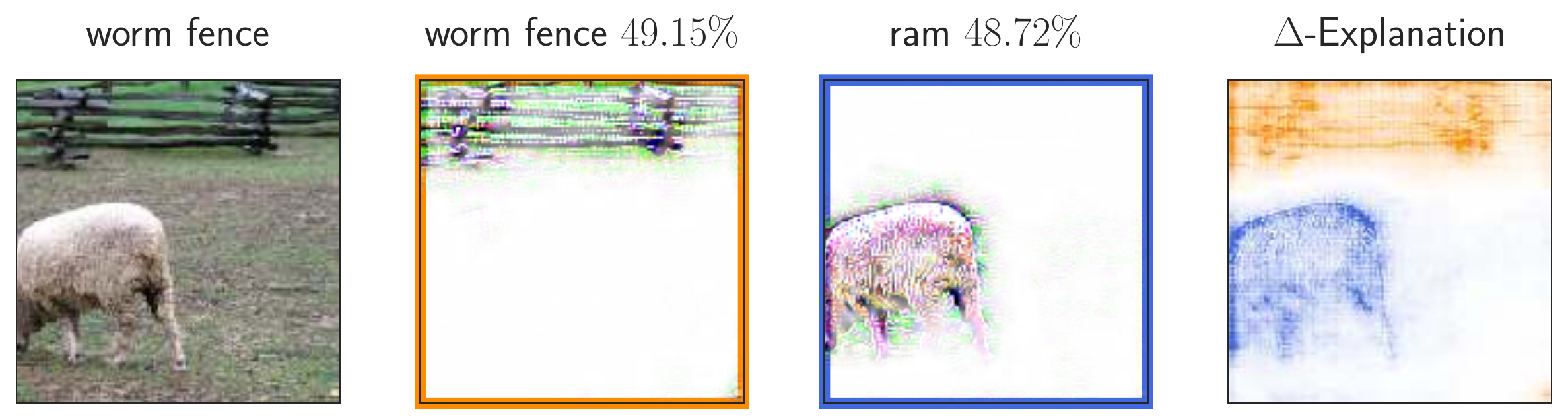}
    \end{subfigure}
    \begin{subfigure}[b]{\linewidth}
    \includegraphics[width=\linewidth, ]{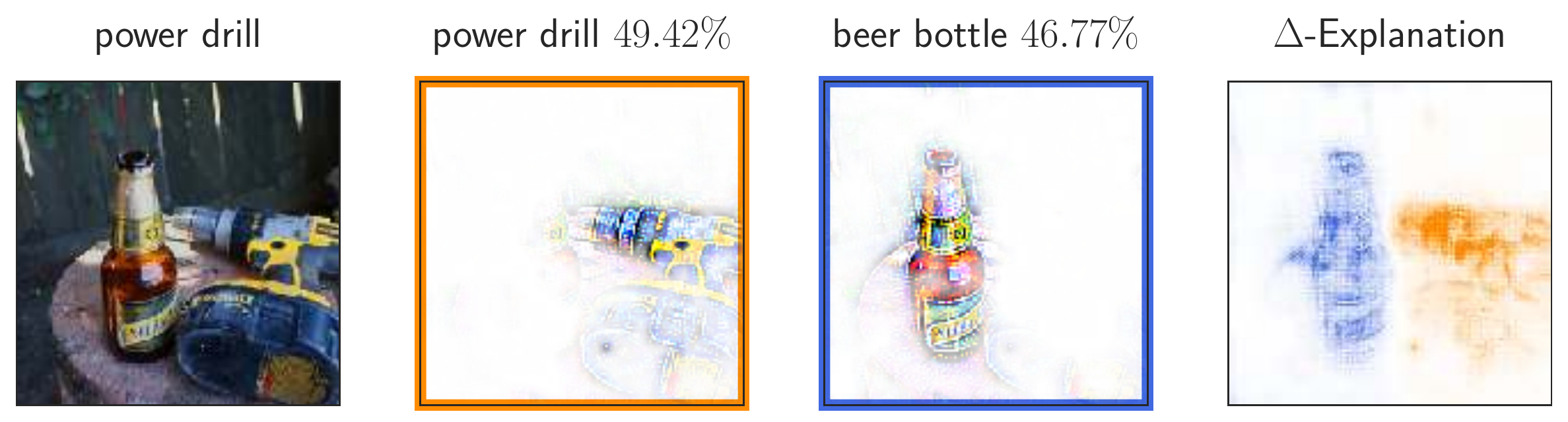}
    \end{subfigure}
    \end{subfigure}
    \caption{\hspace{-.15em}\textbf{Col.~1:} Input image.~\textbf{Cols.~2+3:} Explanations for most likely classes under the model. \textbf{Col.~4:} Difference of contribution maps to the two class logits, i.e., $\vec s_{c_1}^L(\vec x)-\vec s_{c_2}^L(\vec x)$, see \cref{eq:contrib}; positive values shown in orange ($c_1$), negative values in blue ($c_2$).}
    \label{fig:ambiguous_classifications}
    \vspace{-.75em}
\end{figure}

%% file: resources/figures/neurons.tex
\begin{figure*}[t!]
    \vspace{-.75em}
    \centering
    \begin{subfigure}[b]{.975\textwidth}
    \centering
    \begin{subfigure}[b]{\linewidth}
    \begin{subfigure}[b]{\linewidth}
    \includegraphics[width=\linewidth, trim=0 .1em 0 1em, clip]{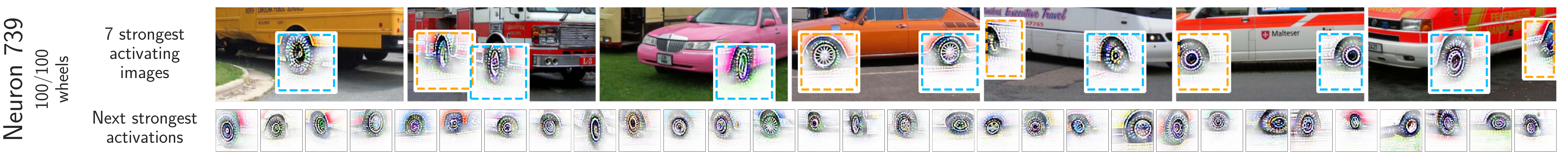}
    \end{subfigure}
    \begin{subfigure}[b]{\linewidth}
    \includegraphics[width=\linewidth, trim=0 .1em 0 1em, clip]{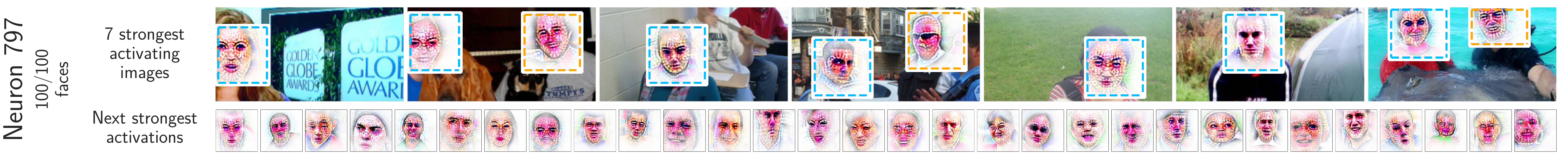}
    \end{subfigure}
    \end{subfigure}
    \begin{subfigure}[b]{\linewidth}
    \begin{subfigure}[b]{\linewidth}
    \includegraphics[width=\linewidth, trim=0 .1em 0 1em, clip]{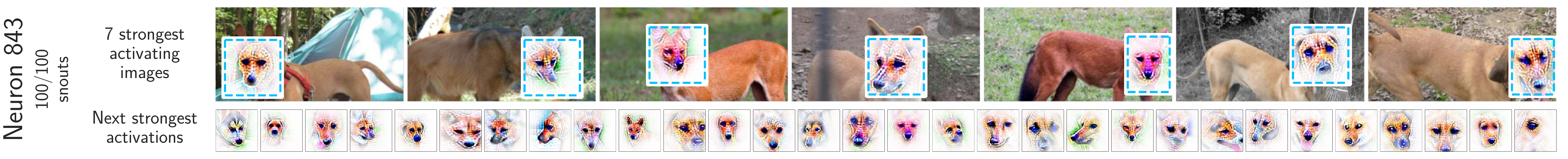}
    \end{subfigure}
    \end{subfigure}
    \begin{subfigure}[b]{\linewidth}
    \begin{subfigure}[b]{\linewidth}
    \includegraphics[width=\linewidth, trim=0 .1em 0 1em, clip]{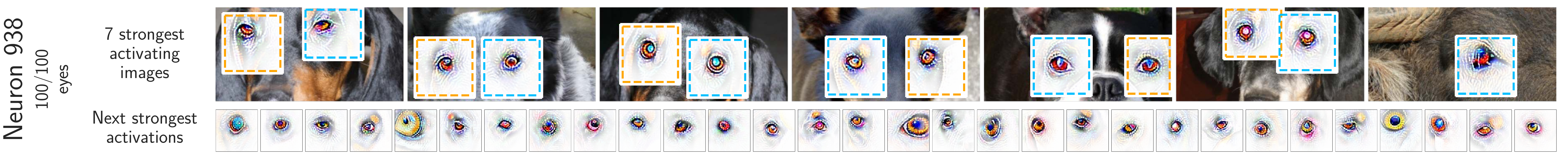}
    \end{subfigure}
    \end{subfigure}
        \begin{subfigure}[b]{\linewidth}
    \begin{subfigure}[b]{\linewidth}
    \includegraphics[width=\linewidth, trim=0 .1em 0 1em, clip]{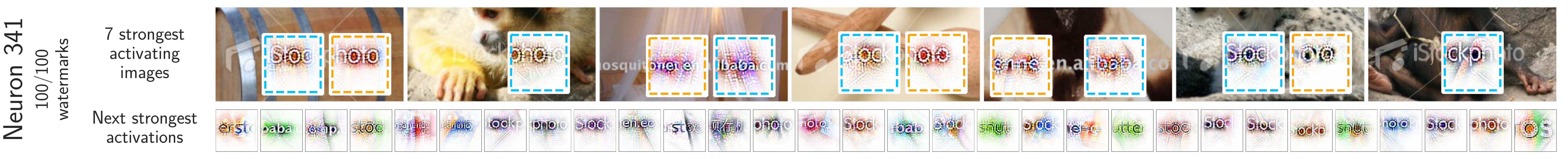}
    \end{subfigure}
    \end{subfigure}
    \end{subfigure}
    \caption[Caption for LOF]{
    Explanations of 5 individual neurons in layer 87 of a DenseNet-121.
    For each neuron, we provide its index number $n$ and its concept description and specificity\footnotemark (\textbf{left}).
    Further, we show the 7 most activating images for each neuron (top row per neuron), in which we visualise the explanation for the highest (\textbf{blue squares}) activation; i.e., visualise the 72$\times$72 center patch of the weighting $[\mat w_{1\rightarrow l}(\vec x)]_n$ for neuron $n$. For some images, we additionally show the explanation for the 2nd highest activation (\textbf{orange squares}). Lastly, we show the explanations of the highest activations (corresponding to the blue squares) for the next 30 images to highlight the neurons' specificity.
    }
    \label{fig:intermediate_protos}
    \vspace{-1.25em}
\end{figure*}

%% file: 6.0-discussion.tex
We presented a novel approach for endowing deep neural networks with a high degree of \emph{inherent interpretability}. 
In particular, we developed the \bcos{} transform as a modification of the linear transform to increase weight-input alignment during optimisation and showed that this can significantly increase interpretability. 
Importantly, the \bcos{} transforms can be used as a drop-in replacement for the ubiquitously used linear transforms in conventional DNNs whilst only incurring minor drops in classification accuracy. 
As such, our approach can increase the interpretability of a wide range of DNNs at a low cost and thus holds great potential to have a significant impact on the deep learning community. In particular, it shows that strong performance and interpretability need not be at odds. Moreover, we demonstrate that by structurally constraining \emph{how} the neural networks are to solve an optimisation task---in the case of \bcos{} networks via \emph{alignment}---allows for extracting explanations that faithfully reflect the underlying model.
We believe this to be  an important step on the road towards interpretable deep learning, which is an essential ingredient for building trust in DNN-based decisions, specifically in safety-critical situations.

%% file: supplement/0.1-main.tex
{
\numberwithin{equation}{section}
\numberwithin{figure}{section}
\numberwithin{table}{section}
\renewcommand{\thefigure}{\thesection\arabic{figure}}
\renewcommand{\thetable}{\thesection\arabic{table}}

\appendix
\onecolumn 
%\flushleft

%\begin{document}
%%%%%%%%% TITLE

\begin{center}
{
\huge\bf \vspace{1em}Supplementary Material\\[2em]}
\end{center}

{
\newcommand{\additem}[2]{%
\item[\textbf{(\ref{#1})}] 
    \textbf{#2} \dotfill\makebox{\textbf{\pageref{#1}}}
}

\newcommand{\addsubitem}[2]{%
    \\[.5em]\indent\hspace{1em}
    \textbf{(\ref{#1})}
    #2 \dotfill\makebox{\textbf{\pageref{#1}}}
}

\newcommand{\adddescription}[1]{\newline
\begin{adjustwidth}{1cm}{1cm}
#1
\end{adjustwidth}
}
% \begin{multicols}{2}

% \begin{minipage}[c]{\textwidth}

{\vspace{2em}\bf\Large Table of Contents\\[2em]}

In this supplement to our work on \bcos{} DNNs, we provide:
\begin{enumerate}[label={({\arabic*})}, topsep=1em, itemsep=.5em]
    \additem{sec:additional_figures}{ 
    Additional qualitative results} 
    \adddescription{In this section, we show additional \emph{qualitative} results of the model-inherent explanations. This includes visualisations for the same model explored in the main paper (DenseNet-121)---both for the class-logits and for intermediate neurons---as well as results for other \bcos{} networks.
    
    Moreover, we provide additional comparisons to post-hoc importance attribution methods that were not shown in the main paper.}
    \additem{sec:additional_quantitative}{ 
    Additional quantitative results} 
    \adddescription{In this section, we show additional \emph{quantitative} results. In particular, we present the localisation metric results for two additional \bcos{} networks as well as those of the pre-trained conventional DNNs. 
    
     Moreover, we present ImageNet results for \bcos{} networks trained without any additional non-linearities (apart from the \bcos{} transform) and for different model sizes.
    
     Finally, we investigate the predictive power of the watermark neurons in more detail.
     }

    \additem{sec:training}{ 
    Implementation details}
    \adddescription{In this section, we describe the model architectures, the training procedure, and the evaluation of model interpretability in more detail.}

    \additem{sec:discussion}{ 
    Additional derivations and discussions} 
    \adddescription{In this section, we provide a short derivation for Eq.~(3)$\rightarrow$Eq.~(9). Further, we provide a more detailed explanation
    of the relevance of the image encoding for visualising the linear transforms $\mat W_{1\rightarrow l}(\vec x)$.}
    
\end{enumerate}

}
\clearpage
% \twocolumn
%\justify
\section{Additional qualitative examples}
\label{sec:additional_figures}
\input{supplement/resources/figures/illustration}
% \vspace{-.}
In \cref{fig:illustration}, we illustrate how the linear mappings $\mat W_{1\rightarrow L}(\vec x)$ are used to compute the outputs of \bcos{} networks. In particular, with this we would like to highlight that these linear mappings do not only constitute qualitatively convincing visualisations. Instead, they in fact constitute the actual linear transformation matrix that the model effectively applies to the input to compute its outputs and thus constitute an accurate summary of the model computations.
\subsection{Additional explanations for class logits [DenseNet-121]}
\myparagraph{Comparisons between explanation methods}
In \cref{fig:add_comp_1}, we show additional comparisons between the model-inherent explanations based on the linear mapping $[\mat w_{1\rightarrow L}(\vec x_i)]_c$ and some post-hoc methods; in particular, we show results for GradCam (GCam) \cite{selvaraju2017grad}, LIME \cite{ribeiro2016lime}, Integrated Gradients (IntG) \cite{sundararajan2017axiomatic}, DeepLIFT \cite{shrikumar2017deeplift}, and RISE \cite{petsiuk2018rise} on the most confidently classified image of the first 15 classes in \cref{fig:add_quali_1}. While GCam highlights similar regions and LIME also yields explanations in color, these explanations are post-hoc approximations of model behaviour. In contrast, the model-inherent explanations are not only of higher visual quality, but also summarise the model computations for the presented classes accurately, cf.~\cref{fig:illustration}.
% \vspace{-1em}
\input{supplement/resources/figures/add_comp}

\myparagraph{Model-inherent explanations.} In \cref{fig:add_quali_1,,fig:add_quali_2}, we present additional qualitative examples of the linear mappings $[\mat w_{1\rightarrow L}(\vec x_i)]_c$ that explain the class logit $c$ in the \bcos{} DenseNet-121 model, see Eq.~(13) in the main paper. Specifically, we show the 3 most confidently classified examples for 48 different classes; these classes were selected as those that had the highest mean confidence (sum of class logits) in the three most confidently classified images.

Note that due to the \textbf{alignment pressure} induced by the \bcos{} transform, the linear mappings $[\mat w_{1\rightarrow L}(\vec x_i)]_c$ align with class-discriminative features in the input images. Interestingly, we find that these features can be highly specific to particular regions in the image (see, e.g., great grey owl, centipede, school bus, planetarium, three-toed sloth, parking meter), but can also cover the entire image and include background features that correlate with the presented classes. For the latter, see e.g., the presented examples of the gondola or the golfcart: for some of these images, the weight matrix also aligns with \emph{context features} in the background. Note, however, that the model has never been explicitly trained to highlight only the respective class objects and it is therefore expected to find that context features are also used by the model to increase its output score for the respective classes.

\input{supplement/resources/figures/add_quali}
\clearpage

\subsection{Additional explanations for intermediate neurons [DenseNet-121]}
In \cref{fig:add_layer7}, we present additional qualitative examples of the linear mappings $[\mat w_{1\rightarrow l}(\vec x_i)]_n$ that explain the activations of intermediate neurons $n$ in layer $l\myeq87$. Specifically, we show 16 out of the 20 most highly activating neurons and their explanations, which were not already shown in the main paper. We find that all neurons seem to represent specific concepts, such as faces, snouts, water, grass, etc.
\input{supplement/resources/figures/add_layer7}

\clearpage

\subsection{Explanations for other \bcos{} networks}
In \cref{fig:add_networks}, we show explanations for neurons in intermediate layers of  a \bcos{} ResNet-34, a \bcos{} InceptionNet, and a \bcos{} VGG-11. We observe that the complexity of the neurons tends to increase, similarly to the neurons in the \bcos{} DenseNet-121 model.
\input{supplement/resources/figures/add_networks}
\clearpage
% In \cref{fig:add_layer7_1,,fig:add_layer7_2}, we present additional qualitative examples of the linear mappings $[\mat w_{1\rightarrow l}(\vec x^i)]_n$ that explain the activations of intermediate neurons $n$ in layer $l\myeq87$. Specifically, we show those neurons and their explanations that had the highest activations over the validation set and that were not already shown in the main paper.
% \input{supplement/resources/figures/add_layer7}

\section{Additional quantitative evaluations}
\label{sec:additional_quantitative}
\subsection{Localisation scores.} 
\noindent In \cref{fig:add_quanti_bcos,,fig:add_quanti_pre}, we present the localisation results of the \emph{grid pointing game} \cite{Boehle2021CVPR}.

In particular, in \cref{fig:add_quanti_bcos}, we show that of all methods, the best explanation for the \bcos{} network is given by the model-inherent linear transforms $\mat W_{1\rightarrow L}(\vec x)$ (Eq.~13, main paper). 

Moreover, from \cref{fig:add_quanti_pre}, we can estimate the \emph{interpretability gain} due to replacing the linear transform in conventional models by the \bcos{} transform: specifically, we see that no method explains the baseline models better than the model-inherent linear transforms explain the respective \bcos{} network. 

We note that LIME and GCam often achieve good localisation scores. However, we would like to highlight the low reliability of those explanations (high variance). Further, LIME requires many forward passes through the model to estimate feature importance, whereas the model-inherent explanations of \bcos{} models can be extracted in a single forward and backward pass.
GCam, on the other hand, only provides explanations with comparably low resolution (cf.~also \cref{fig:add_comp_1}), since it only explains the model's classification head. As such, it does not actually explain the full model, but only a small fraction of it (e.g., 1 out of 121 layers for DenseNet-121). In contrast, the model-inherent explanations of the \bcos{} networks provide high-resolution explanations \emph{in color} and explain the entire model.
\input{supplement/resources/figures/add_quanti}
\clearpage
\subsection{Impact of model size on performance} 
\vspace{-.525em}
\input{supplement/resources/figures/param_ablation}
\vspace{-.525em}
\noindent In \cref{fig:param_ablation,,tbl:param_ablation}, we present the results of two ablation studies. On the one hand, we show the results for a \bcos{} DenseNet-121 model trained without MaxOut, which therefore has a similar number of parameters as the baseline model (a few less due to not using BatchNorm nor biases). On the other hand, we evaluate DenseNet models with two MaxOut units per neuron of different sizes. For this, we modify the \emph{growth factor} of the model architectures, see \cite{huang2017densely}. 

In particular, in \cref{tbl:param_ablation}, we show that despite not employing any non-linearity apart from the \bcos{} transform, the model with no MaxOut units also achieves competitive performance (left-most column in \cref{tbl:param_ablation}). Specifically, we only observe a minor drop in performance with respect to a conventional DenseNet-121 model ($74.4\rightarrow 72.6$, $\Delta\myeq1.8$).

Further, the accuracy of \bcos{} networks improves with model size; note that the \bcos{} DenseNet-121 with a growth factor of 22 and 2 MaxOut units has a similar size to the pretrained baseline DenseNet-121 model. While there is a drop in performance ($74.4\rightarrow 73.2$, $\Delta\myeq1.2$), the \bcos{} version still shows competitive accuracy results. Lastly, note that increasing model size via maxout is computationally more efficient than just increasing the growth factor in a model with a single unit, as the number of feature channels remains unchanged.

\subsection{Class-discriminative information content in watermark neurons} 
\noindent As discussed in the main paper, we observed that some neurons seem to specifically respond to watermarks in images. While this might not seem like a semantically meaningful feature, we find that the distribution of watermarks is in fact highly skewed. In particular, in \cref{fig:watermark}, we plot the distribution of classes among the images corresponding to the 500 highest neuron activations of the `watermark neuron' (index 341); we manually inspected those images and found that neuron 341 consistently activated on text within or overlayed over the images. These images clearly exhibit a non-uniform class distribution, indicating that watermarks indeed represent a highly informative feature for the classification task. 
\input{supplement/resources/figures/watermark}

% \clearpage
\section{Implementation details}
\label{sec:training}
\noindent Here, we provide implementation details regarding implementation of a convolutional B-cos transform (\cref{alg:bcos}), the training procedure (\ref{subsec:training}) and the \emph{post-hoc} attribution methods (\ref{subsec:attributions}). 

{\centering
\RestyleAlgo{ruled}\LinesNumbered\setlength{\algomargin}{.5em}
\begin{algorithm}[htpb]
\caption{\centering Pseudocode for B-cos-Conv2d, cf.~Eq. (9) in the main paper.}
  \label{alg:bcos}
    \setstretch{1}
  \DontPrintSemicolon
  \newcommand{\mycomment}[1]{{\color[RGB]{112, 128, 144}\textit{\# #1}}\;}
  \newcommand{\self}{{\bf\color[RGB]{0,24,128}{self}}}
  \newcommand{\pykey}[1]{{\bf\color[RGB]{0,128,24}{#1}}}
  \newcommand{\pyword}[1]{{\bf\color[RGB]{197,117,50}{#1}}}
  \SetKwFunction{FMain}{bcos\_conv2d}
  \SetKwFunction{Finit}{bcos\_conv2d}
  \SetKwFunction{Ffwd}{forward}
  \SetKwProg{Fn}{\pykey{class}}{:}{}
  \SetKwProg{Imp}{}{}{}
  \SetKwProg{Df}{\pykey{def}}{:}{}
%   \Imp{\normalfont \pykey{from} torch \pykey{import} nn}{}\vspace{-.25em}
%   \Imp{\normalfont \pykey{import} torch.nn.functional \pykey{as} F}{}\vspace{-.6em}\;\vspace{-.25em}
%   \Df{\FMain}{\;
    \mbox{\mycomment{$\vec x$: input, $\widehat{\mat w}$: normed weights, $k$: kernel size, $d_f$: index of feature dimension}}\;
    \Df{\Finit{\textit{$\vec x, \widehat{\mat w}, k$, B} }}{ 
        % \mycomment{in\_channels: `input channels'; out\_channels: `number of capsules'; rank: `rank of matrix $\mat{AB}$'} 
        {linear\_out} = conv2d($\vec x$, $\widehat{\mat w}$)  \mycomment{$\myeq\widehat{\mat w}\vec x$}
        {norm} = sumpool2d($\vec x$.pow(2).sum($d_f$), k).sqrt()\;
        {cos} = linear\_out / norm.unsqueeze($d_f$)\;
        scaling = cos.abs().pow(B-1)  \mycomment{$\myeq|c(\vec x; \widehat{\mat w})|^{\text{B}-1}$}
        \KwRet\ scaling $\ast$ linear\_out \mycomment{$\myeq|c(\vec x; \widehat{\mat w})|^{\text{B}-1}\widehat{\mat w}\vec x$}
        }\end{algorithm}
}

\subsection{Training and evaluation procedure}
\label{subsec:training}
\subsubsection{CIFAR10}
\myparagraph[-.25]{Architecture.}
For our CIFAR10 experiments, we used a 9-layer architecture with the following specifications: 
kernel size $k=[3, 3, 3, 3, 3, 3, 3, 3, 1]$, stride $s=[1, 1, 2, 1, 1, 2, 1, 1, 1]$, padding $p=[1, 1, 1, 1, 1, 1, 1, 1, 0]$, and 
output channels \mbox{$o=[64, 64, 128, 128, 128, 256, 256, 256, 10]$} for layers $l=[1, 2, 3, 4, 5, 6, 7, 8, 9]$ respectively.

When increasing the parameter B, we observed the input signal to decay strongly over the network layers, which resulted in zero outputs and hindered training. To overcome this, we scaled all layer outputs with a fixed scalar $\gamma$, which we set such that $\log_{10}\gamma = 1.5\times B - 1.75$, which improved signal propagation. 
To counteract the artificial upscaling of the signal at the network output, we divided the network output by a fixed constant $T$ for each B, such that $\log_{10}T=[-3, -3, -2, 1, 2, 2, 3]$ for B$\,=[1, 1.25, 1.5, 1.75, 2, 2.25, 2.5]$ respectively. In future work, we aim to examine how to automatically set an optimal scale for a given network in more detail.

\myparagraph{Training.}
We trained our CIFAR10 models for 100 epochs with Adam \cite{kingma2014adam}, an initial learning rate of $1\times10^{-3}$, and a batch size of 64. Further, we used a cosine learning rate schedule and decayed the learning rate to $1\times10^{-5}$ over the 100 epochs and applied horizontal flipping and padded random cropping for augmenting the data. We used a bias term of $\vec b = \log(0.1/0.9)$, which yields a uniform probability distribution for zero inputs ($[\vec f(\vec x=\vec 0)]_i = [\sigma(\mat W_{1\rightarrow L}\, \vec 0 + \vec b)]_i = 0.1\;\forall \; i$).

% In our experiments, we used a bias term $\vec b = \log(0.01/0.99)$ on ImageNet and $\vec b = \log(0.1/0.9)$ for CIFAR10. 
% When applying RandAugment, 
\subsubsection{ImageNet}
\myparagraph{Training.}
{Similar to the CIFAR10 experiments, we observed signals to decay quickly for deep networks and to be dependent on the number of channels used. To overcome this, we scaled the layer outputs by $\gamma = s/\sqrt{d}$ with $s$ a network-dependent hyperparameter and $d$ the input dimensionality (i.e., $k^2c$ for a convolutional layer with kernel size $k$ and an input with $c$ feature channels). Specifically, we chose $s=100$ for DenseNets and ResNets, $s=200$ for the InceptionNet, and $s=1000$ for the VGG model. 

Moreover, as in the CIFAR10 experiments, we divided the network outputs by a temperature parameter $T$.
In detail, for the results in this paper we set $\log_{10}\,T= -3$ for the DenseNet models, $\log_{10}\,T= 1$ for ResNet, $\log_{10}\,T= 0$ for InceptionNet, and $\log_{10}\,T= -1$ for the VGG model. These parameters were experimentally determined to achieve good accuracies and stable training behaviour. In future work, we plan to investigate how to set the temperature parameter automatically.
%Similar to the CoDA Networks~\cite{Boehle2021CVPR}, we found the parameter $T$ to influence the alignment pressure. Specifically, for a larger temperature, the network is required to align better with the inputs to achieve the same classification confidence as a model with low temperature. 
}

Finally, we added the auxiliary loss in the InceptionNet with a weighting of $\lambda=1$, used images of size $s=299$ for Inception (224 otherwise), and employed RandAugment with $n=2$ and $m=9$. The bias term $\vec b$ was set to $\vec b = \log(0.01/0.99)$ for all ImageNet experiments.

\subsection{Attribution methods}

\label{subsec:attributions}
\noindent We compare the model-inherent explanations, given by the linear transform $\mat W_{1\rightarrow L}(\vec x)$, against the following post-hoc attribution methods: the vanilla gradient (Grad, \cite{baehrens2010explain}), `Input$\times$Gradient' (IxG, cf.~\cite{adebayo2018sanity}), Integrated Gradients (IntGrad, \cite{sundararajan2017axiomatic}), DeepLIFT (\cite{shrikumar2017deeplift}), GradCam (GCam, \cite{selvaraju2017grad}), LIME (\cite{ribeiro2016lime}), and (RISE \cite{petsiuk2018rise}).

For all methods except RISE, LIME, and GCam, we rely on the captum library (\url{github.com/pytorch/captum}). For IntGrad, we set $\text{n\_steps}=50$ for integrating over the gradients. For RISE and LIME, we used the official implementations available at \url{github.com/eclique/RISE} and \url{github.com/marotcr/lime} respectively. We generated 500 masks for RISE and set the hyperparameters $s$ and $p$ to their default values of $s=8$ and $p=0.1$. Similarly, we used 500 samples for LIME, and used the default values for the kernel size ($k=4$) and the number of features ($n=5$).
\subsubsection{Localisation metric}
\label{subsubsec:localisation}
We evaluated all attribution methods on the \emph{grid pointing game} \cite{Boehle2021CVPR}. For this, we constructed 500 $3\times3$ grid images. For an example of a $2\times2$ grid, see Fig.~3 in the main paper. As was done in \cite{Boehle2021CVPR}, we sorted the images according to the models' classification confidence for each class and then sampled a random set of classes for each multi-image. For each of the sampled classes, we then included the most confidently classified image in the grid that had not already been used in a previous grid image. 

\section{Additional derivations and discussions}
\label{sec:discussion}
\subsection{On the \bcos{} transform in matrix form}
\label{subsec:derivation}
In the following, we provide additional details on how to express the \bcos{} transform in matrix form.

As shown in Eq.~(3) in the main paper, the \bcos{} transform is given by
\begin{align}
\label{eq:bcos}
    \text{B-cos}(\vec x;\vec w)&=||\widehat{\vec w}||\;||\vec x||\times|c(\vec x, \widehat{\vec w})|^\text{B} \times \text{sgn} (c(\vec x, \widehat{\vec w}))\;,\\
    \text{with} \quad\qquad c(\vec x, \vec w) &= \cos(\angle(\vec x, \vec w))\;,\\
    \widehat{\vec w} &= \vec w / ||\vec w||\;,
\end{align}
$\angle(\vec x, \vec w)$ returning the angle between $\vec x$ and $\vec w$, and sgn the sign function.
Note that the sign function can be expressed as $\text{sgn}(a) = a/|a|$ for $|a|\neq0$ and zero otherwise.
Hence, \cref{eq:bcos} can be expressed as 
\begin{align}
\label{eq:bcos2}
    \text{B-cos}(\vec x;\vec w)&=||\widehat{\vec w}||\;||\vec x||\times|c(\vec x, \widehat{\vec w})|^\text{B} \times \text{sgn} (c(\vec x, \widehat{\vec w}))\\
   \text{\footnotesize(replace sgn)} \quad\qquad &=||\widehat{\vec w}||\;||\vec x||\times|c(\vec x, \widehat{\vec w})|^\text{B} 
                                    {\color{red}\;\times\; c(\vec x, \widehat{\vec w}) / |c(\vec x, \widehat{\vec w})|}\\
    \text{\footnotesize(combine cos terms)} \quad\qquad &=||\widehat{\vec w}||\;||\vec x||
                                        {\;\times\; |c(\vec x, \widehat{\vec w})|^{\color{red}\text{B}-1} \times c(\vec x, \widehat{\vec w})}\\
    \text{\footnotesize(reorder)} \quad\qquad &=||\widehat{\vec w}||\;||\vec x||
                       { \color{red}\times c(\vec x, \widehat{\vec w})  \times |c(\vec x, \widehat{\vec w})|^{\text{B}-1}}\\
    \text{\footnotesize(write first three factors as linear transform)} \quad\qquad &={\color{red}\widehat{\vec w}^T \vec x} \times |c(\vec x, \widehat{\vec w})|^{\text{B}-1}\;. \label{eq:final_bcos}
\end{align}
For clarity, we marked the changes between lines in the above equations in red. 

From \cref{eq:final_bcos} it becomes clear that a \bcos{} transform simply computes a rescaled linear transform. Thus, multiple units in parallel (i.e., a layer $\vec l^*$ of \bcos{} units) can easily be expressed in matrix form via 
\begin{align}
    \vec l^*(\vec x) = |c(\vec x, \widehat{\mat W})|^{\text{B}-1} \times \widehat{\mat W}\vec x\;. \label{eq:main9}
\end{align}
Here, the $\times$, $\cos$, and absolute value operators are applied element-wise and the rows of $\widehat{\mat W}$ are given by $\widehat{\vec w}_n$ of the individual units $n$.

Hence, the output of each unit (entry in output vector $\vec l^*$) is the down-scaled linear transform from \cref{eq:final_bcos}.
Note that \cref{eq:main9} is the same as Eq.~(9) in the main paper.

\subsection{On the relevance of image encoding for the visualisations}
\label{subsec:discussion}
\noindent As we describe in the main paper, we encode image pixels as $[r,g,b,1\myminus r, 1\myminus g, 1\myminus b]$. This has two important advantages. 

On the one hand, as argued by \cite{Boehle2021CVPR}, this overcomes a bias towards bright pixels. For this, note that the model output is computed as a linear transform of the input $\vec x$. As such, the contribution to the output per pixel is given by the weighted input strength. In particular, 
a specific pixel location $(i, j)$ with color channels $c$ contributes $\sum_c w_{(i, j, c)} x_{(i, j, c)}$ to the output. Under the conventional encoding---i.e., $[r,g,b]$---, a black pixel is encoded by $x_{(i, j, c)}\myeq0$ for $c\in\{1, 2, 3\}$ and can therefore not contribute to the model output. Since we train the model to maximise its outputs (binary cross entropy loss, see Sec.~3.2.2 in the main paper), the network will preferentially encode bright pixels, as these can produce higher contributions for maximising the output than dark pixels. In contrast, under the new encoding dark and bright pixels have the same amount of `signal' that can be weighted, i.e., $\sum_c x_{(i, j, c)} \myeq 3\; \forall\; (i, j)$.

Moreover, this encoding allows to unambiguously infer the color of a pixel solely based on the angle of the pixel vector $[r,g,b,1\myminus r, 1\myminus g, 1\myminus b]$. To contrast this with the original encoding, consider a pixel that is (almost) completely black and given by $[r, g, b]$ with $g\myeq0, b\myeq0, r\myeq0.001$. This pixel has the same angle as a red pixel, given by $r\myeq1, g\myeq0, b\myeq0$. Thus, these two colors cannot be disambiguated based on their angle. By adding the three additional color channels $[1\myminus r, 1\myminus g, 1\myminus b]$, each color channel is uniquely encoded by the direction of the color channel vector, e.g.,$[r, 1\myminus r]$. Finally, note that the \bcos{} transform induces an alignment pressure on the weights, i.e., the model weights are optimised such that $\mat w_{1\rightarrow L}$ points in the same direction as (important features in) the input. Consequently, the weights will reproduce the \emph{angles} of the pixels, but there is no constraint on their \emph{norm}. Since the angle is sufficient for inferring the color, we can nevertheless decode the angles of the weight vectors into RGB colors, as, e.g., shown in \cref{fig:add_networks,,fig:add_layer7,,fig:add_comp_1,,fig:add_quali_1,,fig:add_quali_2}.
\clearpage
{
%\bibliographystyle{supplement/ieee_fullname}
%\bibliography{supplement/egbib}
}
% 

%\end{document}

%% file: supplement/resources/figures/illustration.tex
\begin{figure}
    \centering
    \begin{subfigure}[b]{.6\linewidth}
    \centering
    \includegraphics[width=\textwidth]{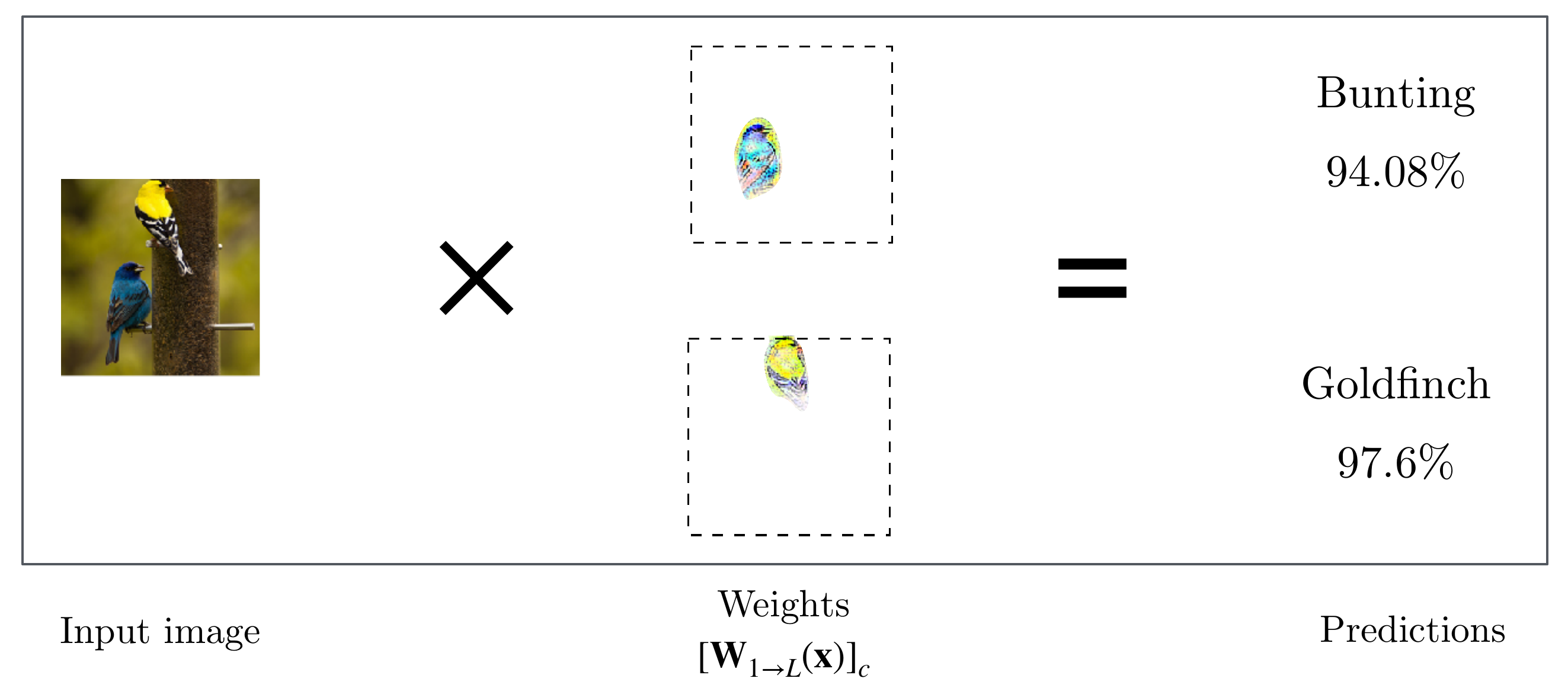}
    \end{subfigure}
    % \vspace{-2em}
    \caption{Illustration of the computations of a \bcos{} network. For a given input image (left), the model computes an \emph{input-dependent} linear transform $\mat w_{1\rightarrow L}(\vec x)$ (center). The scalar product between the input and the weights $[\mat w_{1\rightarrow L}(\vec x)]_c$ for class $c$ (row $c$ of $\mat w_{1\rightarrow L}(\vec x)$), yields the class logits for the respective class. To obtain class probabilities (right), we apply the sigmoid function.
    Since the \bcos{} networks are trained with the BCE loss, they produce probabilities \emph{per class} and \emph{not a probability distribution over classes}. Thus, the probabilities do not sum to 1. For illustration purposes, we only visualise the positive contributions according to $\mat w_{1\rightarrow L}(\vec x)$.
    }
    % \vspace{-.5em}
    \label{fig:illustration}
\end{figure}

%% file: supplement/resources/figures/add_comp.tex
\begin{figure}[h!]
    \centering
    \begin{subfigure}[b]{.32\linewidth}
    \centering
    \begin{subfigure}[b]{.975\linewidth}\centering
        \includegraphics[width=\linewidth]{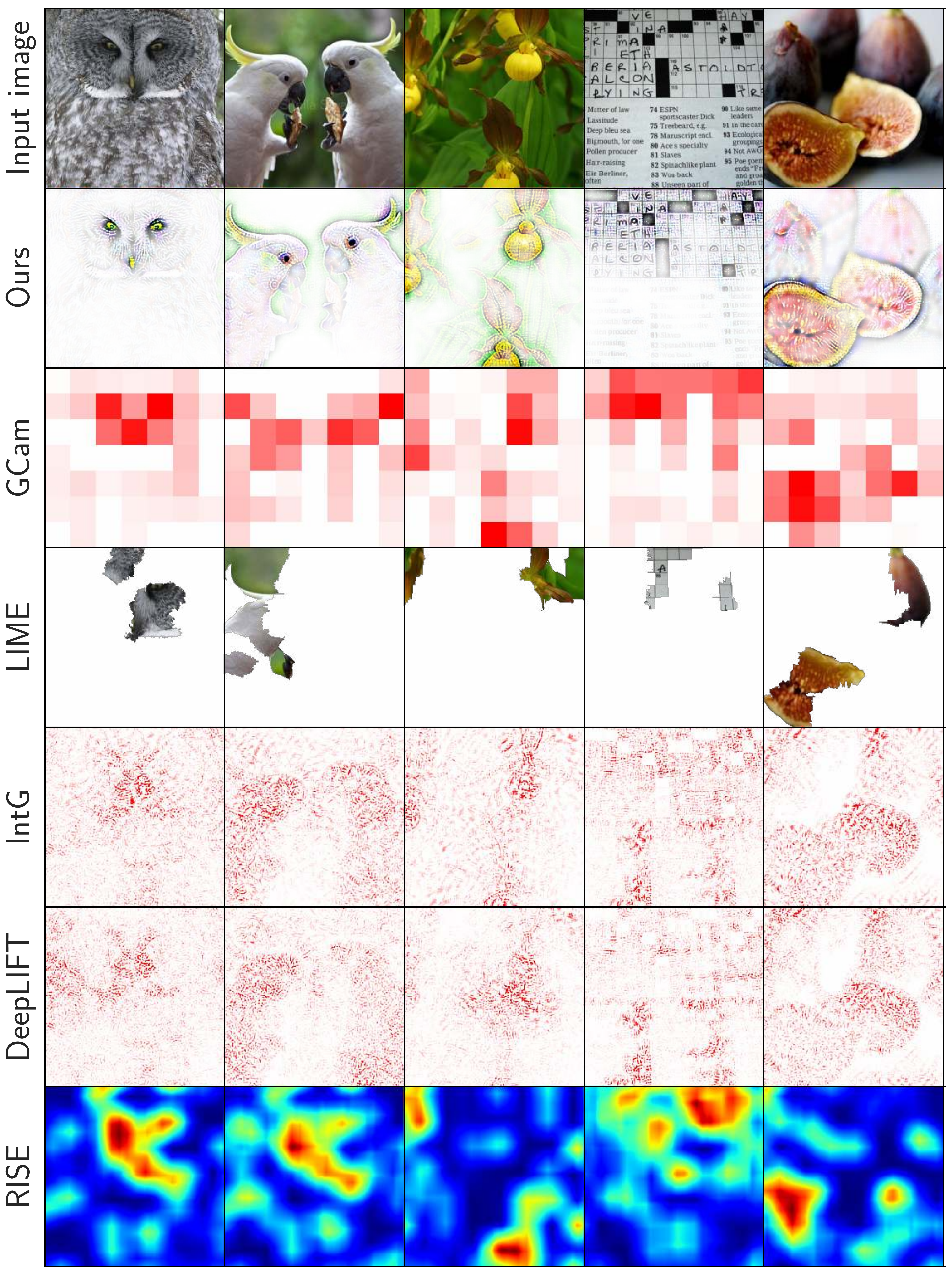}
        \end{subfigure}
    \end{subfigure}
    \begin{subfigure}[b]{.32\linewidth}
    \centering
    \begin{subfigure}[b]{.975\linewidth}\centering
        \includegraphics[width=\linewidth]{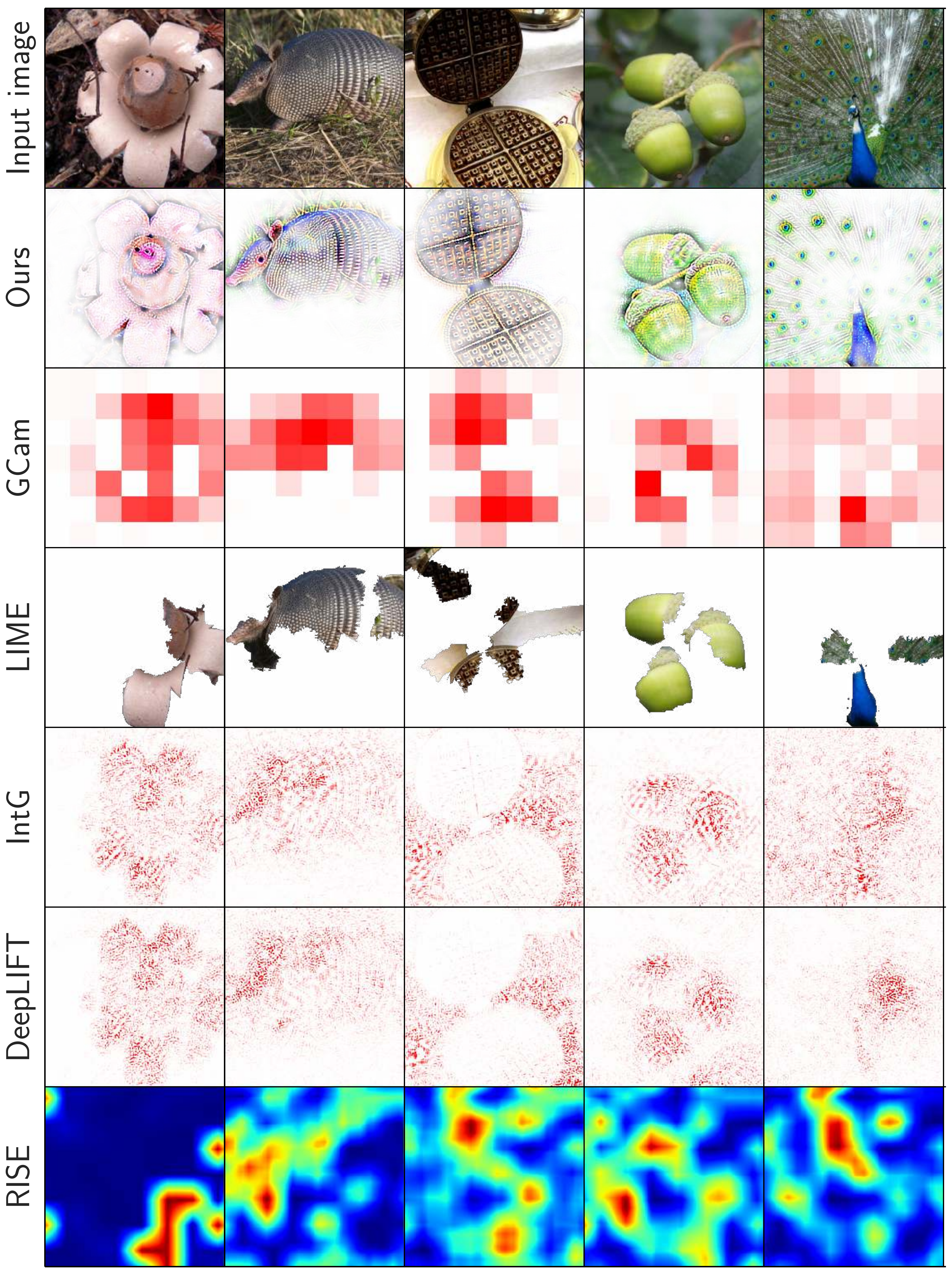}
        \end{subfigure}
    \end{subfigure}
    \begin{subfigure}[b]{.32\linewidth}
    \centering
    \begin{subfigure}[b]{.975\linewidth}\centering
        \includegraphics[width=\linewidth]{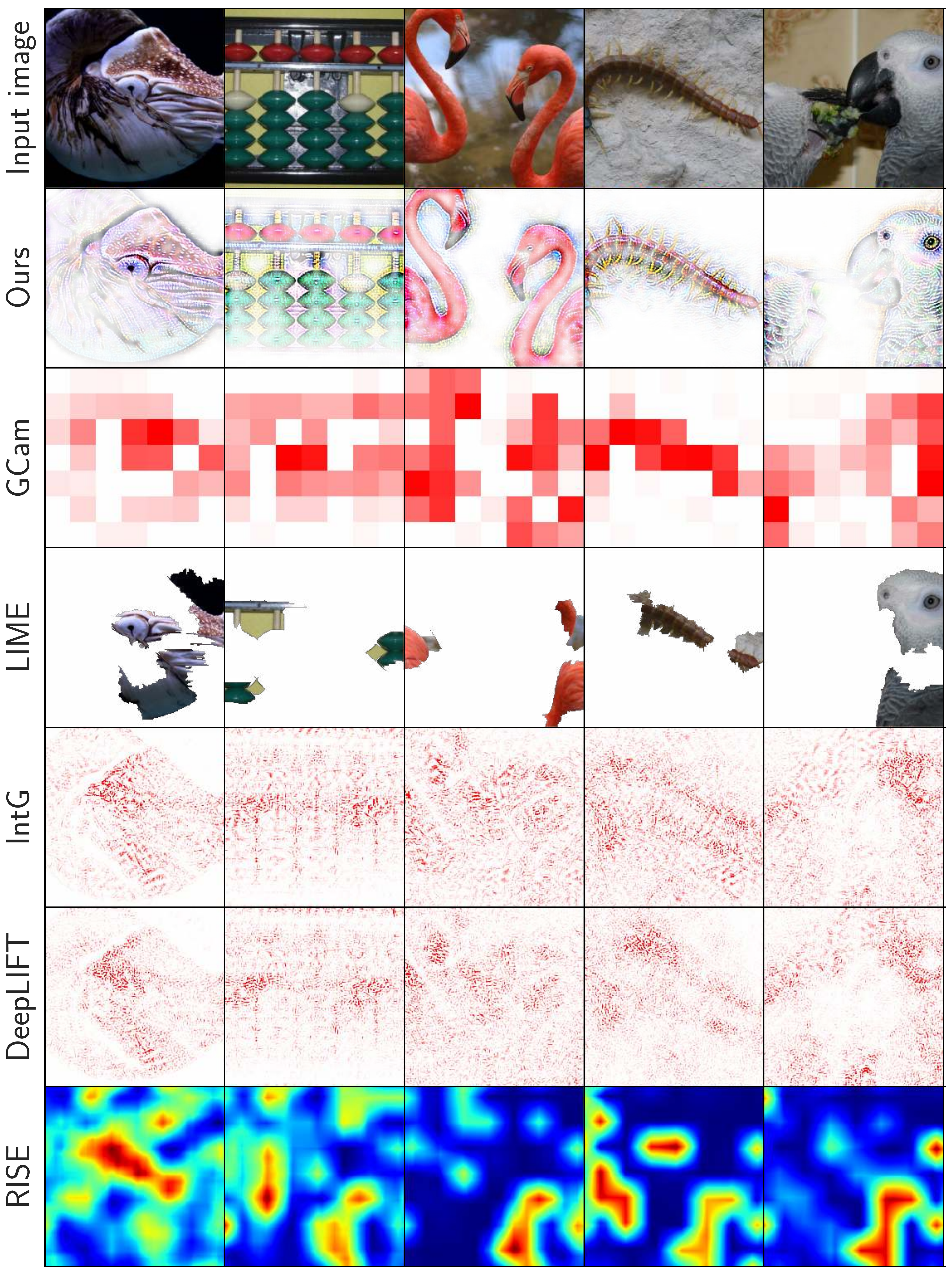}
        \end{subfigure}
    \end{subfigure}
    \caption{Comparison between the model-inherent explanations (`Ours') and various post-hoc explanation methods, evaluated for the most confident image for the first 15 of the classes shown in \cref{fig:add_quali_1,,fig:add_quali_2}. Note that for RISE we use its default colormap.}
    \label{fig:add_comp_1}
\end{figure}

%% file: supplement/resources/figures/add_quali.tex
\begin{figure}[b!]
\vspace{-1em}
    \centering
    \begin{subfigure}[b]{.45\linewidth}
    \centering
    
        \begin{subfigure}[b]{.975\linewidth}\centering
        \includegraphics[width=\linewidth]{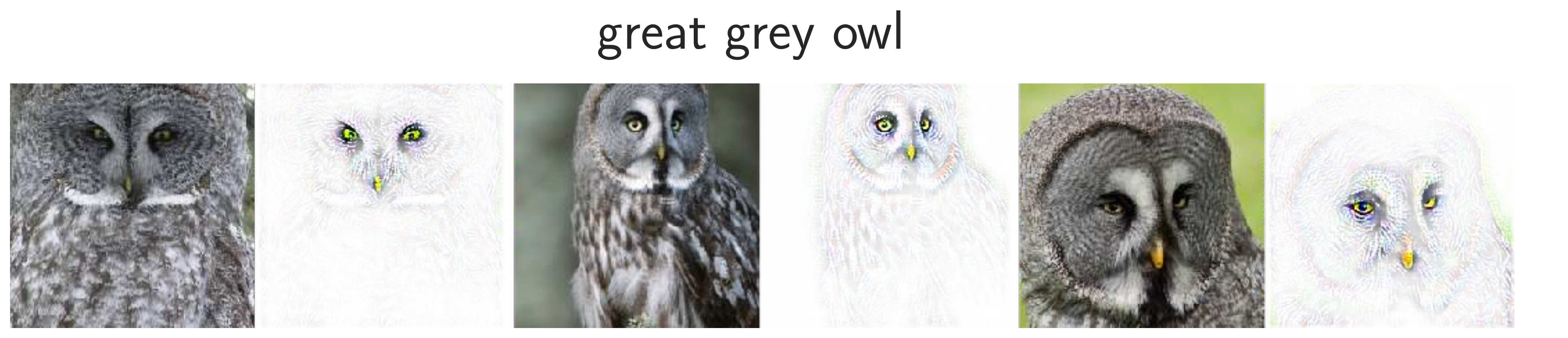}
        \end{subfigure}
    
        \begin{subfigure}[b]{.975\linewidth}\centering
        \includegraphics[width=\linewidth]{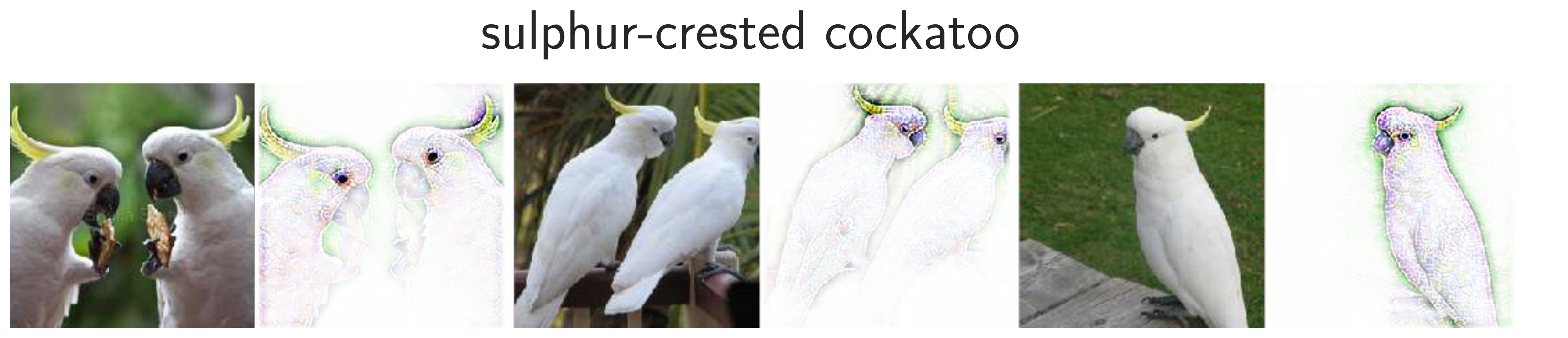}
        \end{subfigure}
    
        \begin{subfigure}[b]{.975\linewidth}\centering
        \includegraphics[width=\linewidth]{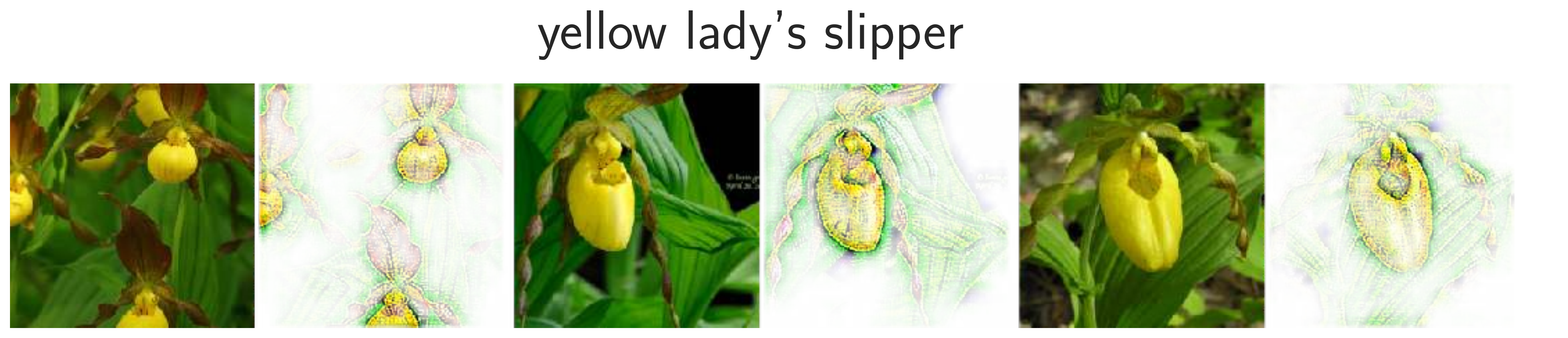}
        \end{subfigure}
    
        \begin{subfigure}[b]{.975\linewidth}\centering
        \includegraphics[width=\linewidth]{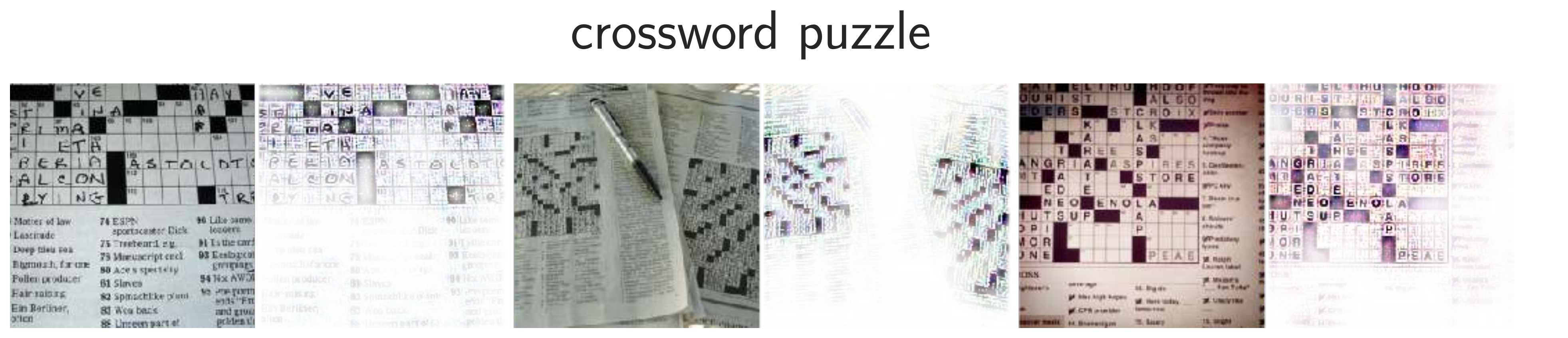}
        \end{subfigure}
    
        \begin{subfigure}[b]{.975\linewidth}\centering
        \includegraphics[width=\linewidth]{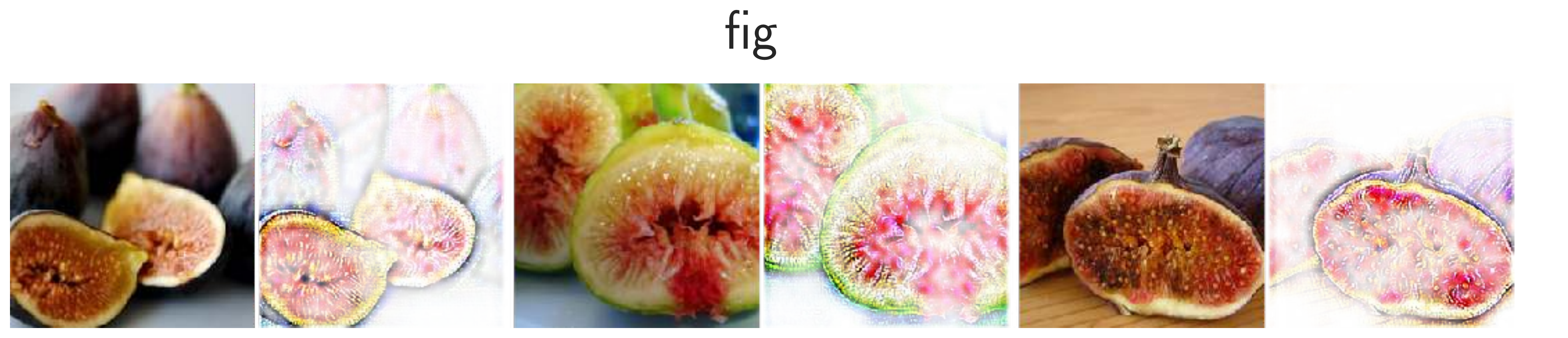}
        \end{subfigure}
    
        \begin{subfigure}[b]{.975\linewidth}\centering
        \includegraphics[width=\linewidth]{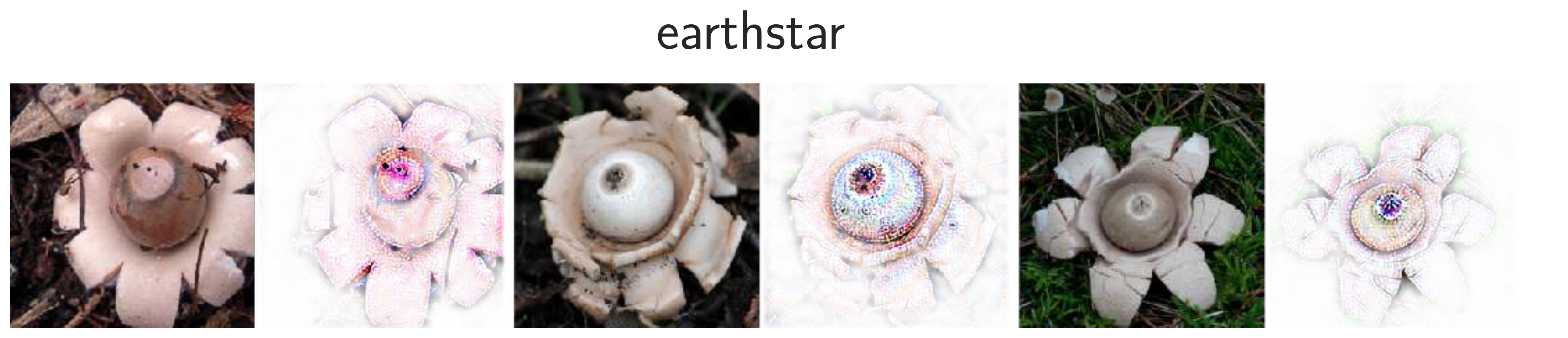}
        \end{subfigure}
    
        \begin{subfigure}[b]{.975\linewidth}\centering
        \includegraphics[width=\linewidth]{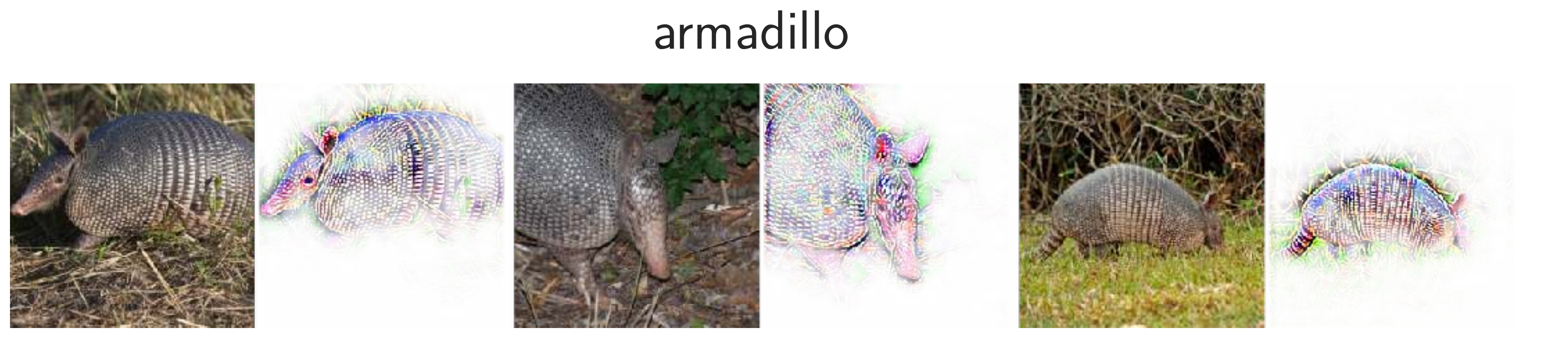}
        \end{subfigure}
    
        \begin{subfigure}[b]{.975\linewidth}\centering
        \includegraphics[width=\linewidth]{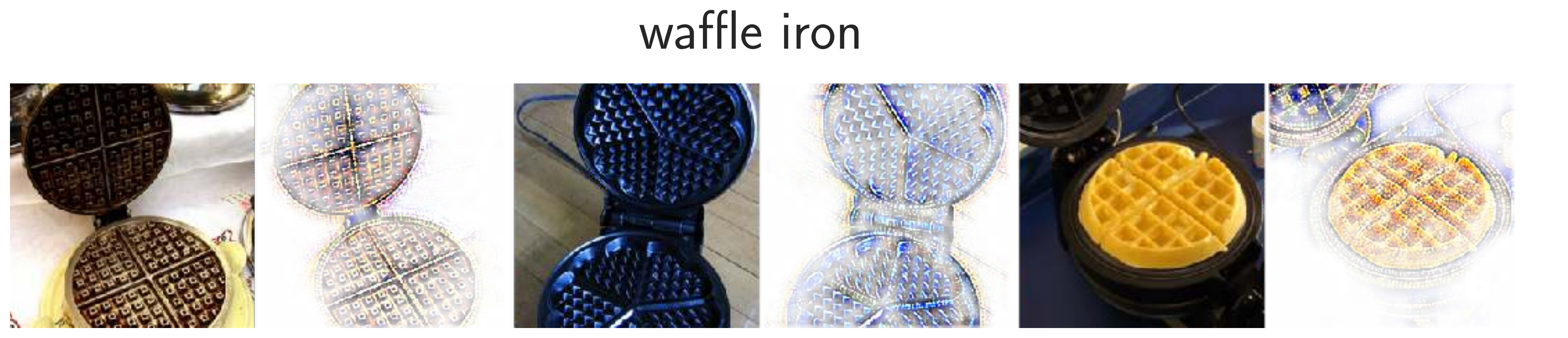}
        \end{subfigure}
    
        \begin{subfigure}[b]{.975\linewidth}\centering
        \includegraphics[width=\linewidth]{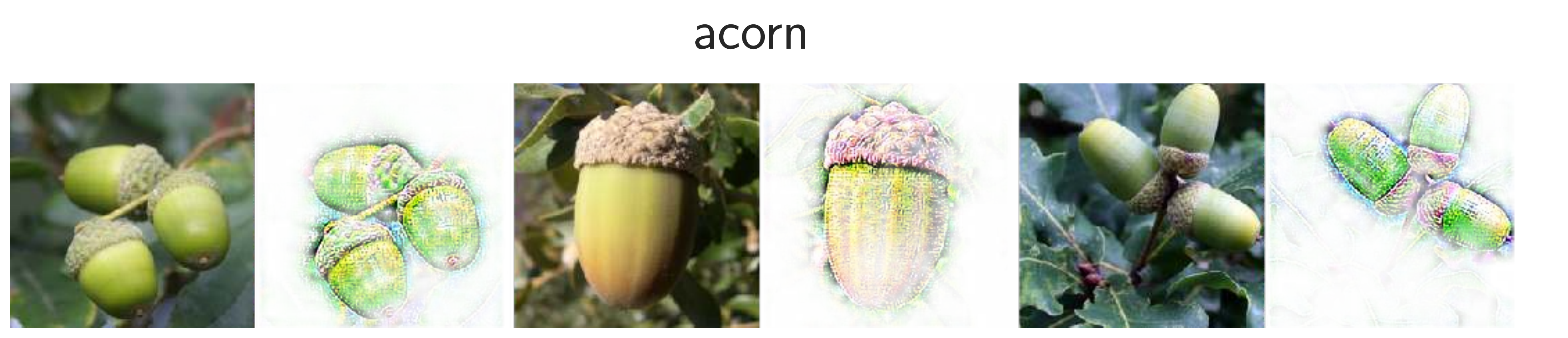}
        \end{subfigure}
    
        \begin{subfigure}[b]{.975\linewidth}\centering
        \includegraphics[width=\linewidth]{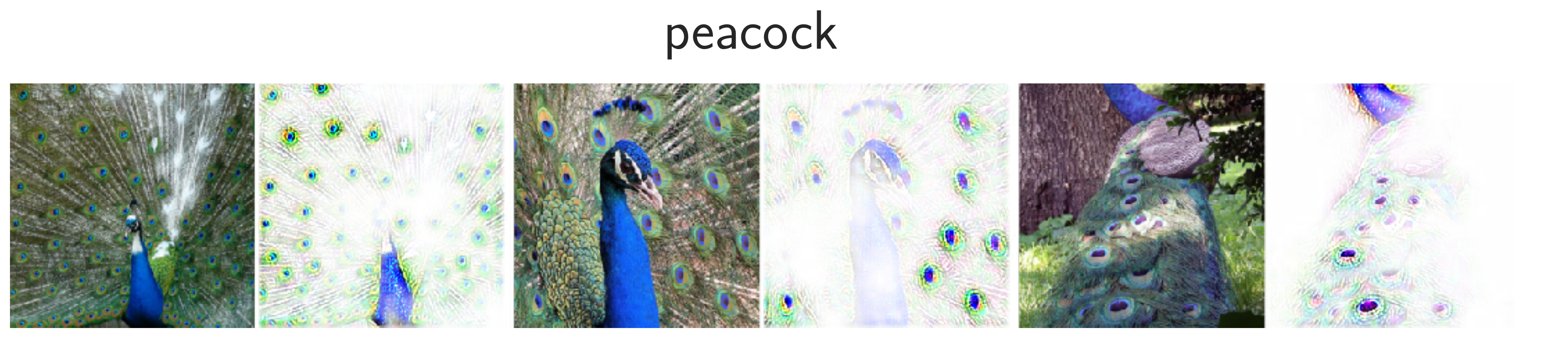}
        \end{subfigure}
    
        \begin{subfigure}[b]{.975\linewidth}\centering
        \includegraphics[width=\linewidth]{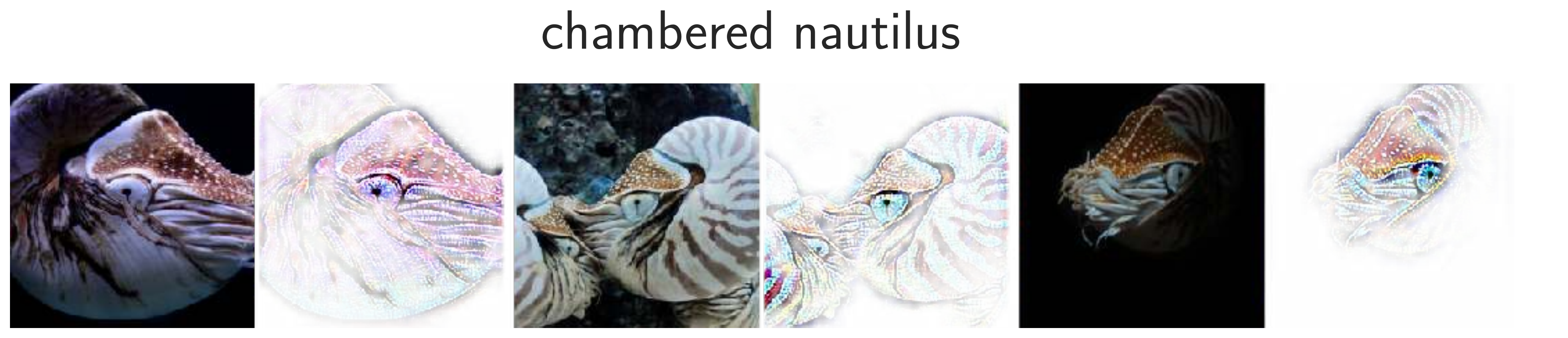}
        \end{subfigure}
    
        \begin{subfigure}[b]{.975\linewidth}\centering
        \includegraphics[width=\linewidth]{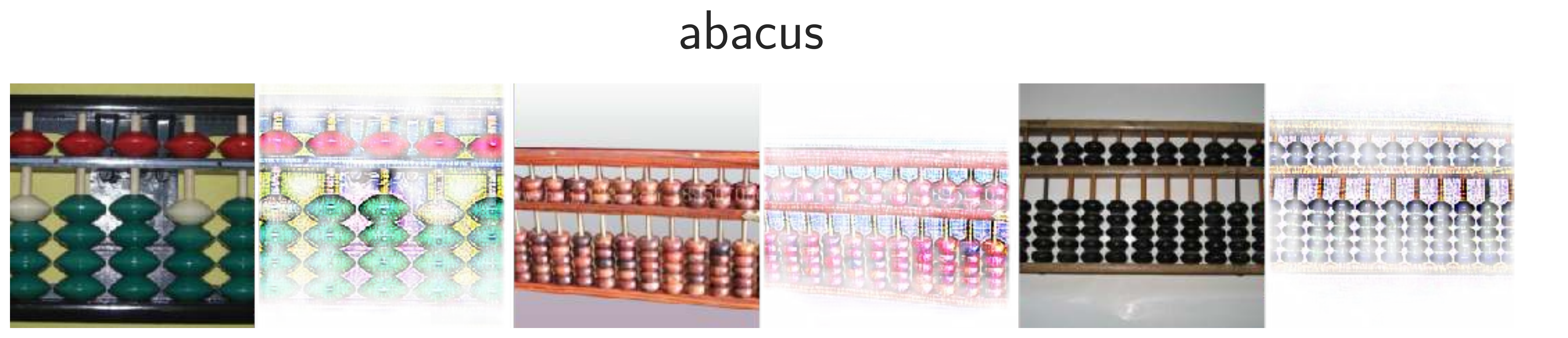}
        \end{subfigure}
    \end{subfigure}
    \begin{subfigure}[b]{.45\linewidth}
    \centering
            \begin{subfigure}[b]{.975\linewidth}\centering
        \includegraphics[width=\linewidth]{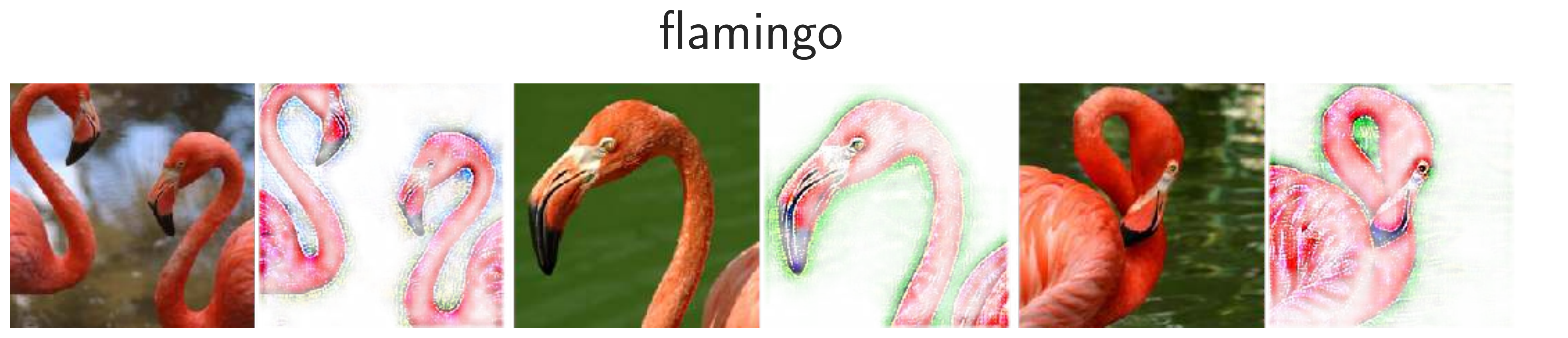}
        \end{subfigure}
    
        \begin{subfigure}[b]{.975\linewidth}\centering
        \includegraphics[width=\linewidth]{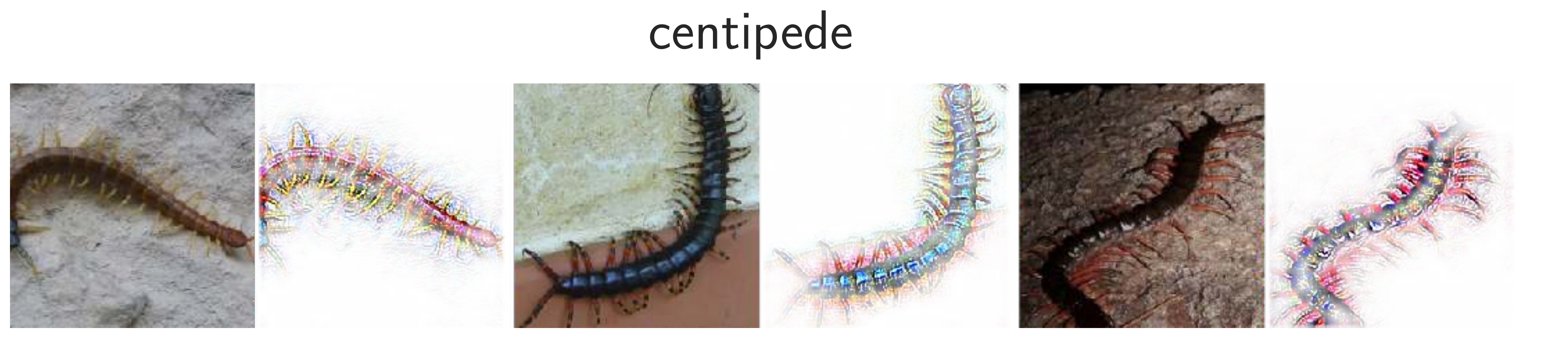}
        \end{subfigure}
    
        \begin{subfigure}[b]{.975\linewidth}\centering
        \includegraphics[width=\linewidth]{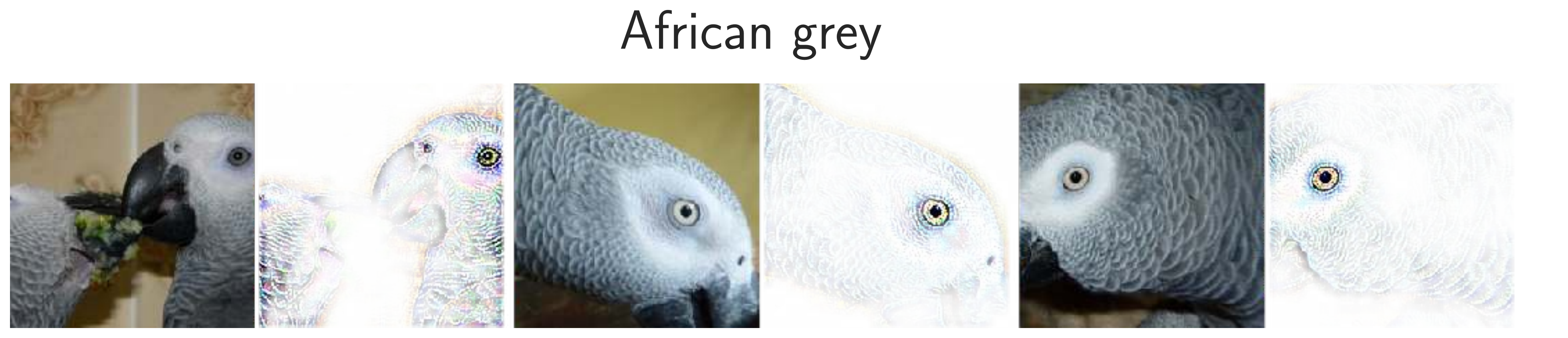}
        \end{subfigure}
    
        \begin{subfigure}[b]{.975\linewidth}\centering
        \includegraphics[width=\linewidth]{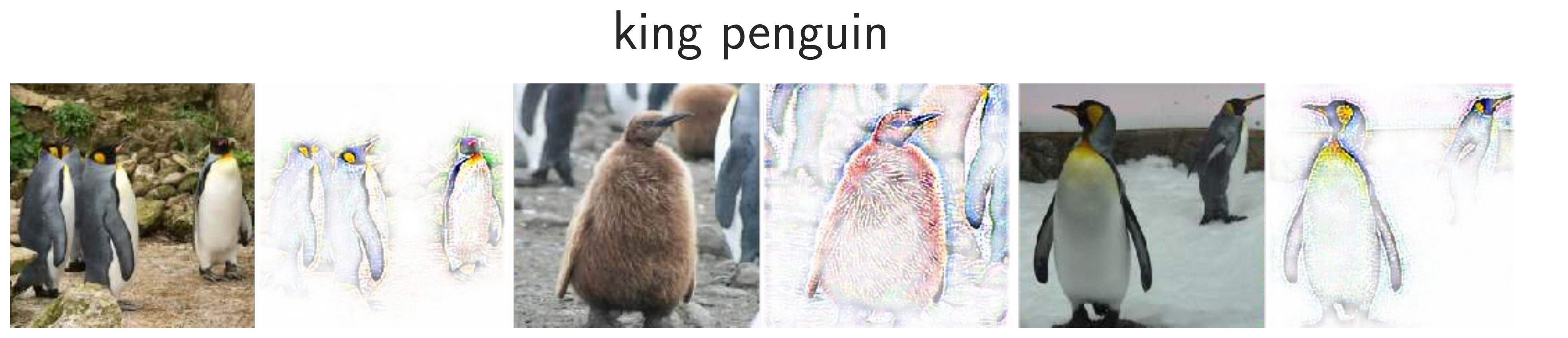}
        \end{subfigure}
    
        \begin{subfigure}[b]{.975\linewidth}\centering
        \includegraphics[width=\linewidth]{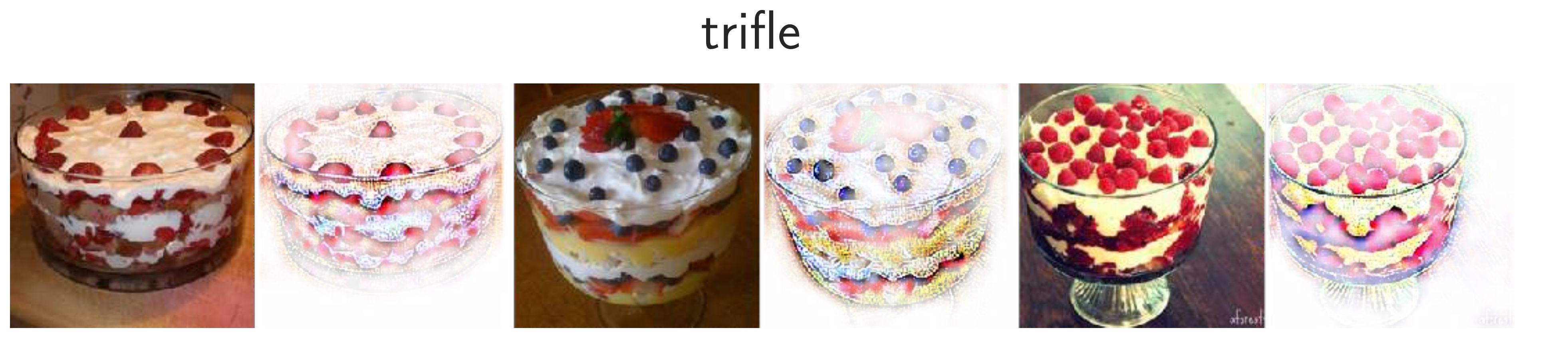}
        \end{subfigure}
    
        \begin{subfigure}[b]{.975\linewidth}\centering
        \includegraphics[width=\linewidth]{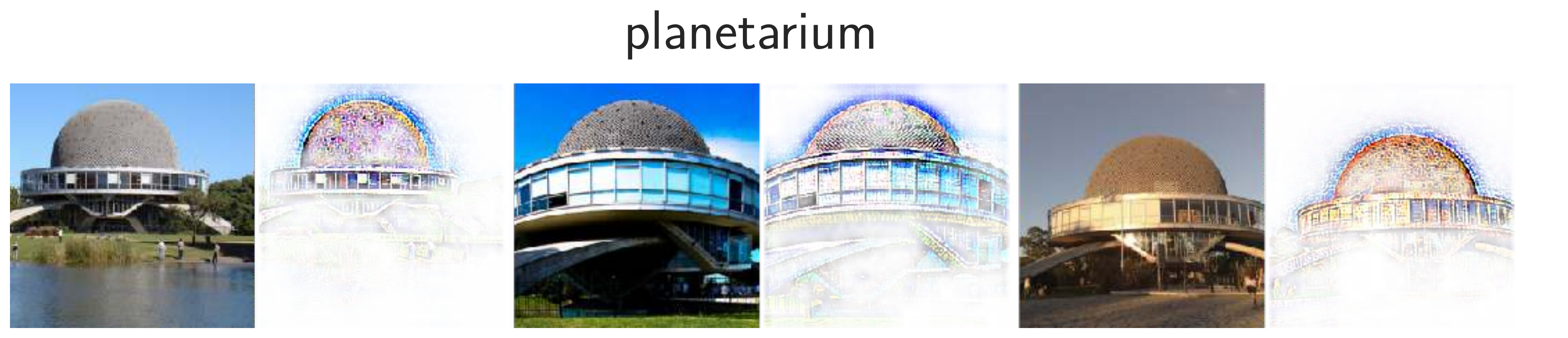}
        \end{subfigure}
    
        \begin{subfigure}[b]{.975\linewidth}\centering
        \includegraphics[width=\linewidth]{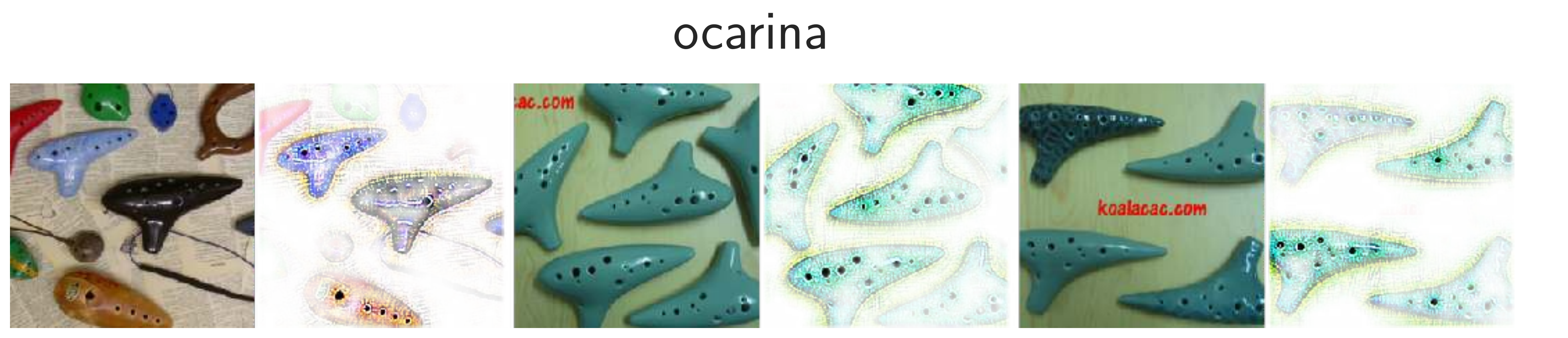}
        \end{subfigure}
    
        \begin{subfigure}[b]{.975\linewidth}\centering
        \includegraphics[width=\linewidth]{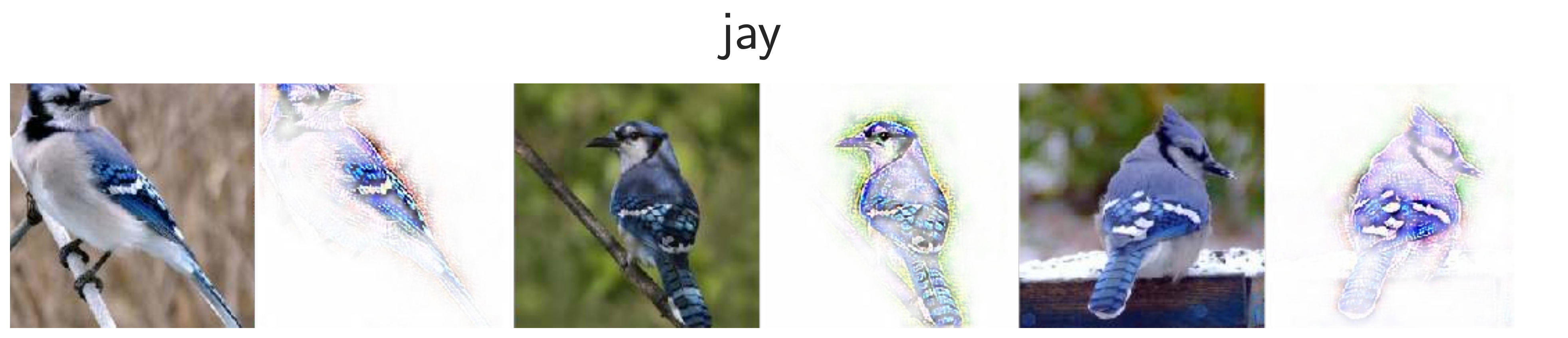}
        \end{subfigure}
    
        \begin{subfigure}[b]{.975\linewidth}\centering
        \includegraphics[width=\linewidth]{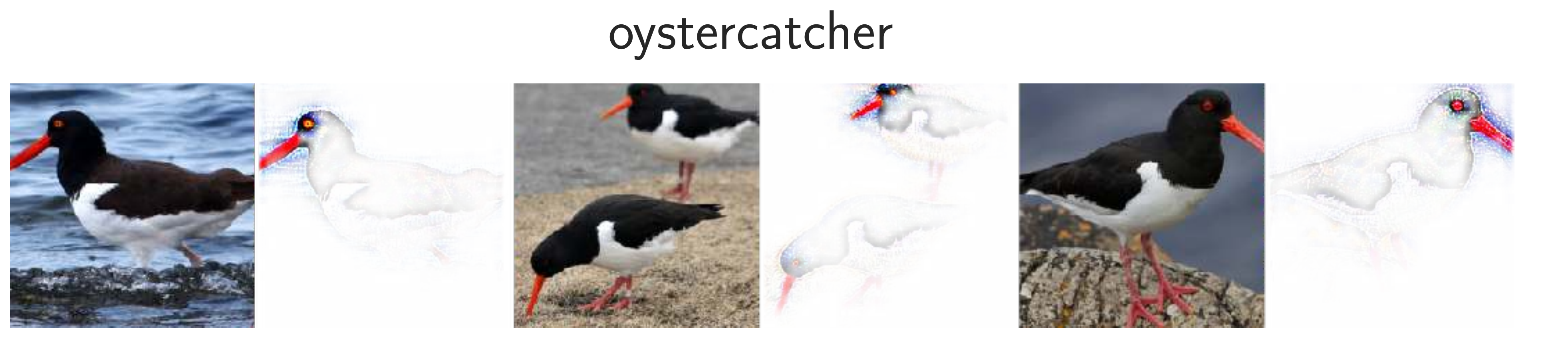}
        \end{subfigure}
    
        \begin{subfigure}[b]{.975\linewidth}\centering
        \includegraphics[width=\linewidth]{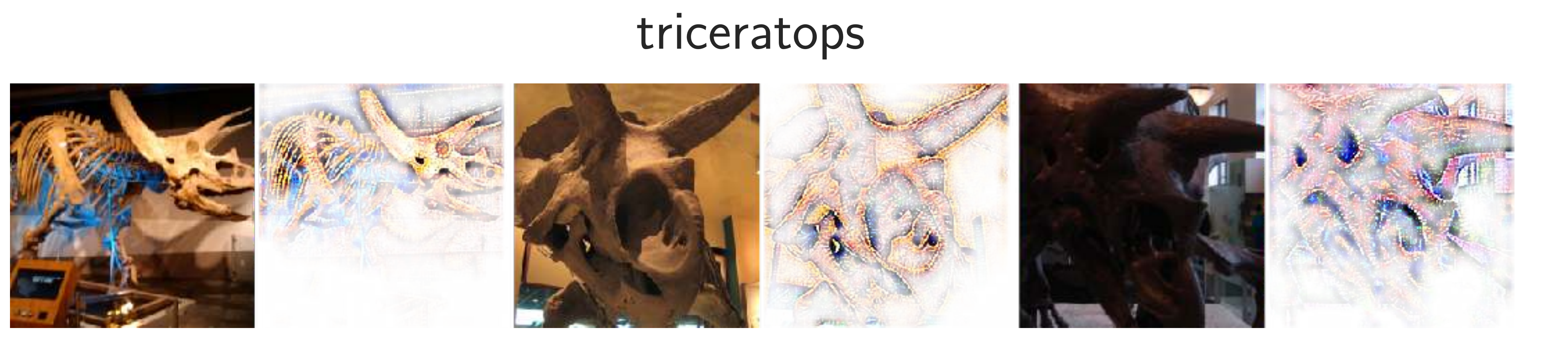}
        \end{subfigure}
    
        \begin{subfigure}[b]{.975\linewidth}\centering
        \includegraphics[width=\linewidth]{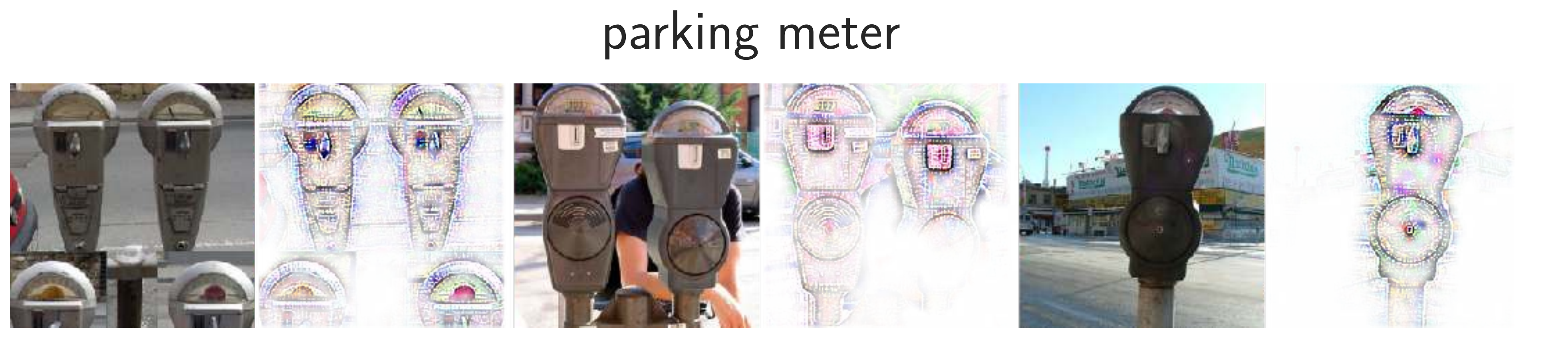}
        \end{subfigure}
    
        \begin{subfigure}[b]{.975\linewidth}\centering
        \includegraphics[width=\linewidth]{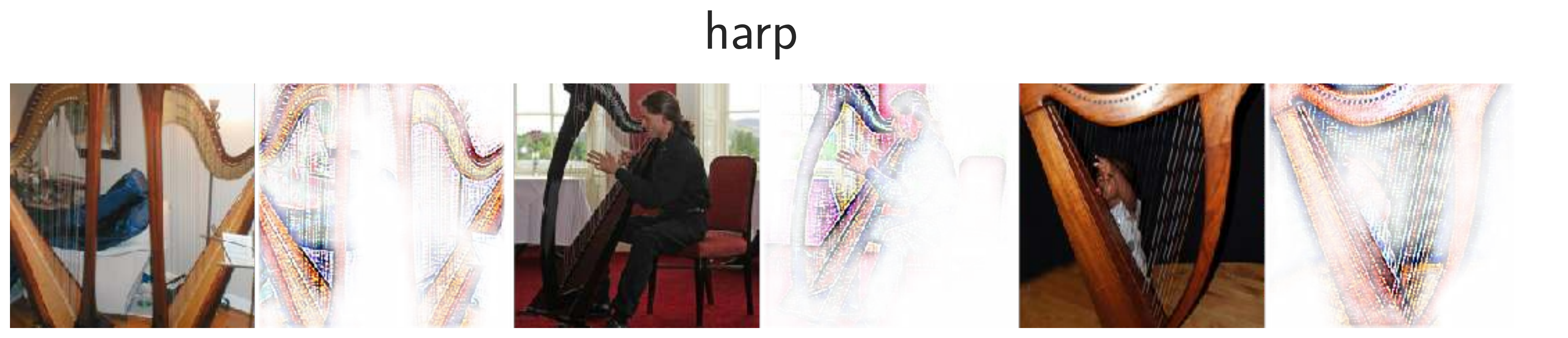}
        \end{subfigure}
    
    \end{subfigure}
    \caption{\hspace{-.15em}First three samples $\vec x_i^c$ and linear mappings $[\mat{w}_{1\rightarrow L}(\vec x_i^c)]_c $ for 24 of the most confidently classified classes $c$ from the Imagenet dataset. Specifically, the classes are sorted by the sum of the logits for those three samples. Left: Classes 1-12. Right: Classes 13-24.}
    \label{fig:add_quali_1}
\end{figure}

\begin{figure}[t!]
    \centering
    \begin{subfigure}[b]{.45\linewidth}\centering
                \begin{subfigure}[b]{.975\linewidth}\centering
        \includegraphics[width=\linewidth]{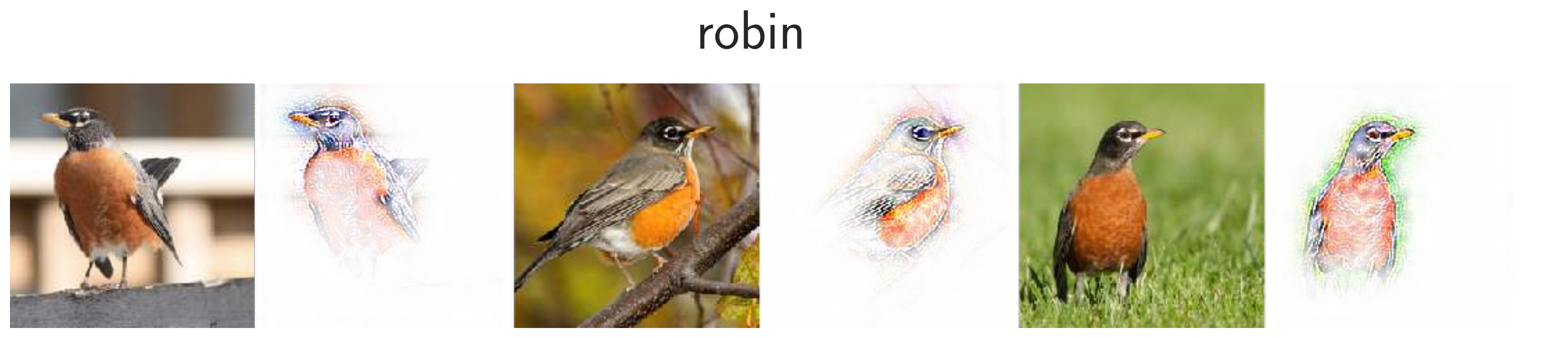}
        \end{subfigure}
    
        \begin{subfigure}[b]{.975\linewidth}\centering
        \includegraphics[width=\linewidth]{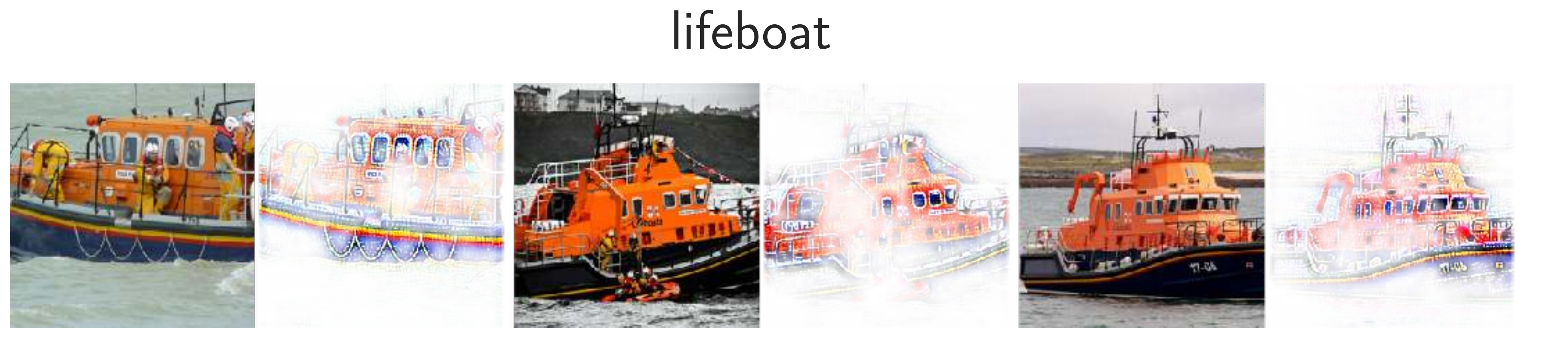}
        \end{subfigure}
    
        \begin{subfigure}[b]{.975\linewidth}\centering
        \includegraphics[width=\linewidth]{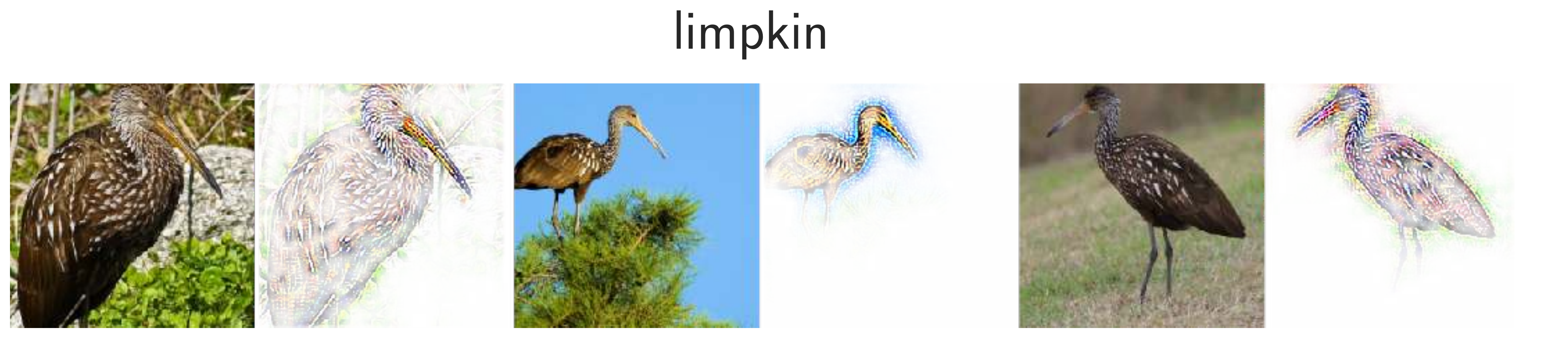}
        \end{subfigure}
    
        \begin{subfigure}[b]{.975\linewidth}\centering
        \includegraphics[width=\linewidth]{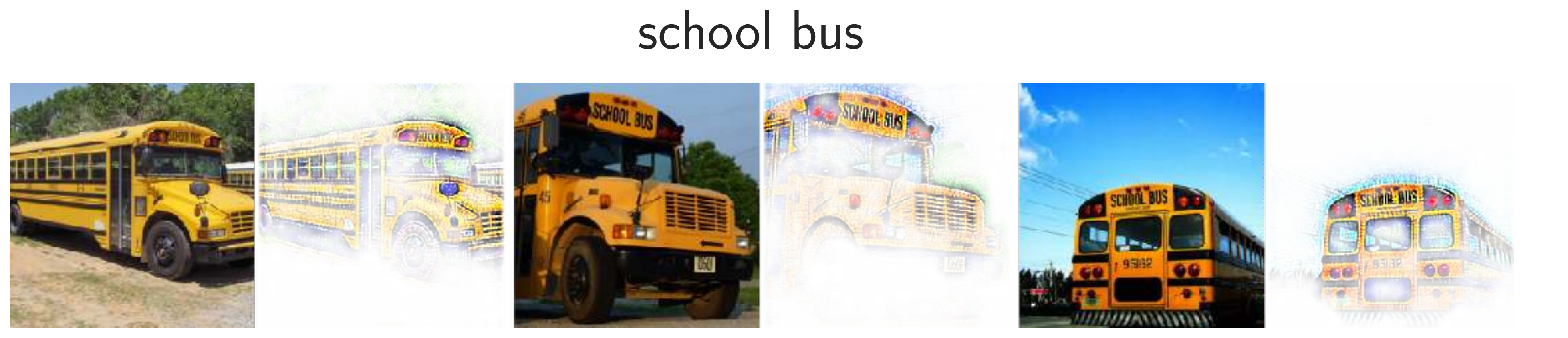}
        \end{subfigure}
    
        \begin{subfigure}[b]{.975\linewidth}\centering
        \includegraphics[width=\linewidth]{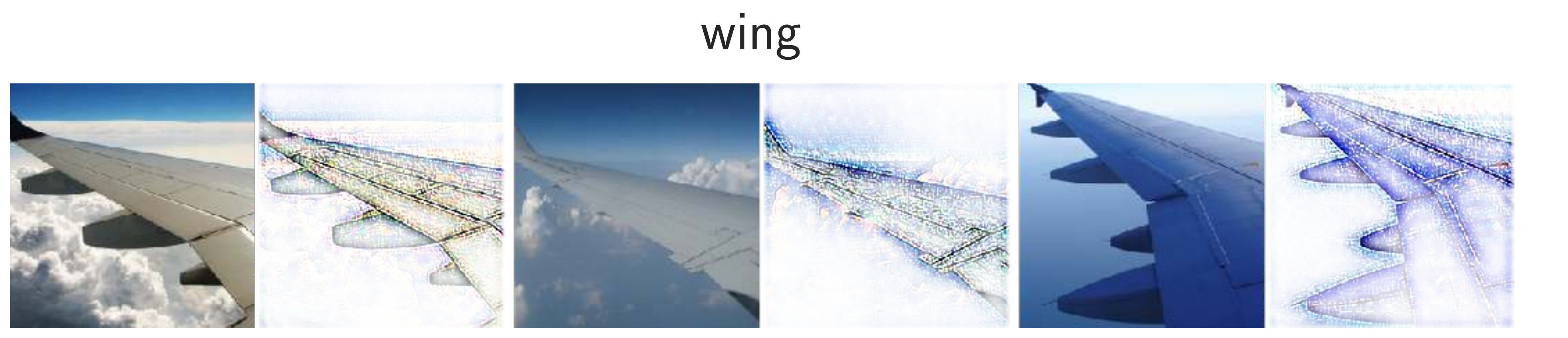}
        \end{subfigure}
    
        \begin{subfigure}[b]{.975\linewidth}\centering
        \includegraphics[width=\linewidth]{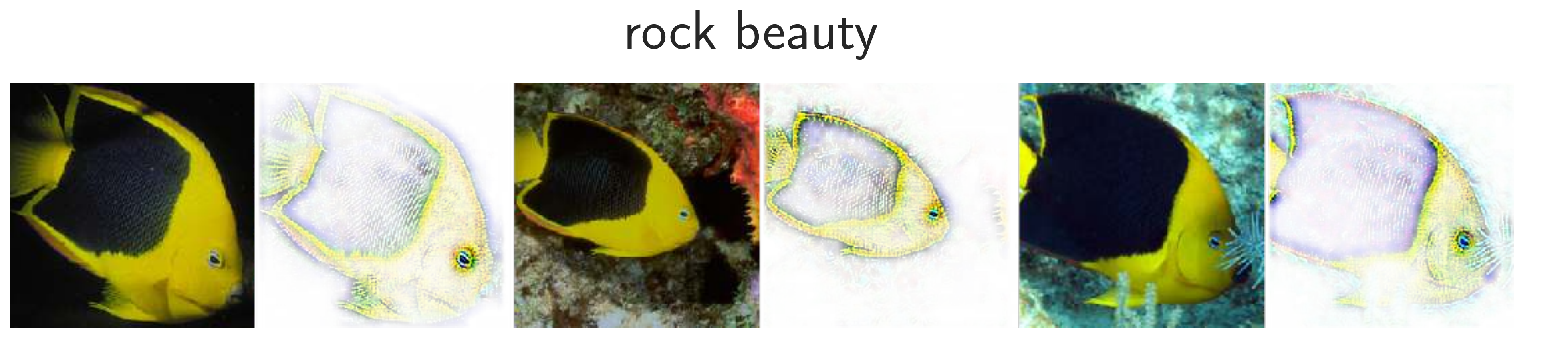}
        \end{subfigure}
    
        \begin{subfigure}[b]{.975\linewidth}\centering
        \includegraphics[width=\linewidth]{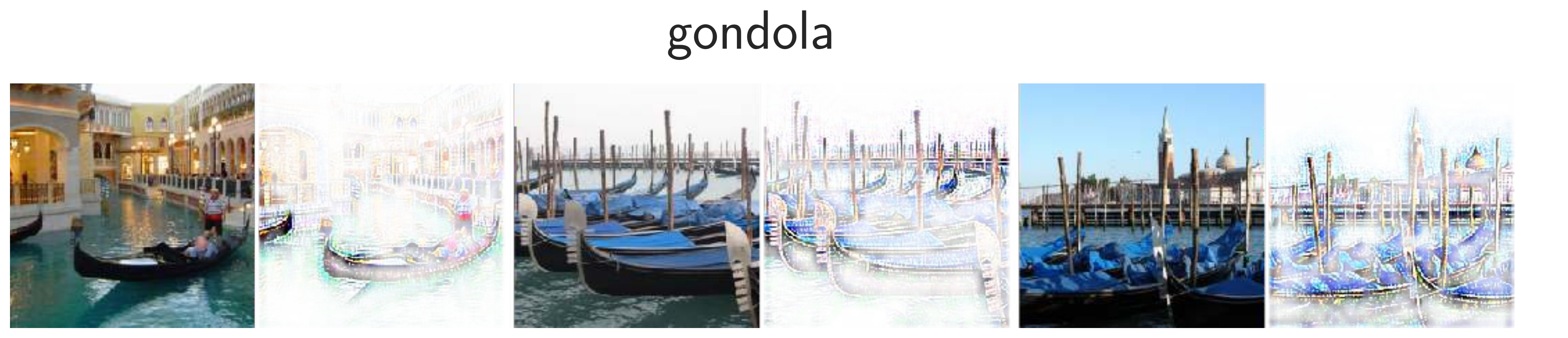}
        \end{subfigure}
    
        \begin{subfigure}[b]{.975\linewidth}\centering
        \includegraphics[width=\linewidth]{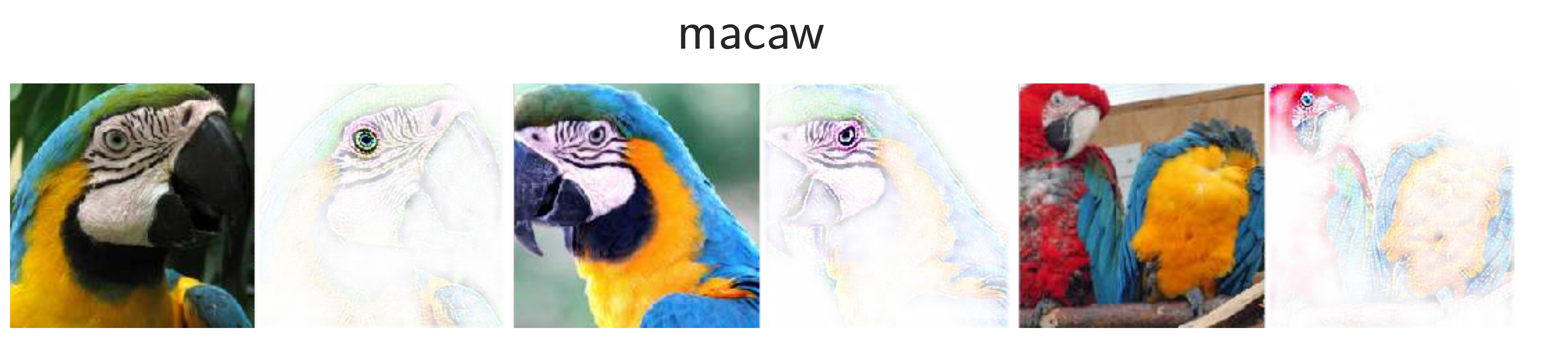}
        \end{subfigure}
    
        \begin{subfigure}[b]{.975\linewidth}\centering
        \includegraphics[width=\linewidth]{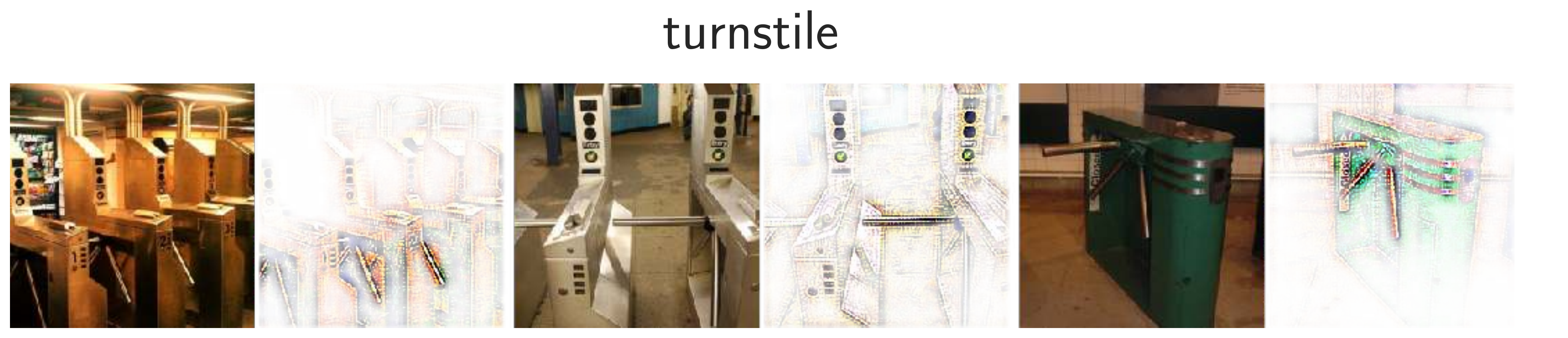}
        \end{subfigure}
    
        \begin{subfigure}[b]{.975\linewidth}\centering
        \includegraphics[width=\linewidth]{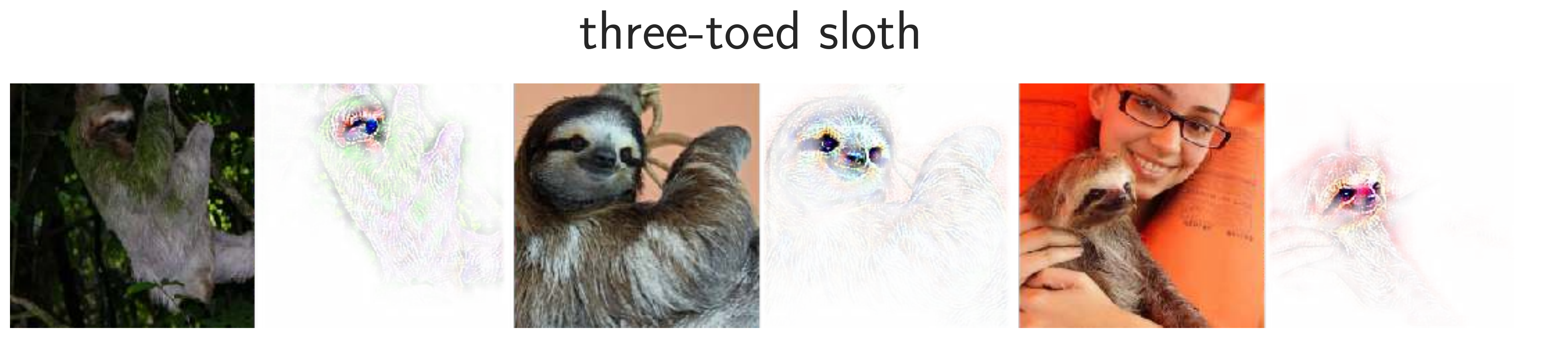}
        \end{subfigure}
    
        \begin{subfigure}[b]{.975\linewidth}\centering
        \includegraphics[width=\linewidth]{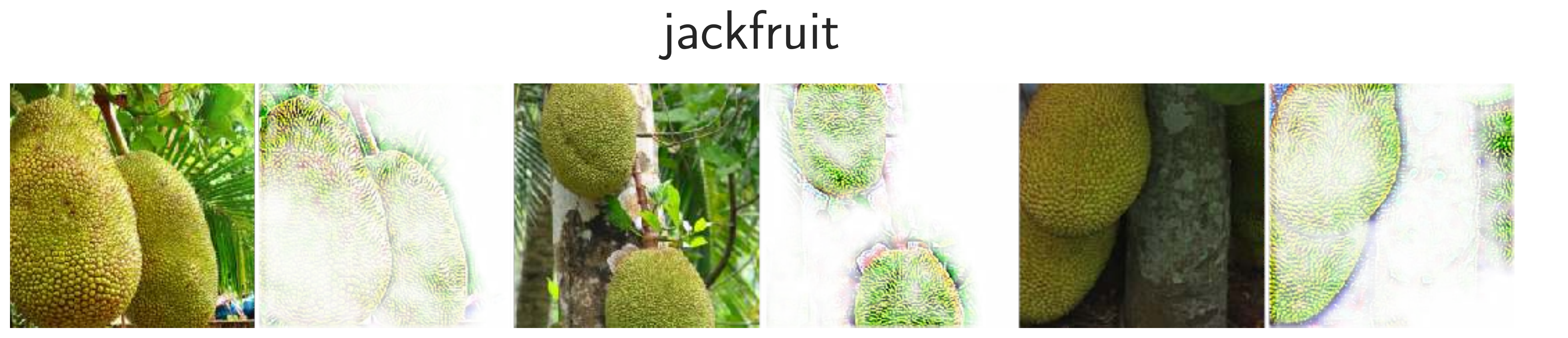}
        \end{subfigure}
    
        \begin{subfigure}[b]{.975\linewidth}\centering
        \includegraphics[width=\linewidth]{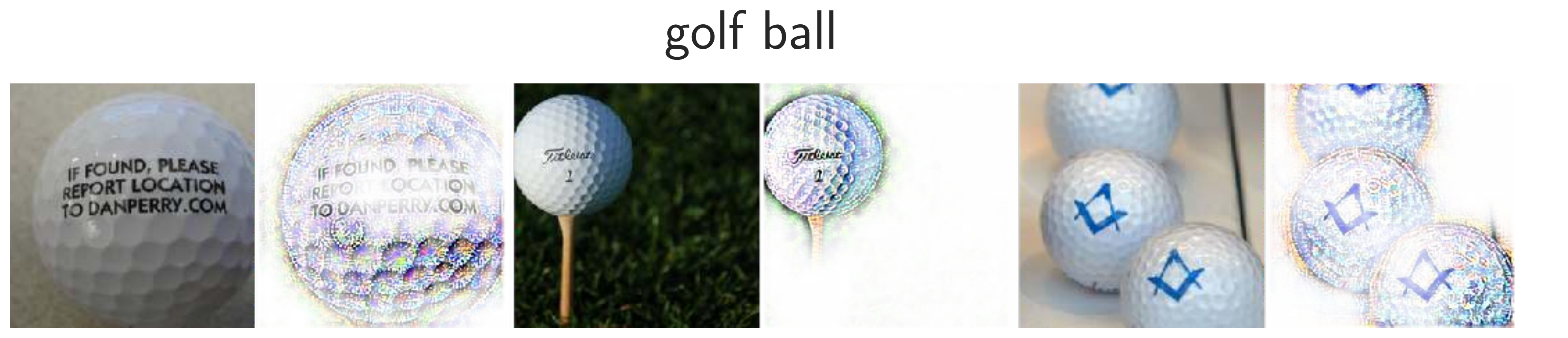}
        \end{subfigure}
    \end{subfigure}
    \begin{subfigure}[b]{.45\linewidth}\centering
        
        \begin{subfigure}[b]{.975\linewidth}\centering
        \includegraphics[width=\linewidth]{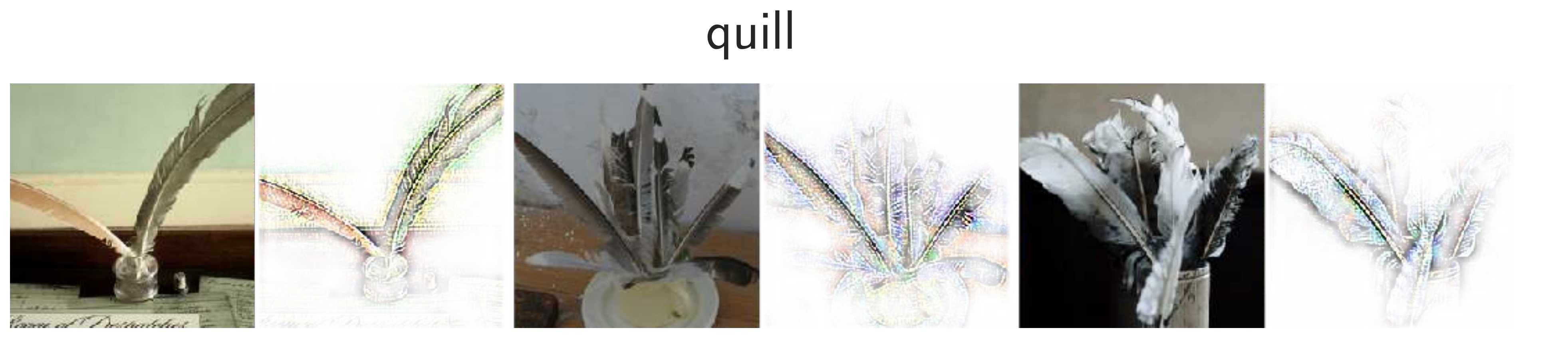}
        \end{subfigure}
    
        \begin{subfigure}[b]{.975\linewidth}\centering
        \includegraphics[width=\linewidth]{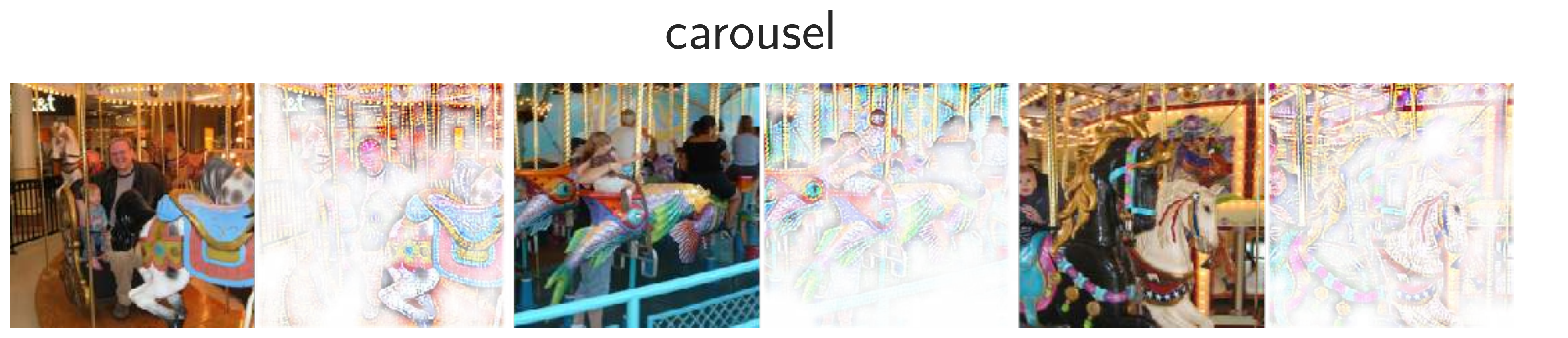}
        \end{subfigure}
    
        \begin{subfigure}[b]{.975\linewidth}\centering
        \includegraphics[width=\linewidth]{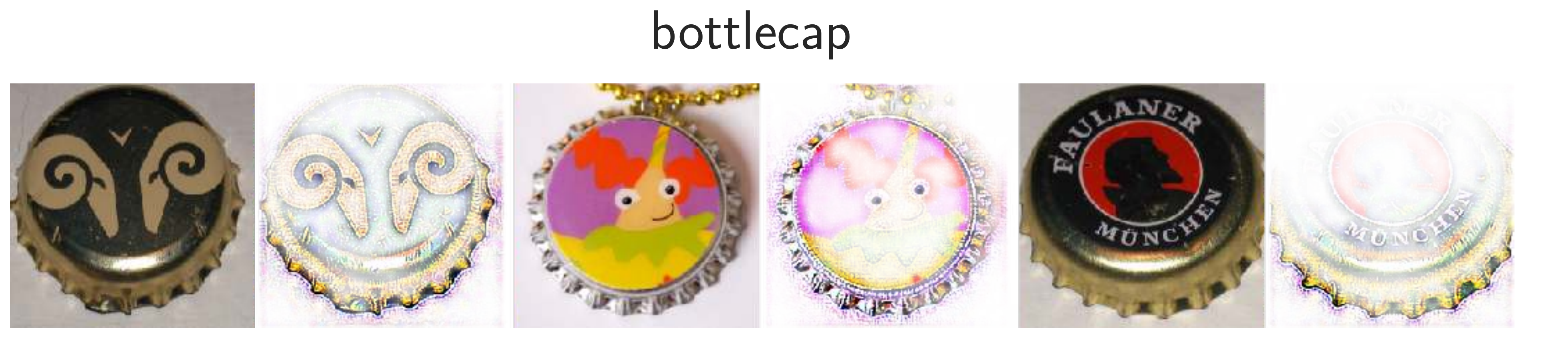}
        \end{subfigure}
    
        \begin{subfigure}[b]{.975\linewidth}\centering
        \includegraphics[width=\linewidth]{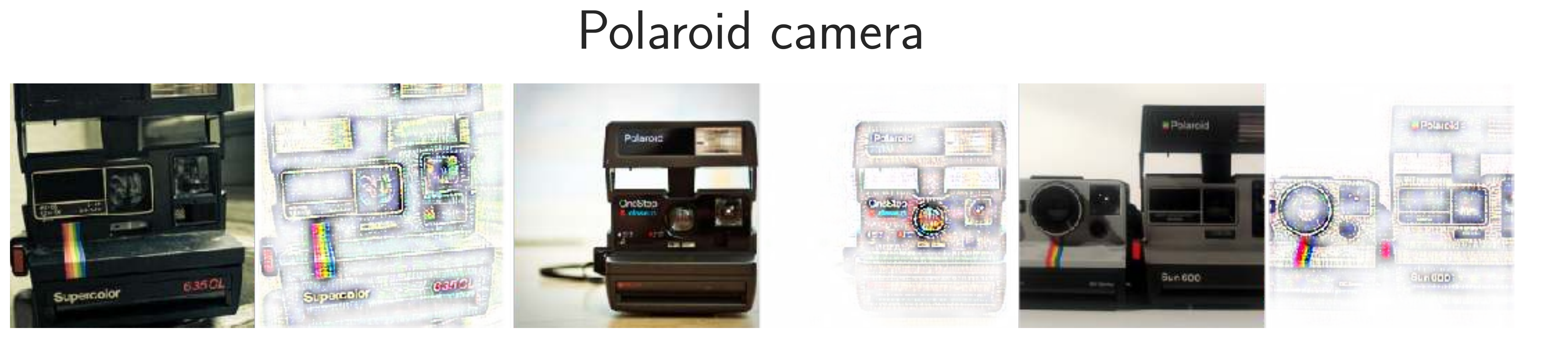}
        \end{subfigure}
    
        \begin{subfigure}[b]{.975\linewidth}\centering
        \includegraphics[width=\linewidth]{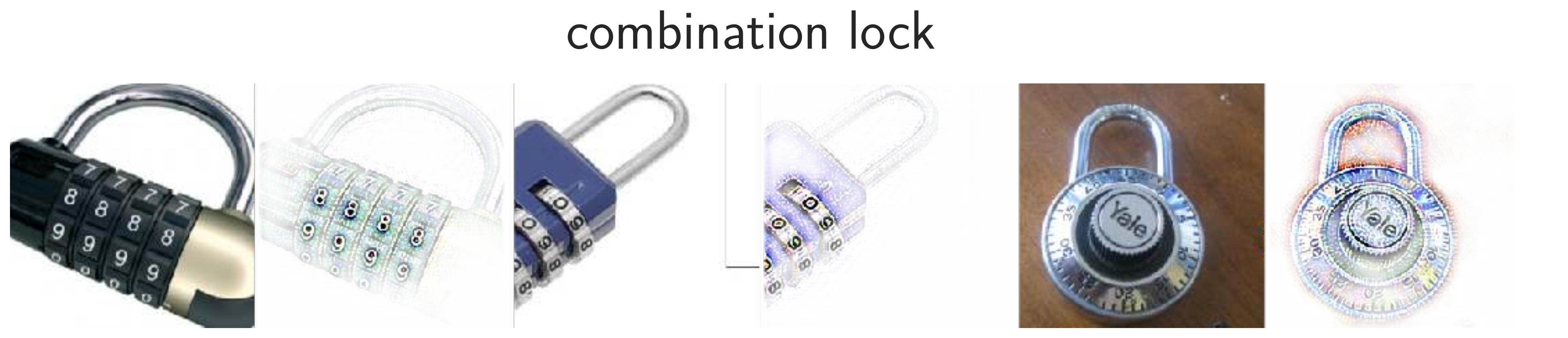}
        \end{subfigure}
    
        \begin{subfigure}[b]{.975\linewidth}\centering
        \includegraphics[width=\linewidth]{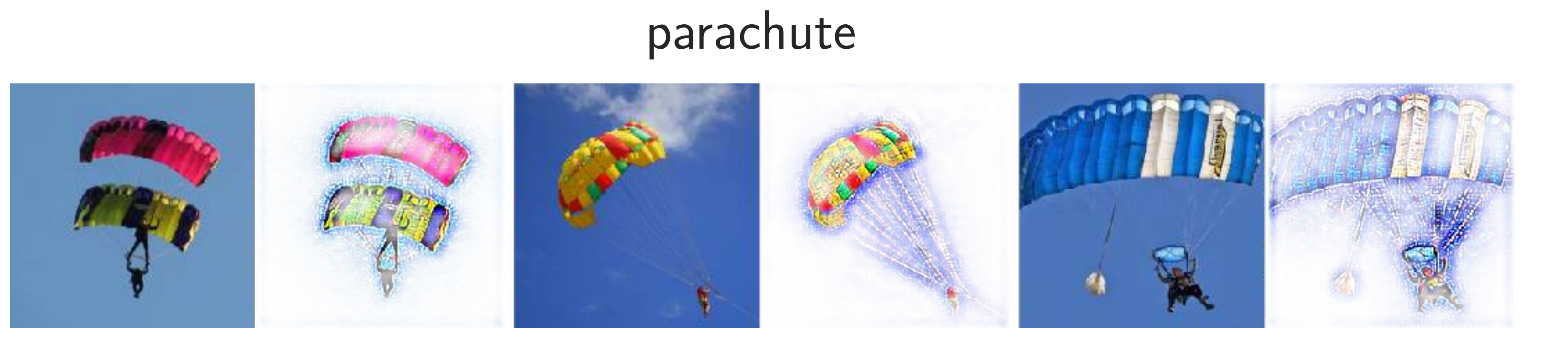}
        \end{subfigure}
    
        \begin{subfigure}[b]{.975\linewidth}\centering
        \includegraphics[width=\linewidth]{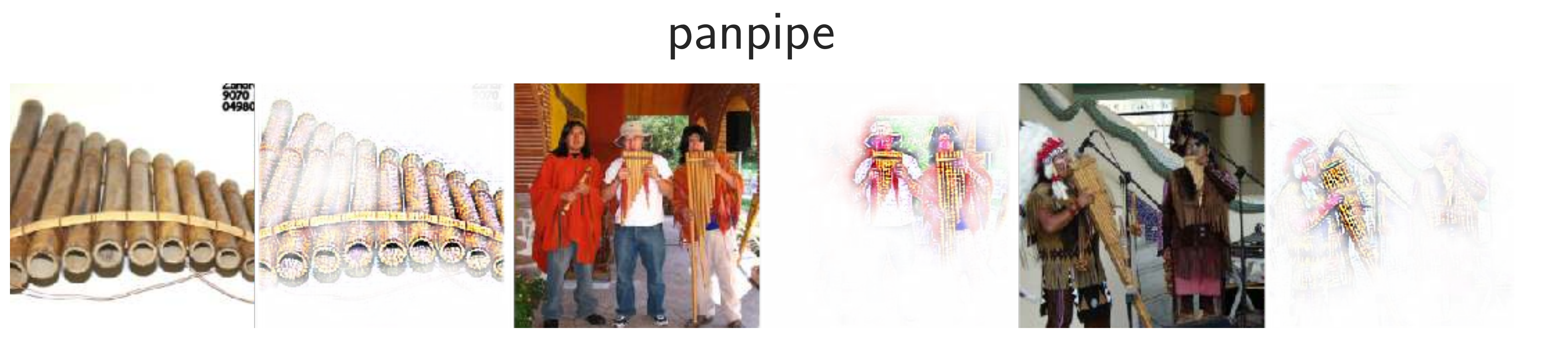}
        \end{subfigure}
    
        \begin{subfigure}[b]{.975\linewidth}\centering
        \includegraphics[width=\linewidth]{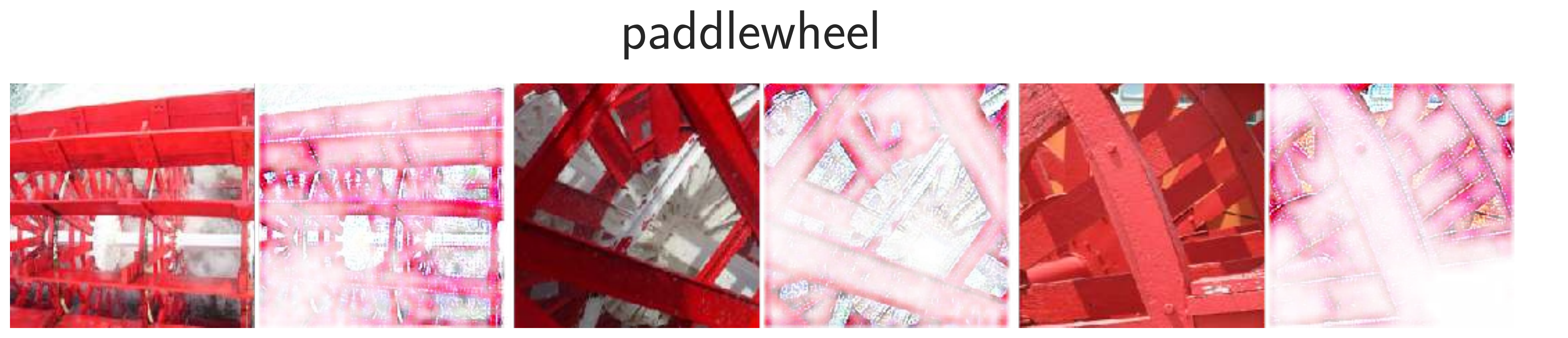}
        \end{subfigure}
    
        \begin{subfigure}[b]{.975\linewidth}\centering
        \includegraphics[width=\linewidth]{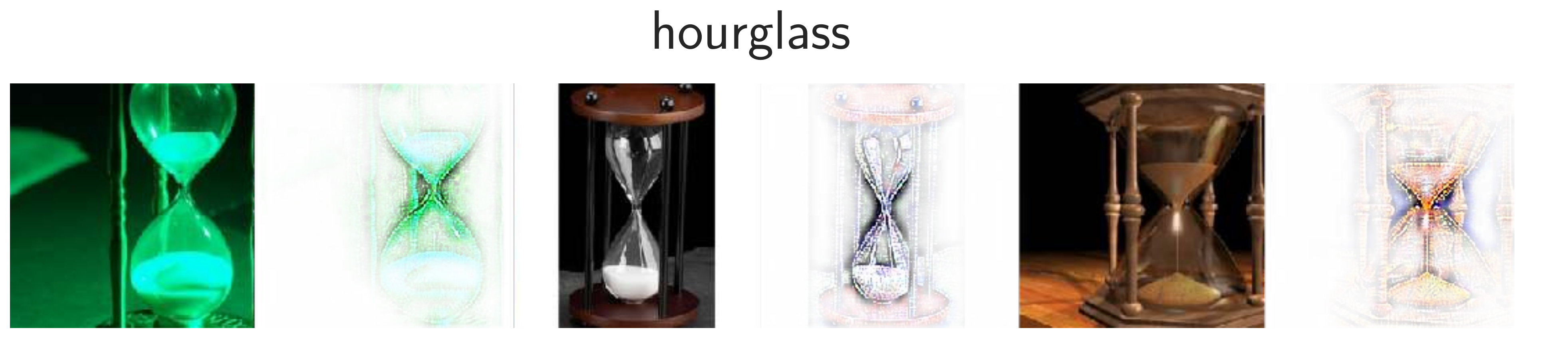}
        \end{subfigure}
    
        \begin{subfigure}[b]{.975\linewidth}\centering
        \includegraphics[width=\linewidth]{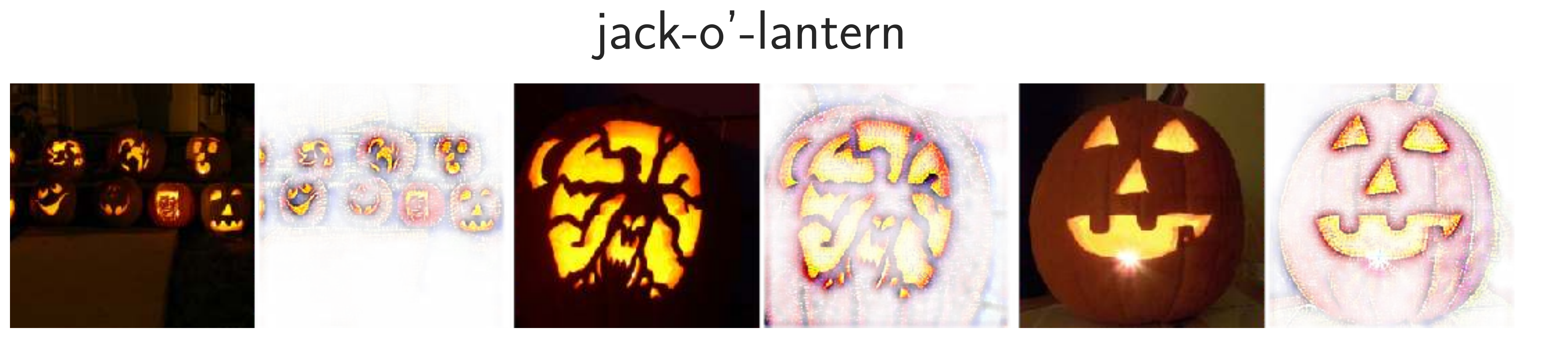}
        \end{subfigure}
    
        \begin{subfigure}[b]{.975\linewidth}\centering
        \includegraphics[width=\linewidth]{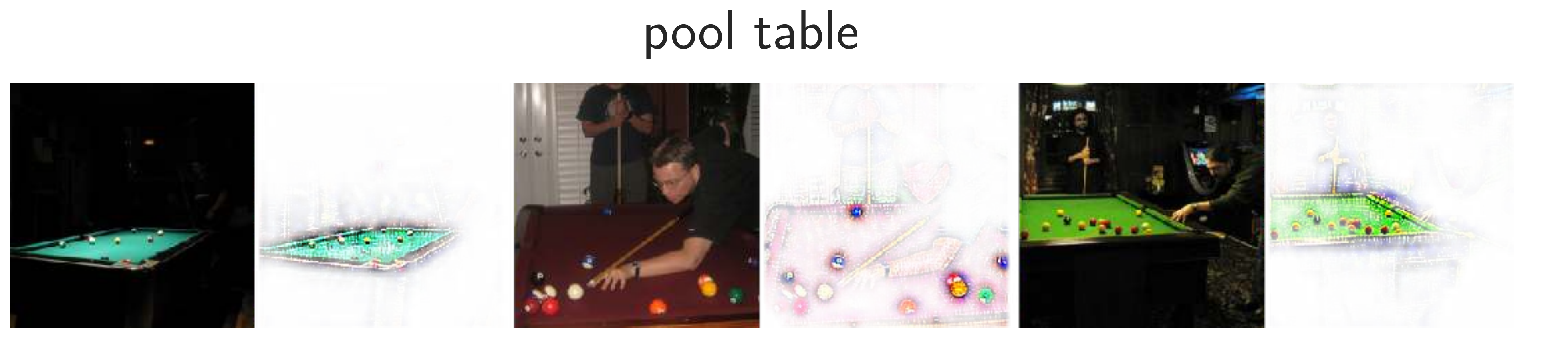}
        \end{subfigure}
    
        \begin{subfigure}[b]{.975\linewidth}\centering
        \includegraphics[width=\linewidth]{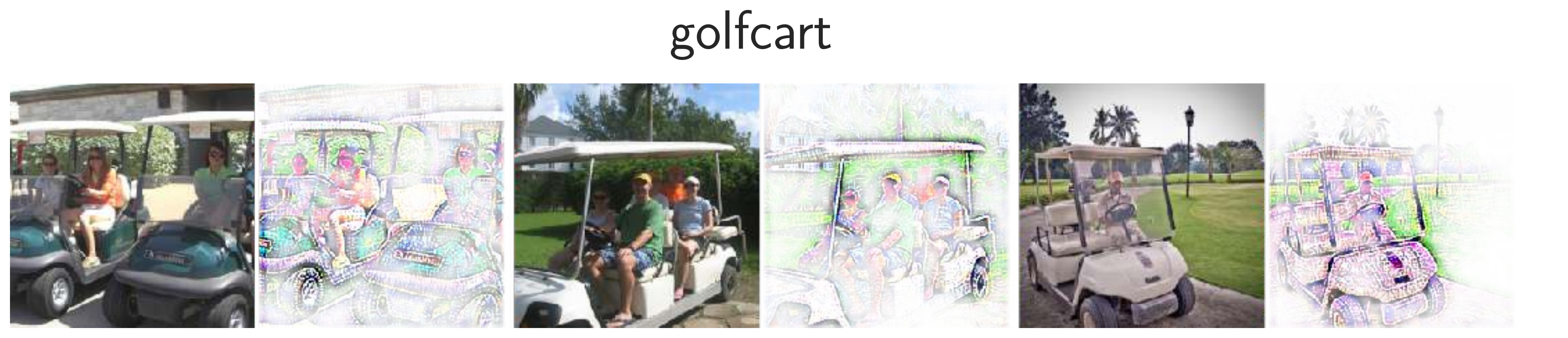}
        \end{subfigure}
    \end{subfigure}
    
    \caption{\hspace{-.15em}First three samples $\vec x_i^c$ and linear mappings $[\mat{w}_{1\rightarrow L}(\vec x_i^c)]_c $ for 24 of the most confidently classified classes $c$ from the Imagenet dataset. Specifically, the classes are sorted by the sum of the logits for those three samples. Left: Classes 25-36. Right: Classes 37-48.}
    \label{fig:add_quali_2}
\end{figure}

%% file: supplement/resources/figures/add_layer7.tex
\begin{figure}[h]
    \centering
    \begin{subfigure}[b]{.49\linewidth}
    \centering
        \begin{subfigure}[b]{1\linewidth}\centering
        \includegraphics[width=\linewidth]{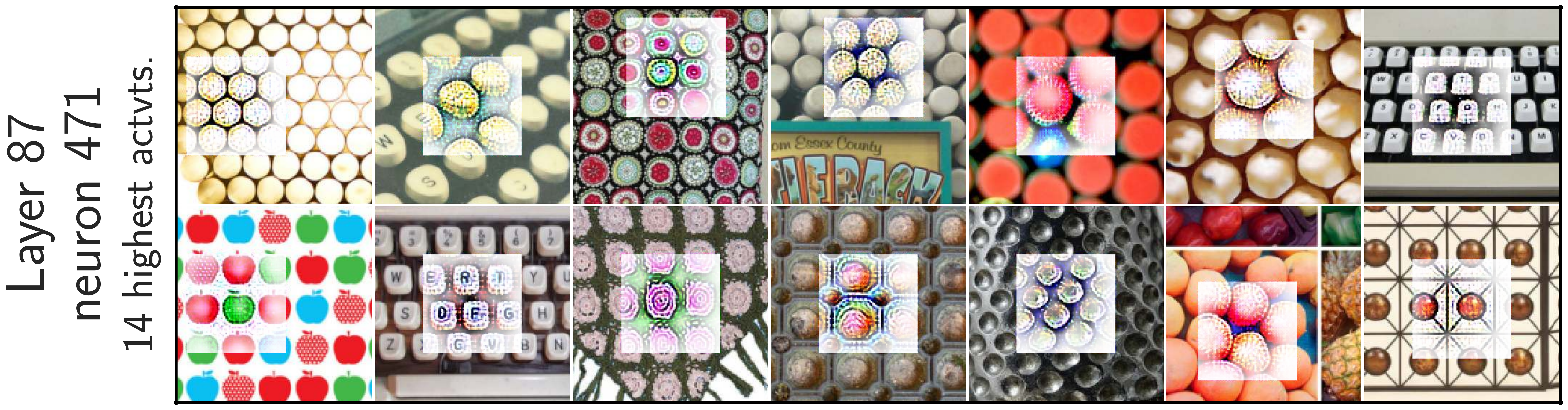}
        \end{subfigure}
        \begin{subfigure}[b]{1\linewidth}\centering
        \includegraphics[width=\linewidth]{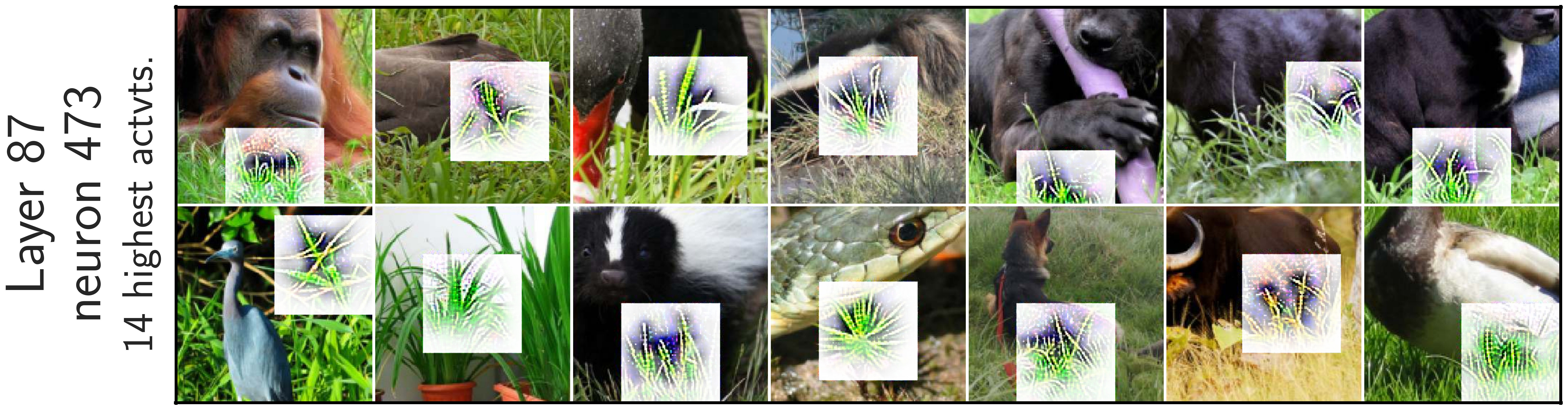}
        \end{subfigure}
    \end{subfigure}
    \begin{subfigure}[b]{.49\linewidth}
    \centering
    \begin{subfigure}[b]{1\linewidth}\centering
        \includegraphics[width=\linewidth]{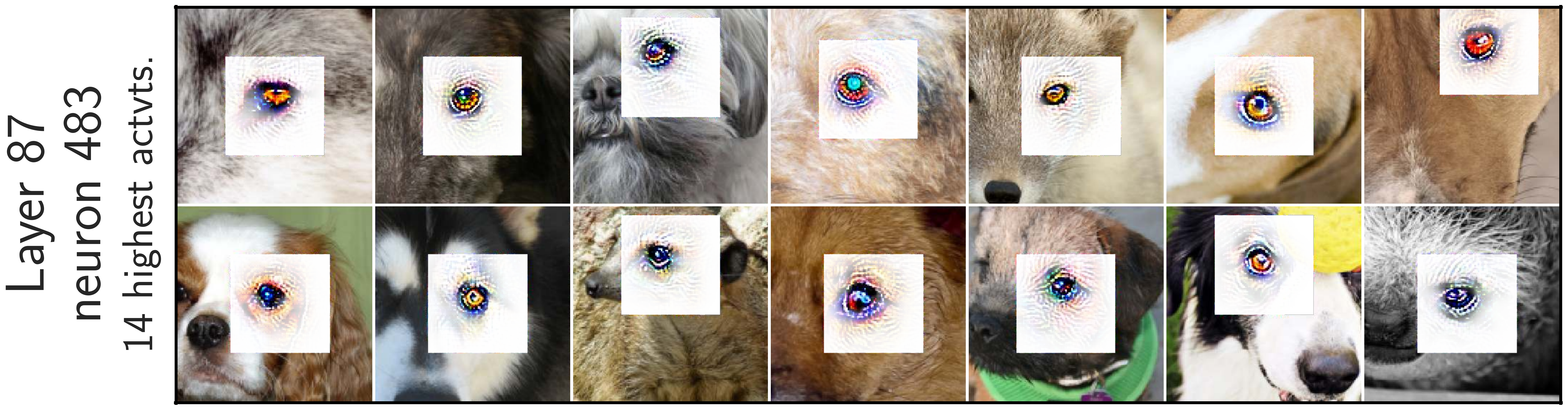}
        \end{subfigure}
        \begin{subfigure}[b]{1\linewidth}\centering
        \includegraphics[width=\linewidth]{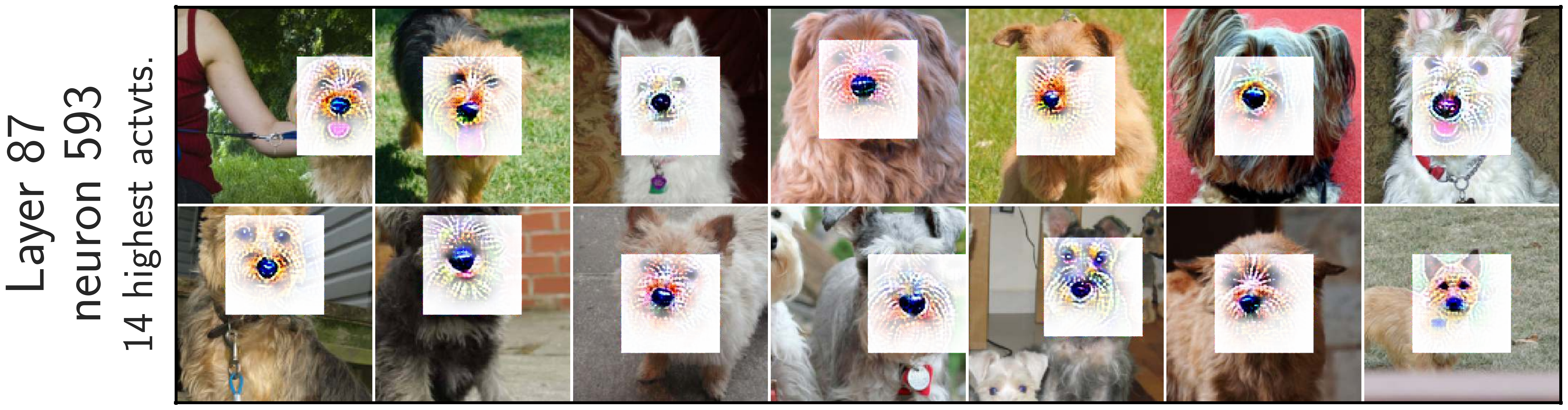}
        \end{subfigure}
    \end{subfigure}
    \begin{subfigure}[b]{.49\linewidth}
    \centering
        \begin{subfigure}[b]{1\linewidth}\centering
        \includegraphics[width=\linewidth]{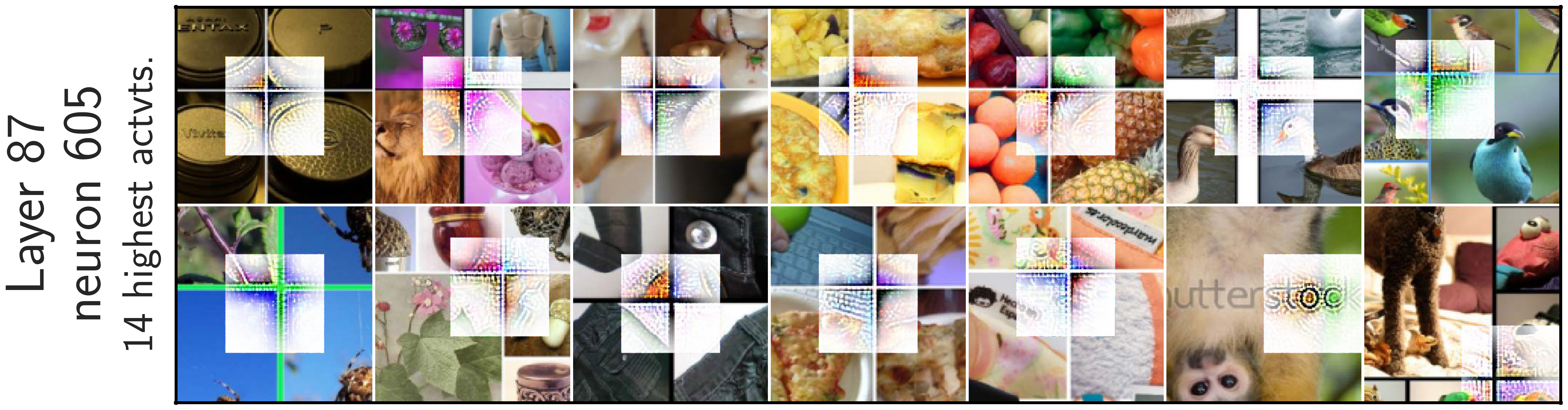}
        \end{subfigure}
        \begin{subfigure}[b]{1\linewidth}\centering
        \includegraphics[width=\linewidth]{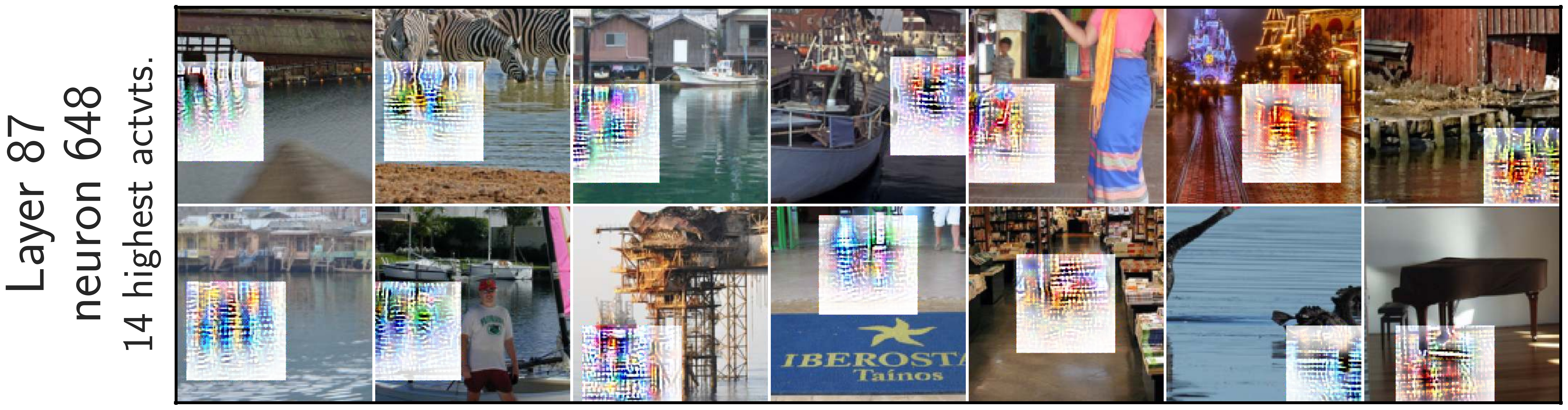}
        \end{subfigure}
        \begin{subfigure}[b]{1\linewidth}\centering
        \includegraphics[width=\linewidth]{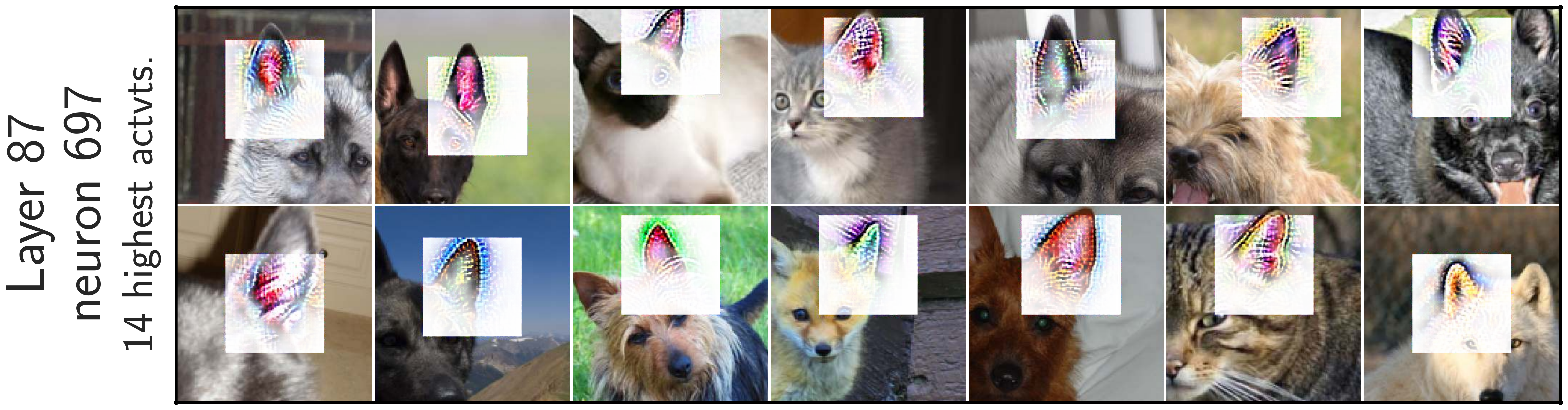}
        \end{subfigure}
        \begin{subfigure}[b]{1\linewidth}\centering
        \includegraphics[width=\linewidth]{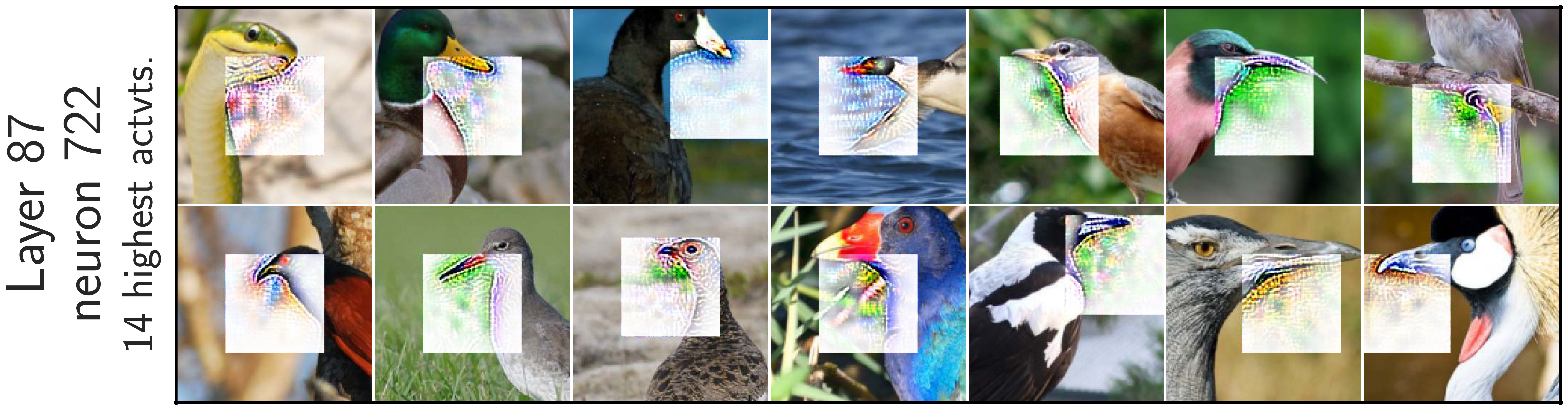}
        \end{subfigure}
                \begin{subfigure}[b]{1\linewidth}\centering
        \includegraphics[width=\linewidth]{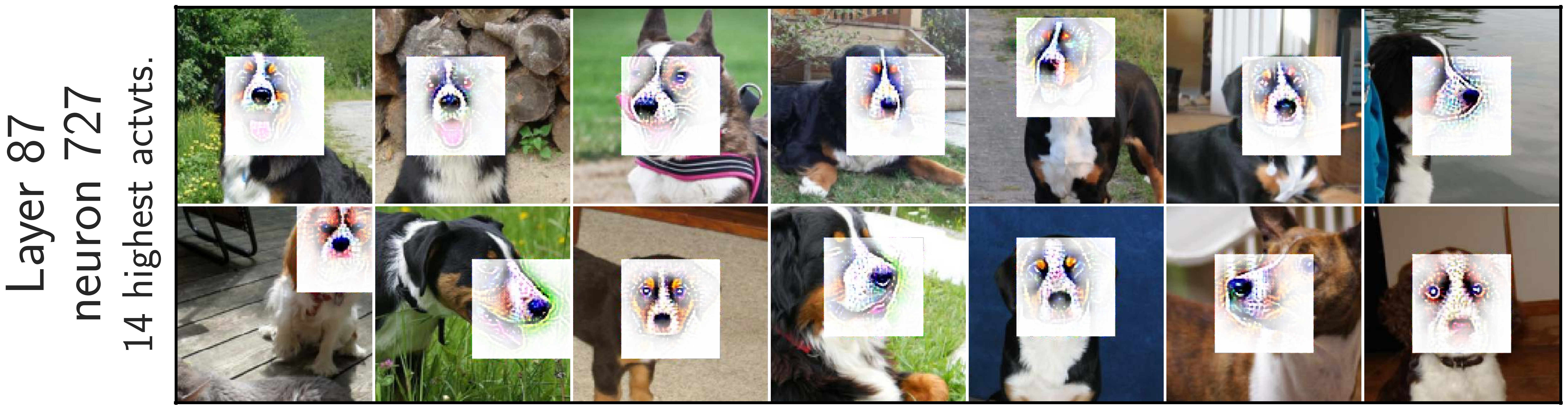}
        \end{subfigure}
        \begin{subfigure}[b]{1\linewidth}\centering
        \includegraphics[width=\linewidth]{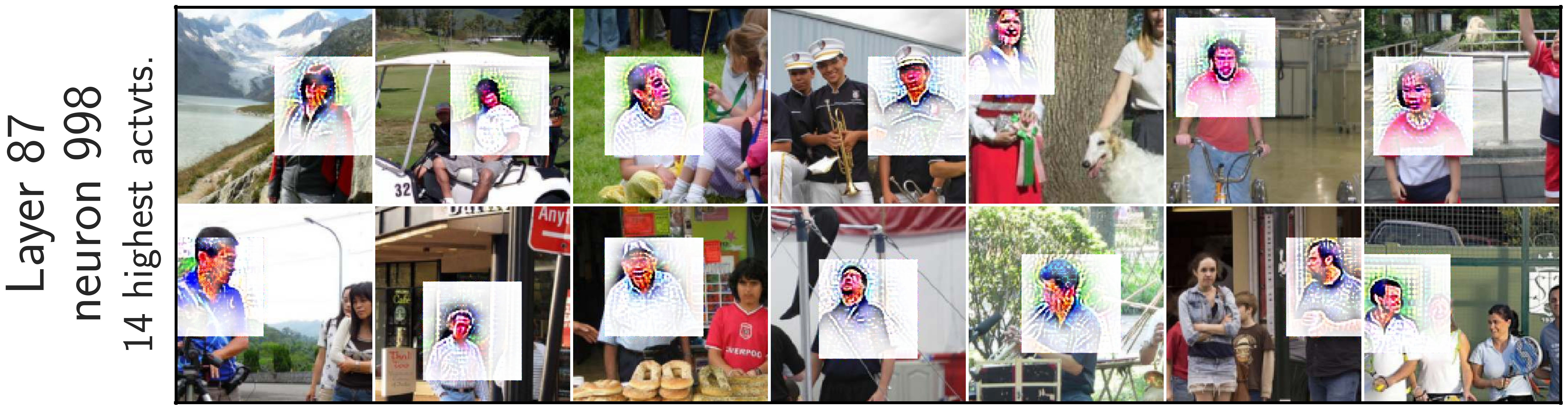}
        \end{subfigure}
    \end{subfigure}
    \begin{subfigure}[b]{.49\linewidth}
    \centering
        \begin{subfigure}[b]{1\linewidth}\centering
        \includegraphics[width=\linewidth]{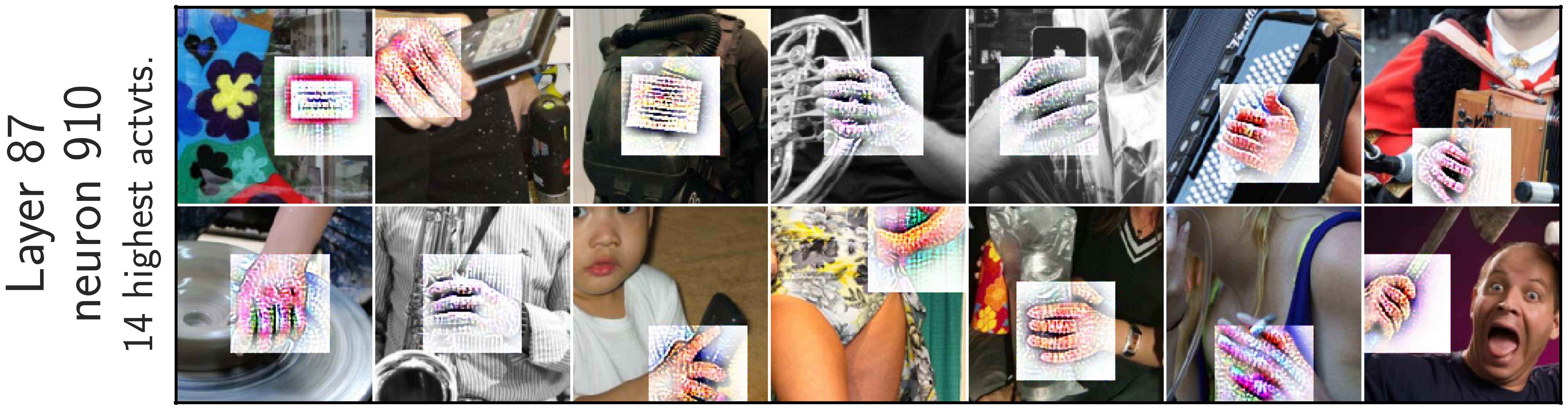}
        \end{subfigure}
        \begin{subfigure}[b]{1\linewidth}\centering
        \includegraphics[width=\linewidth]{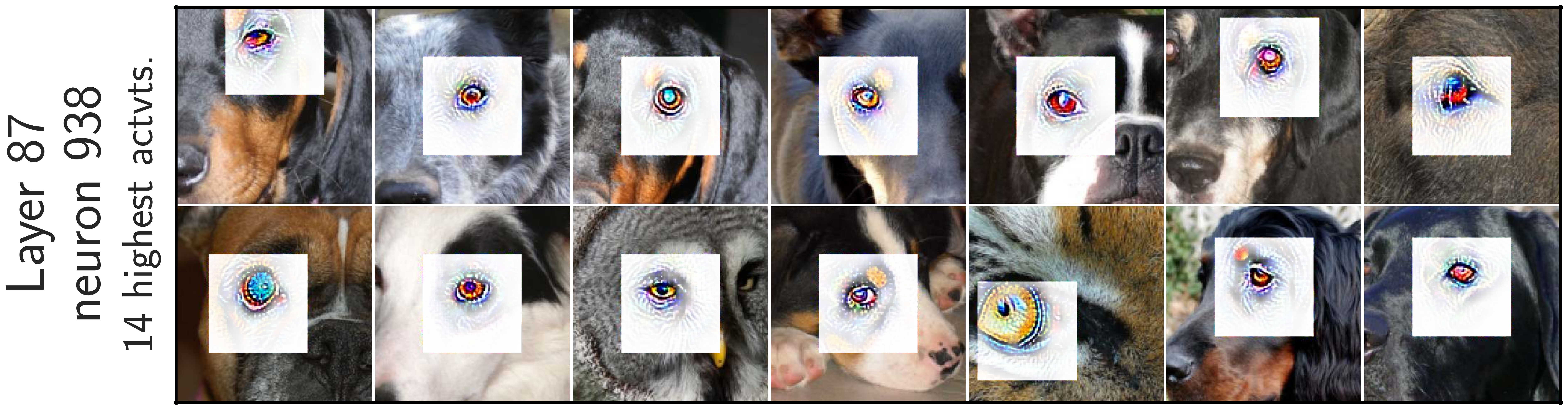}
        \end{subfigure}
        \begin{subfigure}[b]{1\linewidth}\centering
        \includegraphics[width=\linewidth]{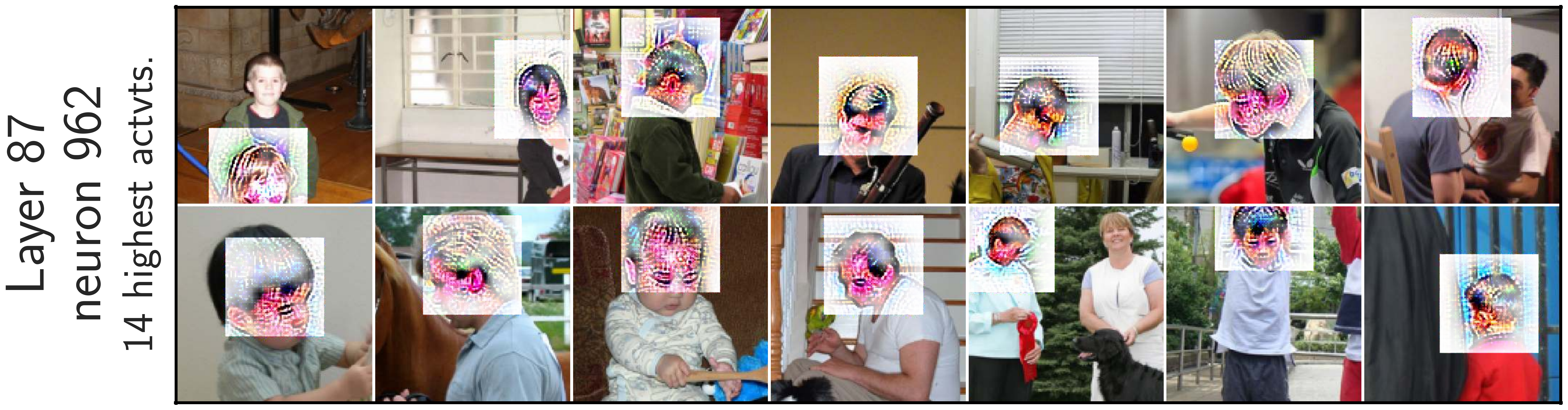}
        \end{subfigure}
        \begin{subfigure}[b]{1\linewidth}\centering
        \includegraphics[width=\linewidth]{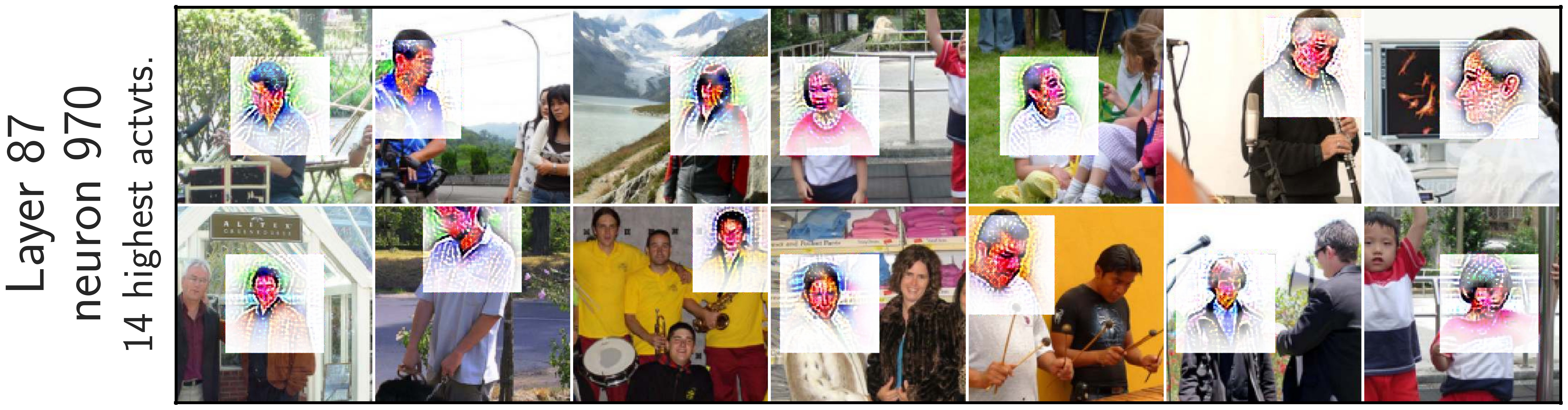}
        \end{subfigure}
        \begin{subfigure}[b]{1\linewidth}\centering
        \includegraphics[width=\linewidth]{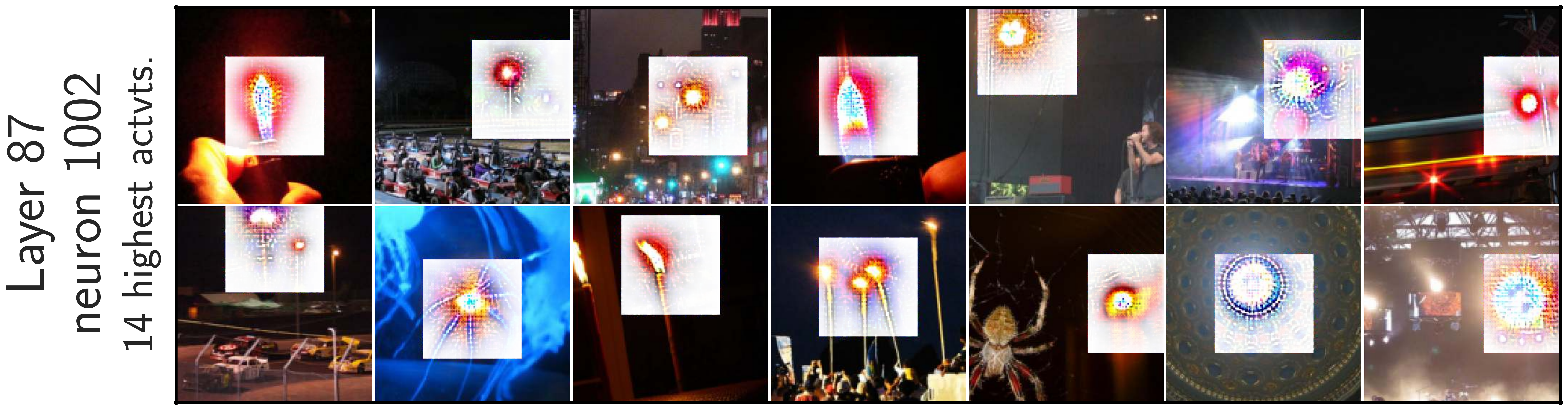}
        \end{subfigure}
        \begin{subfigure}[b]{1\linewidth}\centering
        \includegraphics[width=\linewidth]{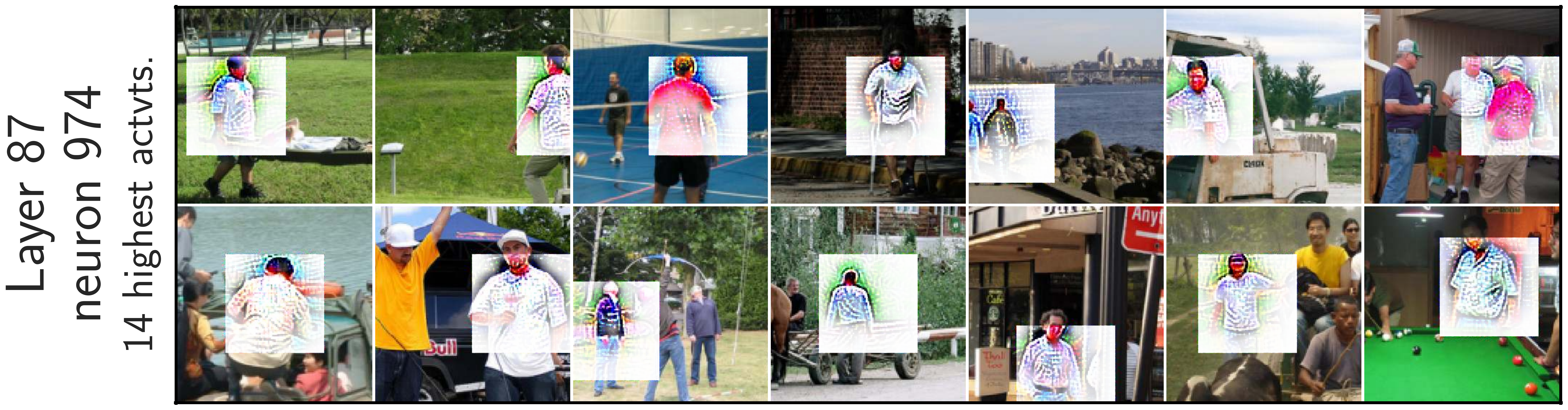}
        \end{subfigure}
    \end{subfigure}
    \caption{\hspace{-.15em}Additional examples of some of the 20 most highly activating neurons in layer 87 of the \bcos{} DenseNet-121 model. Similar to
    the results shown in the main paper, we observe the neurons to represent highly specific concepts.}
    \label{fig:add_layer7}
\end{figure}

%% file: supplement/resources/figures/add_networks.tex
\begin{figure}[h]
    \centering
    \begin{subfigure}[b]{1\linewidth}
    \centering
        \begin{subfigure}[b]{.32\linewidth}\centering
        \includegraphics[width=\linewidth]{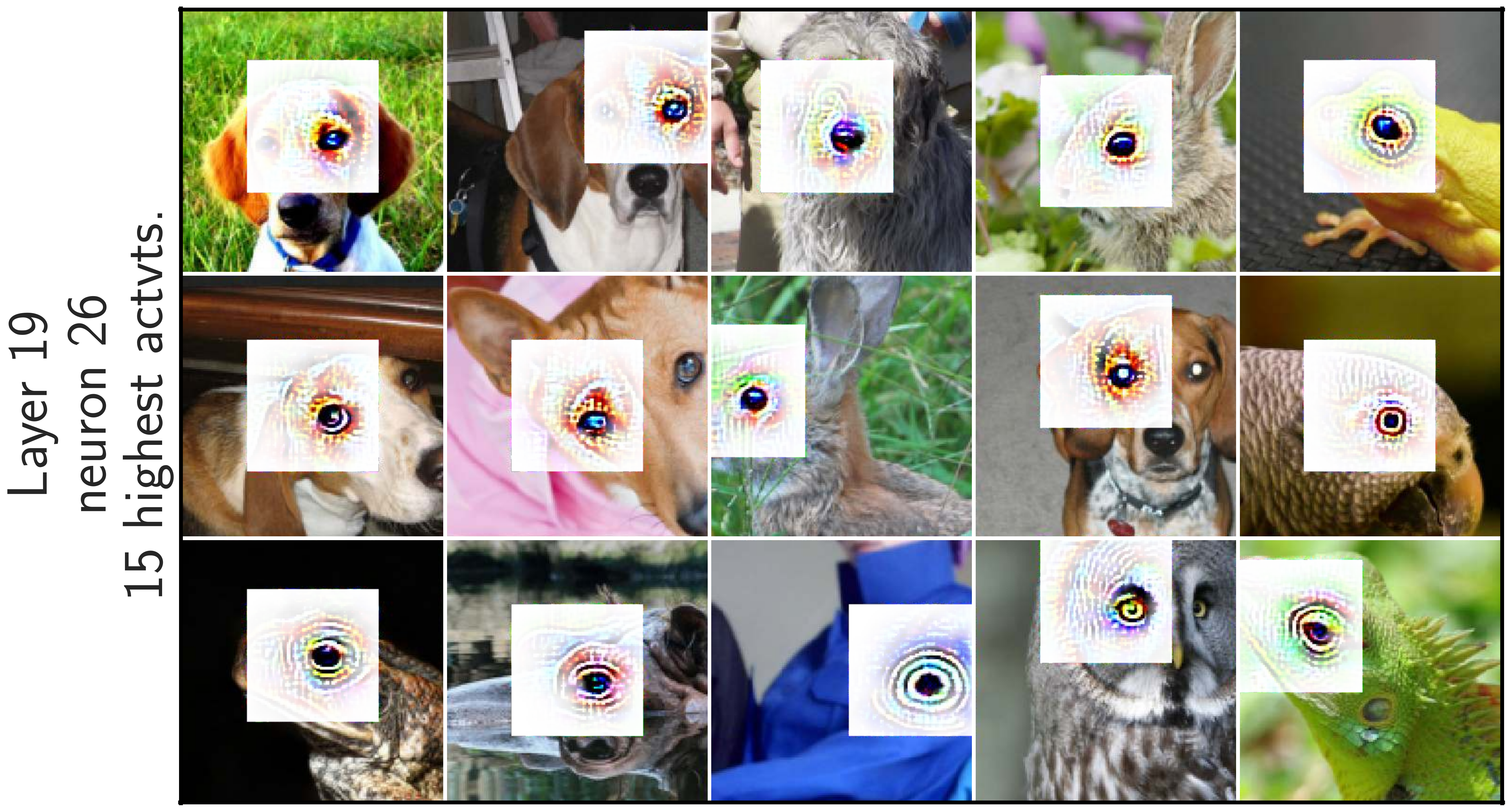}
        \end{subfigure}
        \begin{subfigure}[b]{.32\linewidth}\centering
        \includegraphics[width=\linewidth]{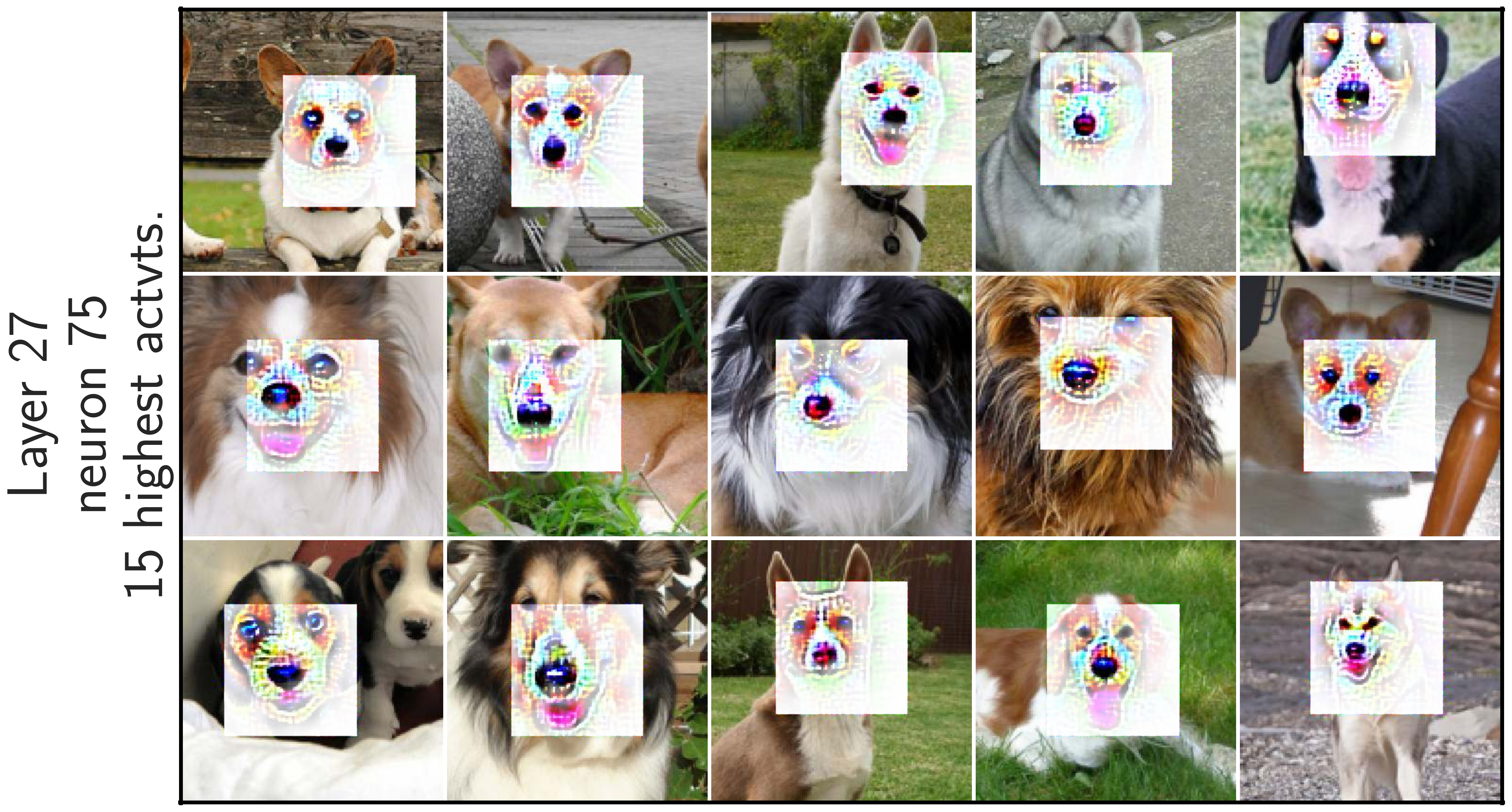}
        \end{subfigure}
        \begin{subfigure}[b]{.32\linewidth}\centering
        \includegraphics[width=\linewidth]{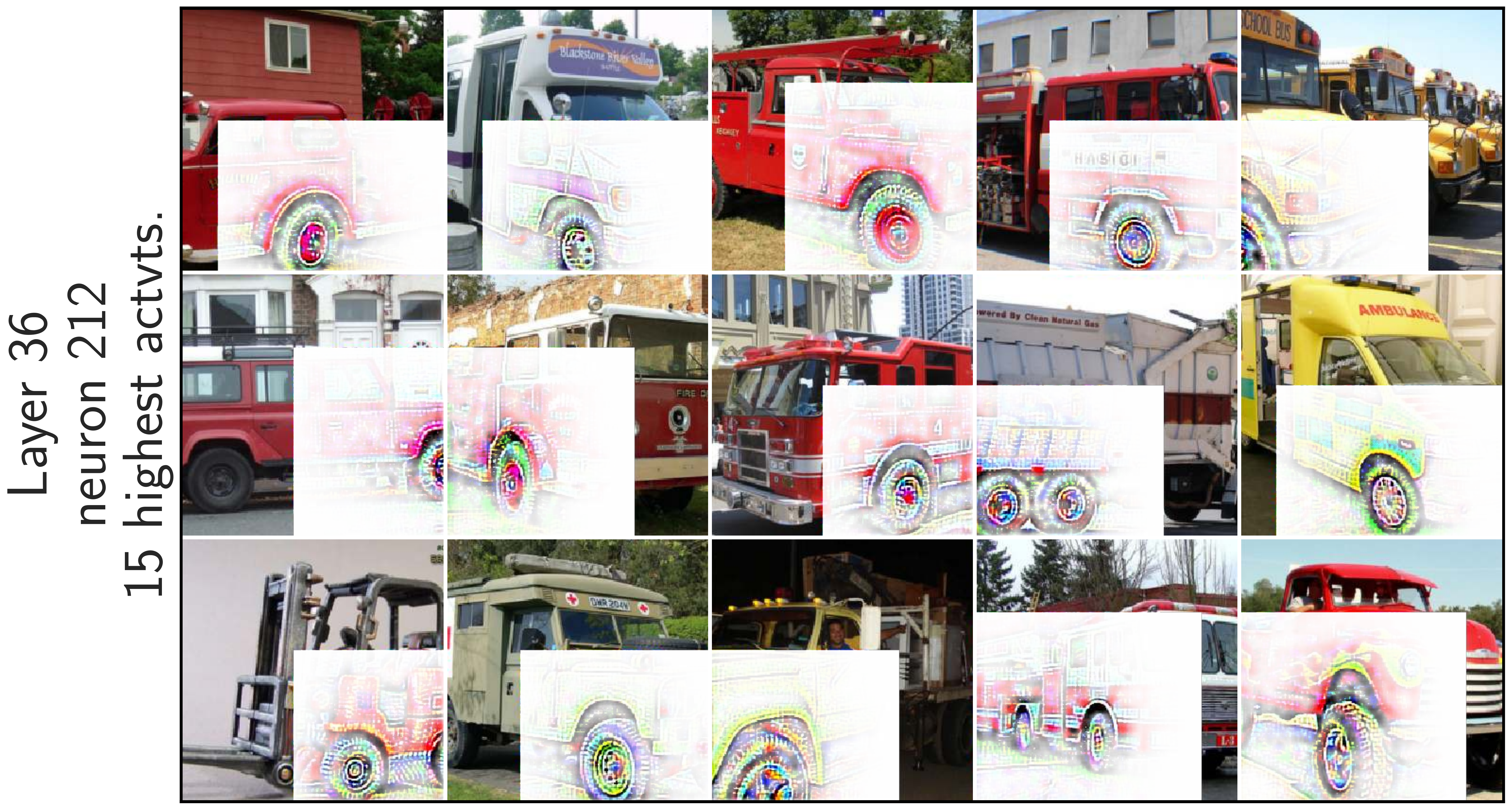}
        \end{subfigure}
        \caption{\bcos{} ResNet-34}
        \vspace{2em}
    \end{subfigure}
    \begin{subfigure}[b]{1\linewidth}
    \centering
        \begin{subfigure}[b]{.32\linewidth}\centering
        \includegraphics[width=\linewidth]{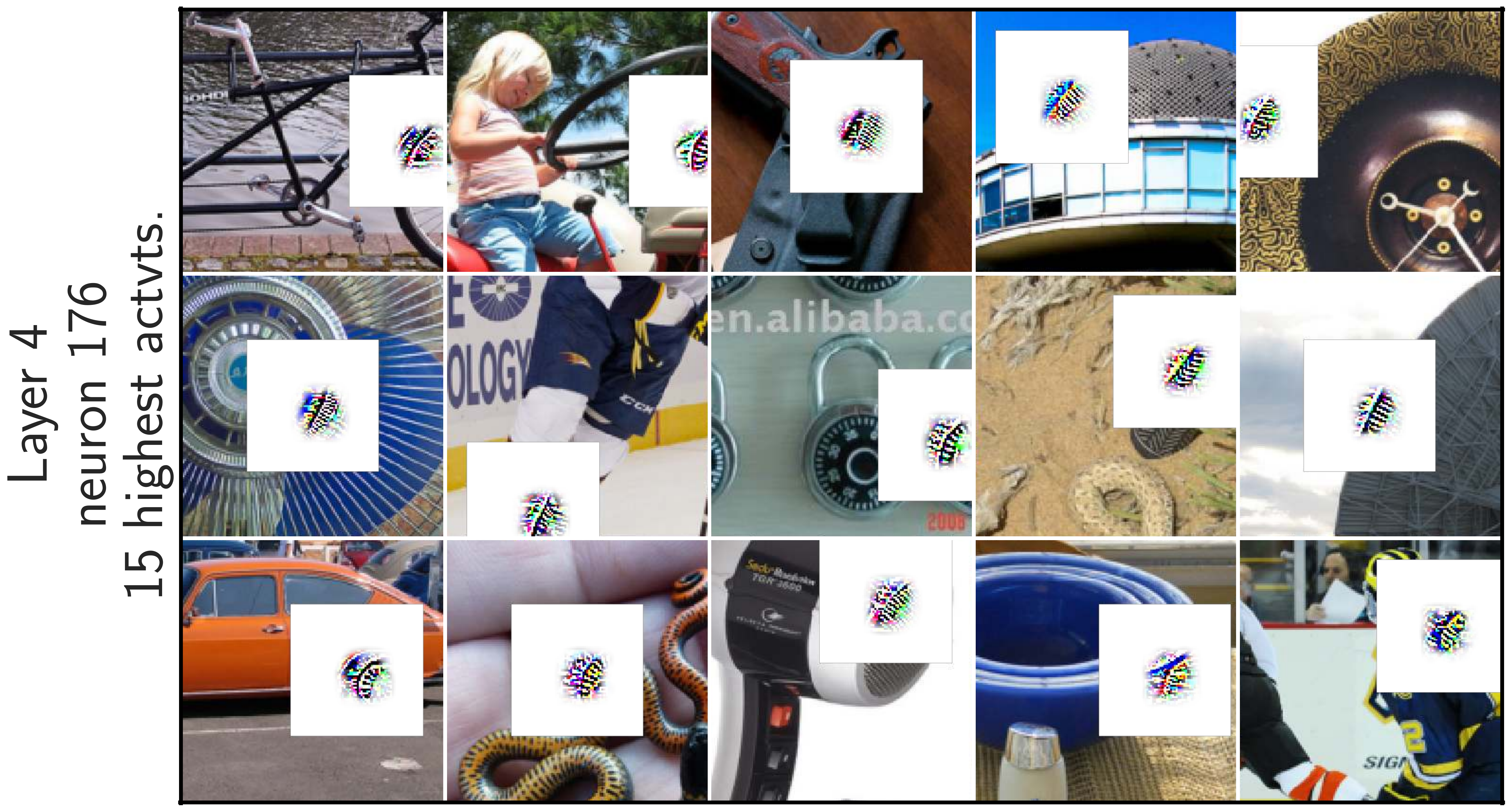}
        \end{subfigure}
        \begin{subfigure}[b]{.32\linewidth}\centering
        \includegraphics[width=\linewidth]{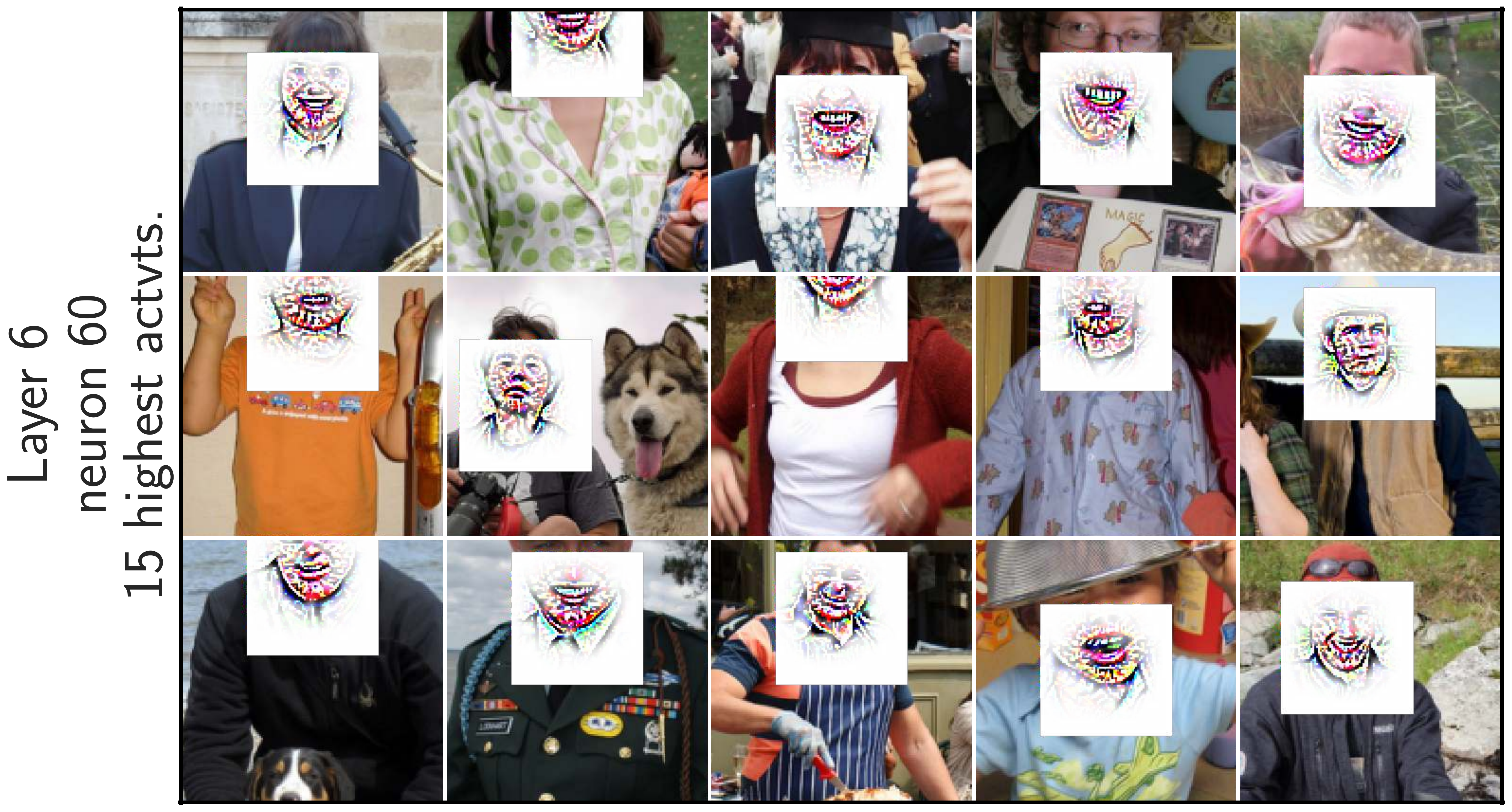}
        \end{subfigure}
        \begin{subfigure}[b]{.32\linewidth}\centering
        \includegraphics[width=\linewidth]{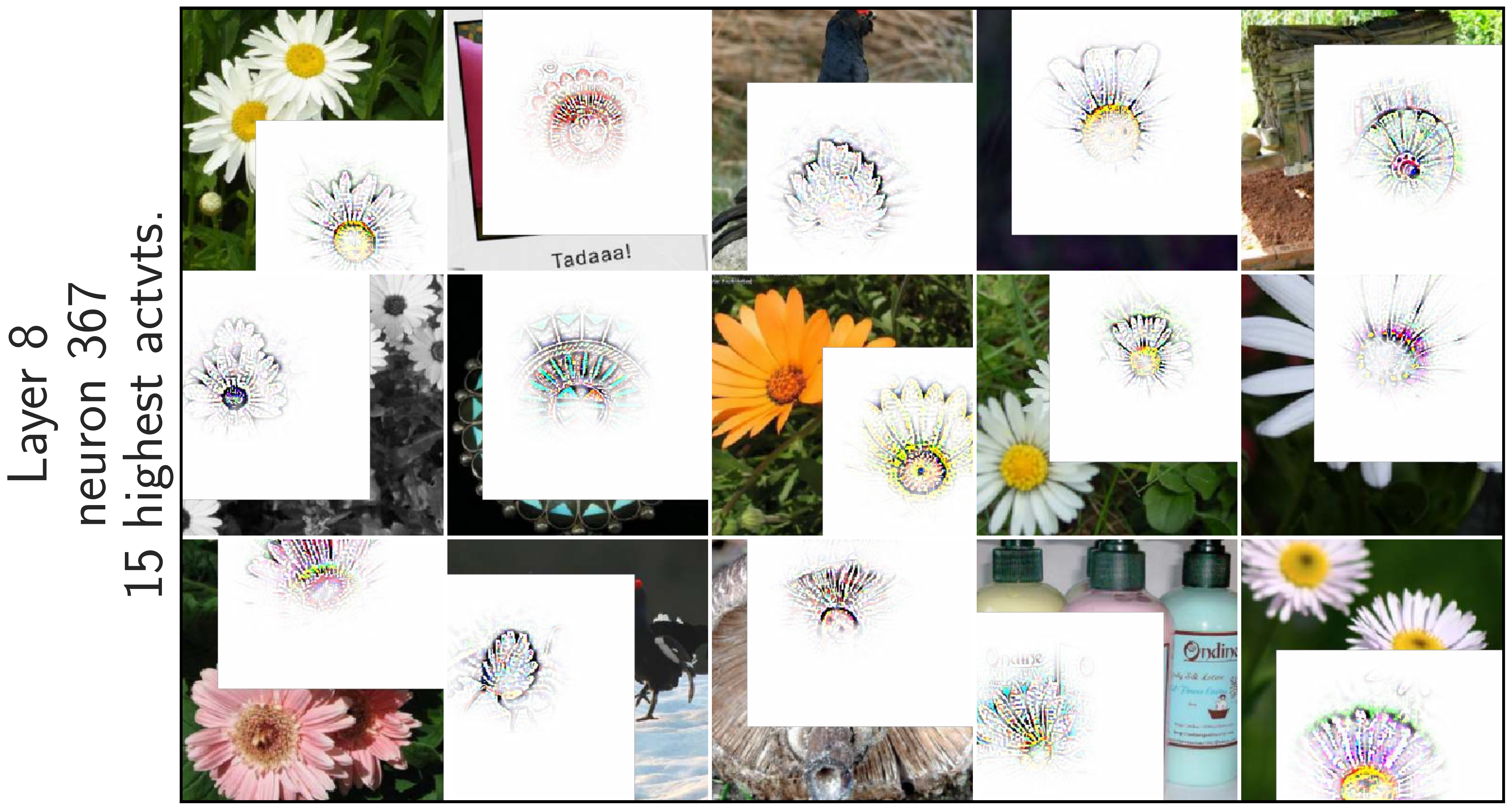}
        \end{subfigure}
        \caption{\bcos{} VGG-11}
                \vspace{2em}
    \end{subfigure}
    \begin{subfigure}[b]{1\linewidth}
    \centering
        \begin{subfigure}[b]{.32\linewidth}\centering
        \includegraphics[width=\linewidth]{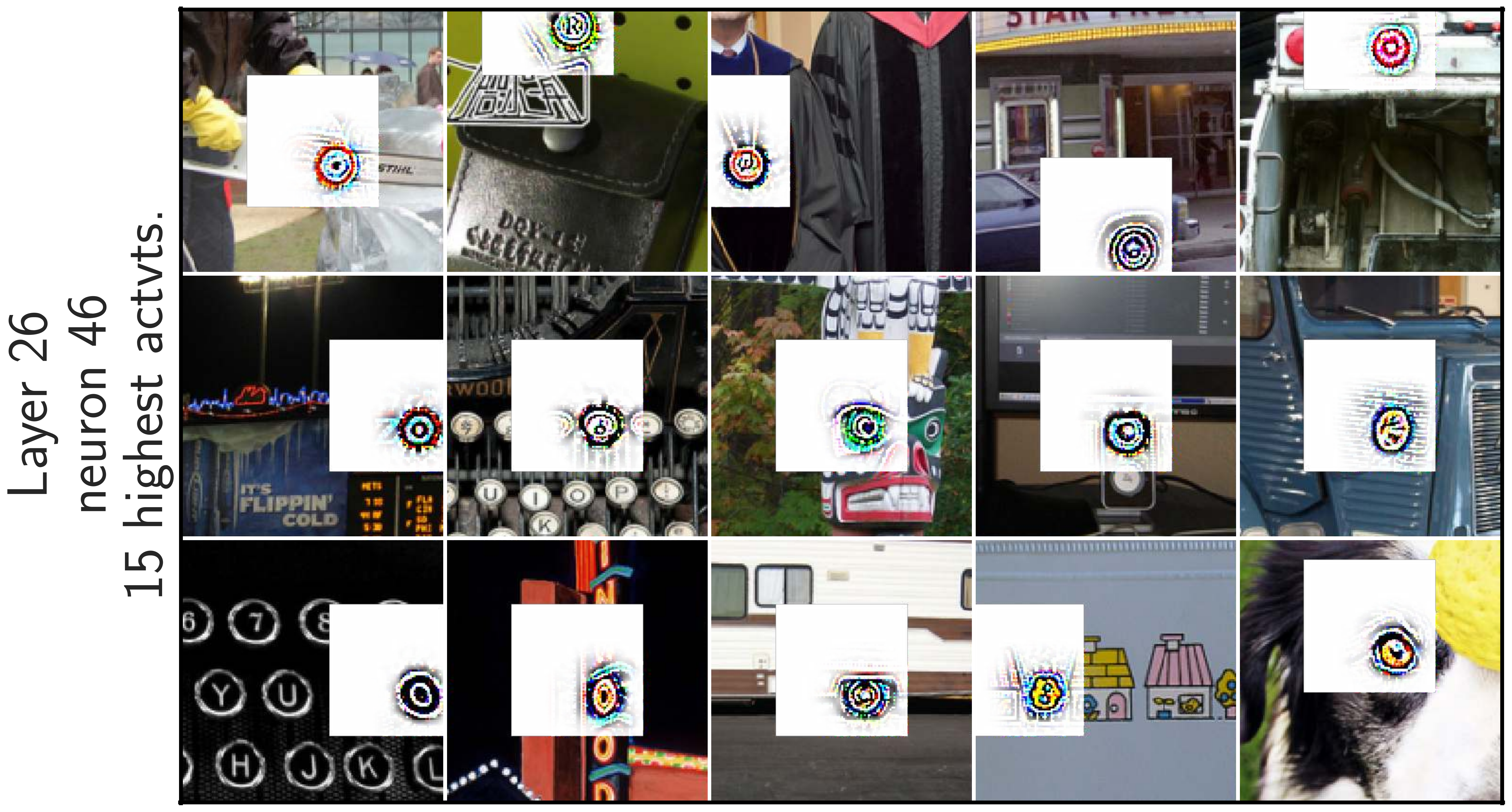}
        \end{subfigure}
        \begin{subfigure}[b]{.32\linewidth}\centering
        \includegraphics[width=\linewidth]{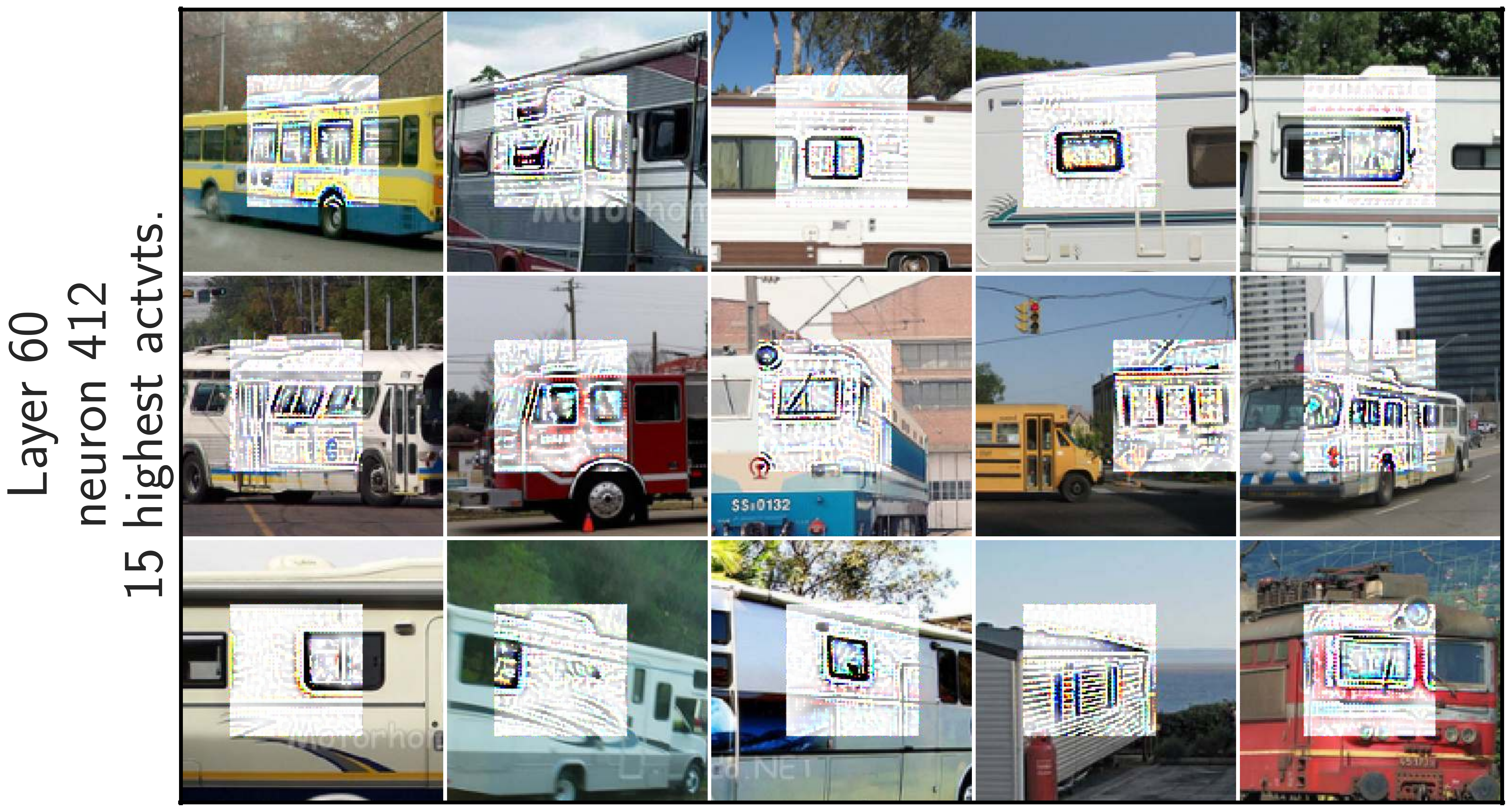}
        \end{subfigure}
        \begin{subfigure}[b]{.32\linewidth}\centering
        \includegraphics[width=\linewidth]{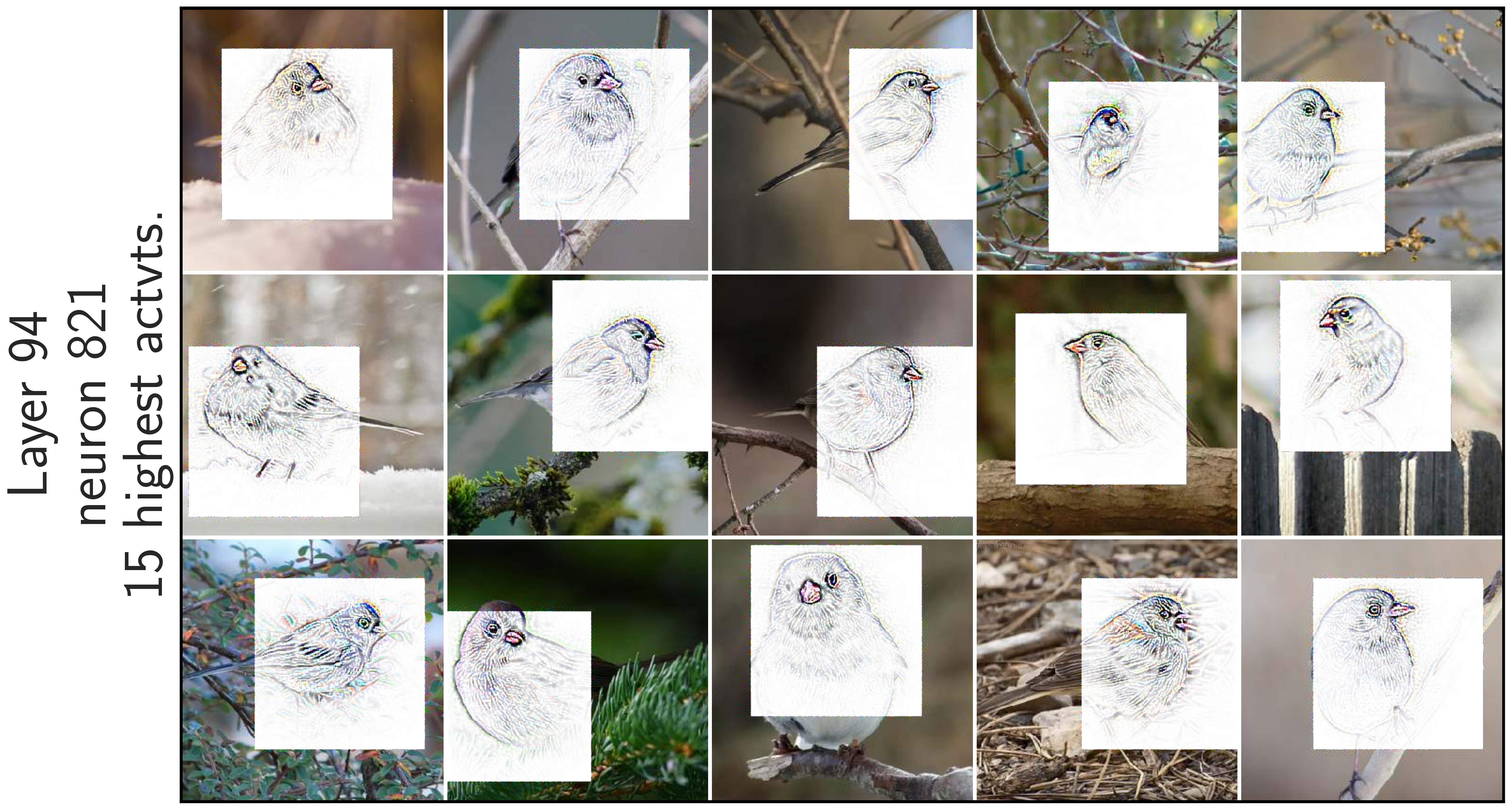}
        \end{subfigure}
        \caption{\bcos{} InceptionNet}
    \end{subfigure}
    \caption{\hspace{-.15em}Explanations for intermediate neurons for other \bcos{} networks. In particular, we show
    results for \bcos{} ResNet-34 (a), \bcos{} VGG-11 (b), and \bcos{} InceptionNet (c); cf.~Tab.~2 in the main paper.
    Similarly to the DenseNet-121 model, we observe the linear mappings $[\mat w_{1\rightarrow l}]_n$ to be of high visual quality and increase in complexity throughout the layers for all networks.
    }
    \label{fig:add_networks}
\end{figure}

%% file: supplement/resources/figures/add_quanti.tex
\begin{figure}[h]
    \centering
    \begin{subfigure}[b]{.975\linewidth}\centering
        \includegraphics[width=\linewidth]{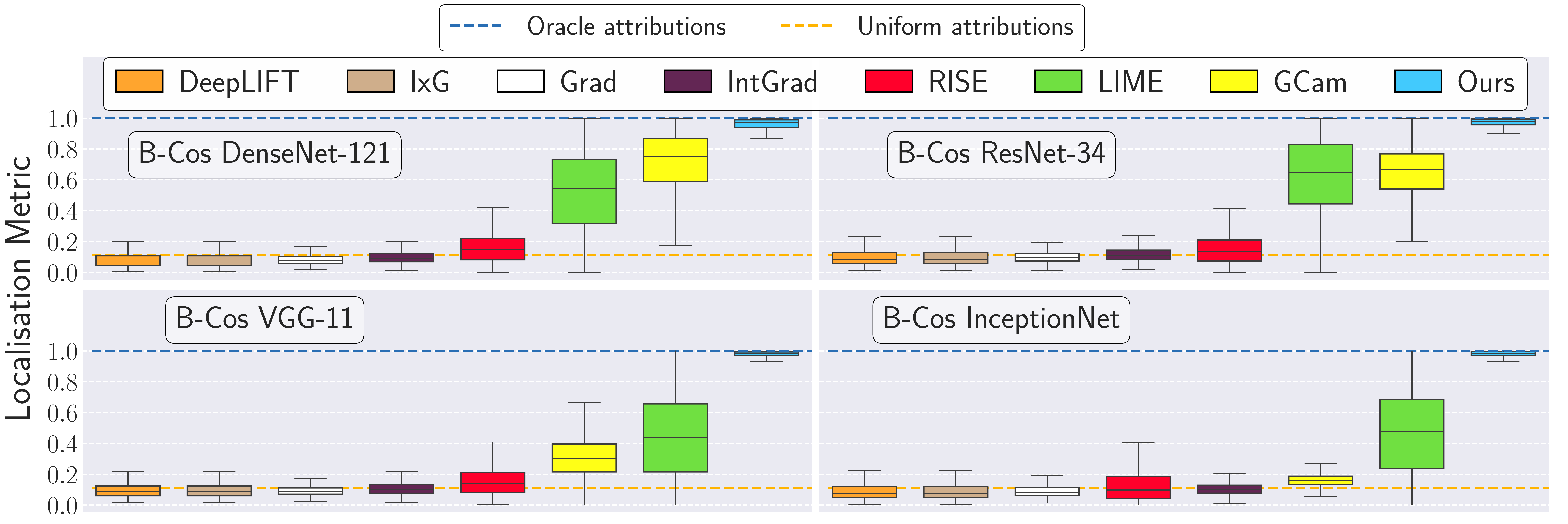}
    \end{subfigure}
    \caption{Localisation metric results for all attribution methods for the converted \bcos{} models. Note that the DenseNet-121 and InceptionNet results are the same as in the main paper in Fig.~5.}
    \label{fig:add_quanti_bcos}
\end{figure}

\begin{figure}[h]
    \centering
    \begin{subfigure}[b]{.975\linewidth}\centering
        \includegraphics[width=\linewidth]{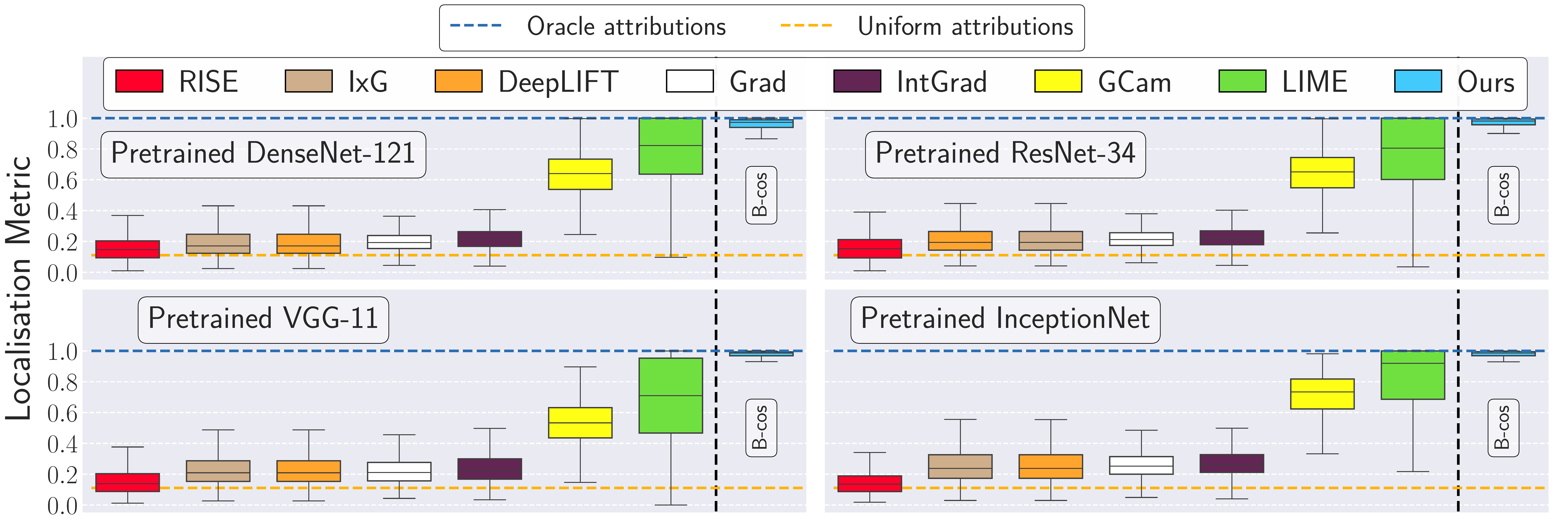}
    \end{subfigure}
    \caption{Results of the localisation metric for all post-hoc attribution methods for the original, pre-trained models. Additionally, we show the \bcos{} results of the converted models as a reference; note that the equivalent to `Ours' in piece-wise linear models is given by `IxG'. }
    \label{fig:add_quanti_pre}
\end{figure}

%% file: supplement/resources/figures/param_ablation.tex
{\begin{center}
\begin{adjustbox}{minipage=.45\linewidth, scale=1}
    \centering
    \vspace{1em}
    \captionsetup{type=figure}
    \centering
          \begin{subfigure}[b]{\linewidth}
            \includegraphics[width=\textwidth]{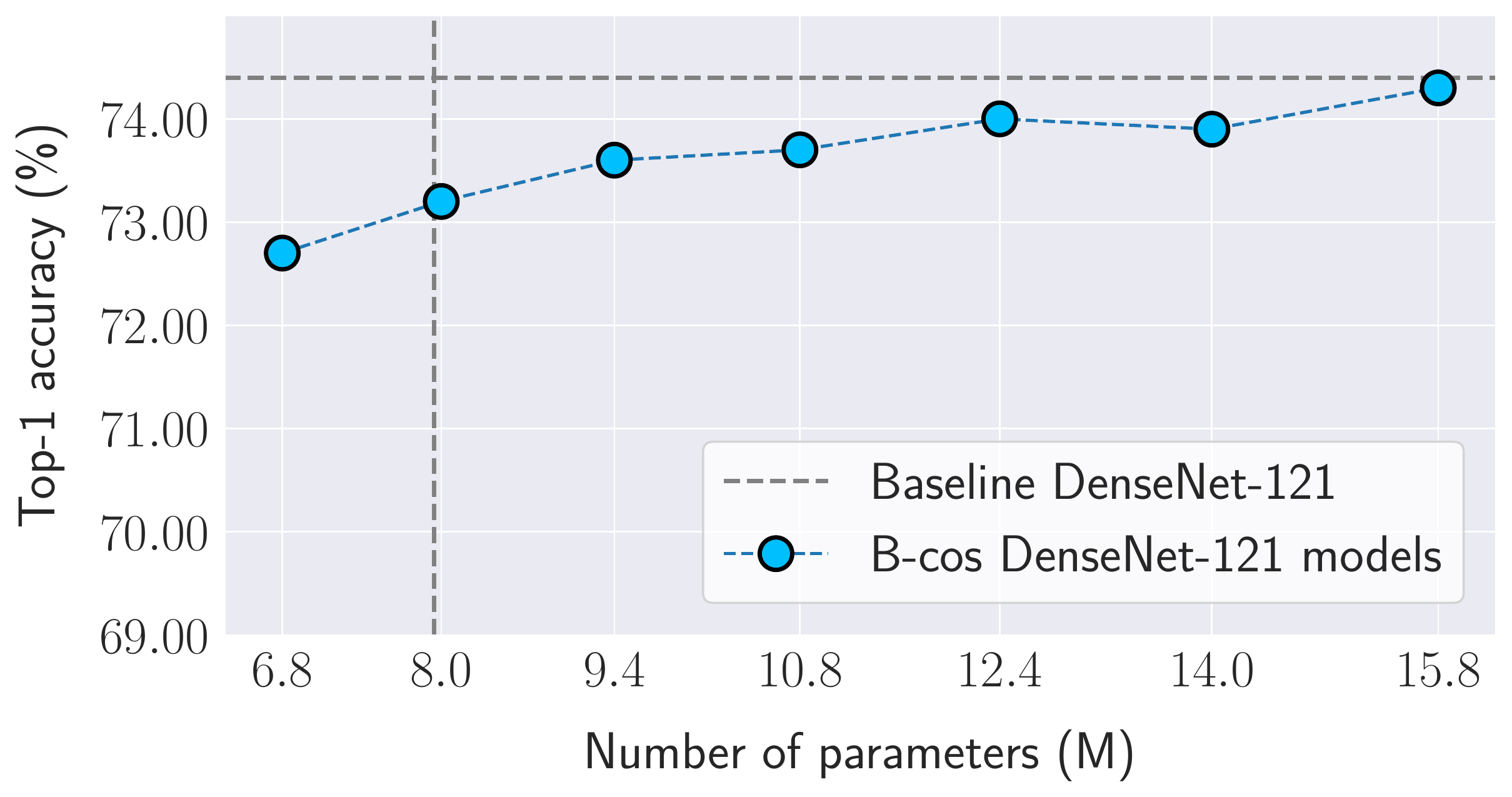}
            \end{subfigure}
            % \vspace{-2em}
            
        {%
          \caption{Top-1 accuracy on ImageNet of \bcos{} DenseNet-121 models of different sizes (i.e., \emph{growth factors}, see~\cite{huang2017densely}). These models were trained with 2 MaxOut units. For reference, we
          indicate the number of parameters and the accuracy results of a conventional DenseNet-121 model (Baseline, dashed lines).
            }
            % \vspace{-.5em}
            \label{fig:param_ablation}
        }
\end{adjustbox}
\hspace{2em}
\begin{adjustbox}{minipage=.45\linewidth, scale=1}
    \centering
    \vspace{2em}
    \captionsetup{type=table}
    \centering
          \input{supplement/resources/tables/param_ablation}
        \vspace{1.5em}
            % \vspace{-2em}
            
        {%
          \caption{Top-1 accuracies ($\%$) on ImageNet of \bcos{} DenseNet-121 models with different growth factors (g-factors), see~\cite{huang2017densely}, and with and without MaxOut. Note that the \bcos{} DenseNet-121 model without MaxOut does not employ any non-linearities other than \bcos{}. Results of a standard DenseNet-121 model shown for reference.
          }%
          \label{tbl:param_ablation}
        }

\end{adjustbox}

\end{center}%
\vspace{1em}
}

%% file: supplement/resources/tables/param_ablation.tex
% \begin{table}
    \centering
    {\setlength{\tabcolsep}{0.25em}\setlength\extrarowheight{-1pt}
    \begin{tabular}{|>{\centering\arraybackslash}  >{\centering\arraybackslash}p{1.85cm} |
    >{\centering\arraybackslash}p{.6cm} |
    >{\centering\arraybackslash}p{.6cm} |
    >{\centering\arraybackslash}p{.6cm} |
    >{\centering\arraybackslash}p{.6cm} |
    >{\centering\arraybackslash}p{.6cm} |
    >{\centering\arraybackslash}p{.6cm} |
    >{\centering\arraybackslash}p{.6cm} |
    >{\centering\arraybackslash}p{.6cm} |
    }
    \cline{2-9}
    % \hline
    % \footnotesize \textbf{Model} 
%   \footnotesize \bf Model type$\,\rightarrow$ 
    \multicolumn{1}{c}{} &\multicolumn{8}{|c|}{\footnotesize \bf \bcos{} DenseNet-121 models}
    \\\hline
  \multicolumn{1}{|c|}{\footnotesize \bf MaxOut}&\footnotesize \bf no &\multicolumn{7}{c|}{\footnotesize \bf Maxout with 2 units}
    \\\hline
    \footnotesize \textbf{g-factor}
    & {\footnotesize 32}
    & {\footnotesize 20}
    & {\footnotesize 22}
    & {\footnotesize 24}
    & {\footnotesize 26}
    & {\footnotesize 28}
    & {\footnotesize 30}
    & {\footnotesize 32}
    \\\hline
    \footnotesize \textbf{\#Params.~(M)}
    & {\footnotesize 7.9}
    & {\footnotesize 6.8}
    & {\footnotesize 8.0}
    & {\footnotesize 9.4}
    & {\footnotesize 10.8}
    & {\footnotesize 12.4}
    & {\footnotesize 14.0}
    & {\footnotesize 15.8}
    \\\hline
    \footnotesize \textbf{Accuracy (\%)}
    &\footnotesize 72.6 % single maxout
    &\footnotesize 72.8 % G=20
    &\footnotesize 73.2 % G=22
    &\footnotesize 73.6 % G=24
    &\footnotesize 73.7 % G=26
    &\footnotesize 74.0 % G=28
    &\footnotesize 73.9 % G=30
    &\footnotesize 74.3 % G=32
    \\\thickhline
    \multicolumn{9}{c}{\footnotesize }
    \\\hline
    \multicolumn{9}{|c|}{\footnotesize {\bf Standard DenseNet-121:} accuracy 74.4; parameters 7.9; g-factor 32}
    \\\hline
    \end{tabular}}

%% file: supplement/resources/figures/watermark.tex
\begin{figure}[h]
    \vspace{-1em}
    \centering
    \includegraphics[width=.7\textwidth]{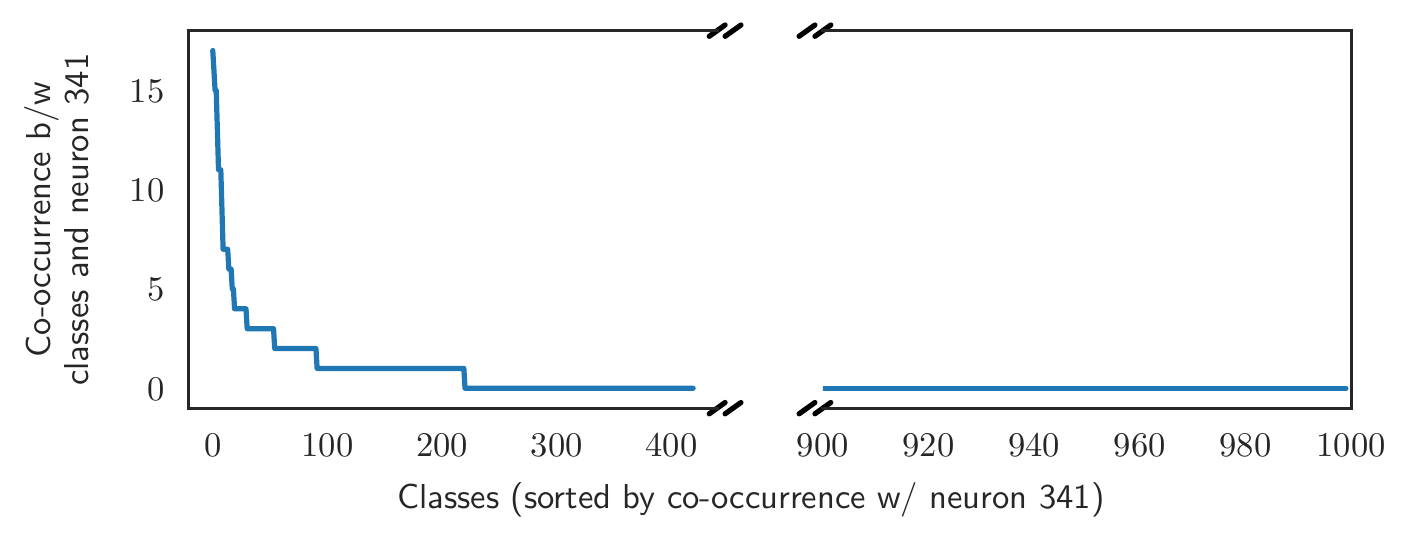}
    \caption{Class distribution among the images corresponding to the 500 highest activations of the watermark neuron (341). The co-occurrence distribution between classes and watermarks is indeed highly skewed and only a fraction of all classes is represented among these images. This indicates  
    %Under manual inspection, we found that 965 of these images indeed had a watermark, i.e., had text artificially overlayed over a natural image. As can be seen, only a fraction all classes is represented, indicating 
    high discriminative power of the watermark for classification.
    }
    \label{fig:watermark}
\end{figure}